\documentclass{article}

\usepackage{microtype}
\usepackage{graphicx}
\usepackage{subfigure}
\usepackage{booktabs} % for professional tables
\usepackage{bm,bbm}
\usepackage{multirow}
\usepackage{wrapfig}
\usepackage{threeparttable}
\usepackage{caption}

\usepackage{amsmath, amsthm, amssymb, multirow, paralist}

\usepackage[final]{neurips_2023}

\usepackage{amsmath}
\usepackage{amssymb}
\usepackage{mathtools}
\usepackage{amsthm}
\DeclareUnicodeCharacter{FF0C}{ }

\usepackage[utf8]{inputenc} % allow utf-8 input
\usepackage[T1]{fontenc}    % use 8-bit T1 fonts
\usepackage{hyperref}       % hyperlinks
\usepackage{url}            % simple URL typesetting
\usepackage{booktabs}       % professional-quality tables
\usepackage{amsfonts}       % blackboard math symbols
\usepackage{nicefrac}       % compact symbols for 1/2, etc.
\usepackage{microtype}      % microtypography
\usepackage{xcolor}         % colors

\usepackage[capitalize,noabbrev]{cleveref}
\usepackage{stfloats}
\def \R {\mathbb{R}}

\def \ph {\widehat{p}}

\def \V {\mathcal{V}}

\def \V {\mathcal{V}}

\def \U {\mathcal{U}}
\theoremstyle{plain}
\newtheorem{theorem}{Theorem}[section]

\theoremstyle{definition}

\theoremstyle{remark}

\newtheorem{thm}{Theorem}

\newtheorem{lemma}[theorem]{Lemma}

\usepackage[textsize=tiny]{todonotes}

\title{One Fits All:\\ Power General Time Series Analysis by Pretrained LM}

\author{%
Tian Zhou$^*$\quad Peisong Niu$^*$ \quad  Xue Wang$^*$ \quad Liang Sun \quad \textbf{Rong Jin}$^\dagger$ \\
\texttt{\{tian.zt,niupeisong.nps,xue.w,liang.sun,jinrong.jr\}@alibaba-inc.com }\\
}
\begin{document}
\maketitle

\let\thefootnote\relax\footnote{$*$ Equal contribution}
\let\thefootnote\relax\footnote{$\dagger$ Corresponding authors}

\begin{abstract}
Although we have witnessed great success of pre-trained models in natural language processing (NLP) and computer vision (CV), limited progress has been made for general time series analysis. Unlike NLP and CV where a unified model can be used to perform different tasks, specially designed approach still dominates in each time series analysis task such as classification, anomaly detection, forecasting, and few-shot learning. The main challenge that blocks the development of pre-trained model for time series analysis is the lack of a large amount of data for training. In this work, we address this challenge by leveraging language or CV models, pre-trained from billions of tokens, for time series analysis. Specifically, we refrain from altering the self-attention and feedforward layers of the residual blocks in the pre-trained language or image model. This model, known as the Frozen Pretrained Transformer (FPT), is evaluated through fine-tuning on all major types of tasks involving time series. Our results demonstrate that pre-trained models on natural language or images can lead to a comparable or state-of-the-art performance in all main time series analysis tasks, as illustrated in Figure~\ref{fig:representation}. We also found both theoretically and empirically that the self-attention module behaviors similarly to principle component analysis (PCA), an observation that helps explains how transformer bridges the domain gap and a crucial step towards understanding the universality of a pre-trained transformer. 
The code is publicly available at \url{https://github.com/DAMO-DI-ML/One_Fits_All}.

% Obtaining sufficient training data for downstream performance is a major obstacle for machine learning tasks that either demand a significant investment in data collection or are impracticable to acquire enough data, such as time series analysis. Additionally, the diversity and domain-specific characteristics of time series data pose substantial challenges in transferring learning to time series analysis tasks. To leverage the wealth of knowledge available from domains with rich data, we investigate the efficacy of using a transformer model that has been pre-trained on natural language or image data and subsequently fine-tuned for time series analysis tasks with minimal modifications. Specifically, we refrain from altering the self-attention and feedforward layers of the residual blocks. This model, known as the Frozen Pretrained Transformer (FPT), is evaluated through fine-tuning on time series analysis tasks including imputation, anomaly detection, classification, and especially forecasting under few-shot/normal/zero-shot settings. Our results demonstrate that pre-training on natural language or images can lead to a comparable or state-of-the-art performance in all cross-modality time series analysis tasks, in contrast to previous studies that focused on fine-tuning within the same modality as the pre-training data. Additionally, we provide a comprehensive theoretical analysis of the universality and the functionality of the FPT. 
%The code is publicly available at https://anonymous.4open.science/r/Pretrained-LM-for-TSForcasting-C561.

\begin{figure}[h]
    \centering
    \includegraphics[width=0.8\columnwidth]{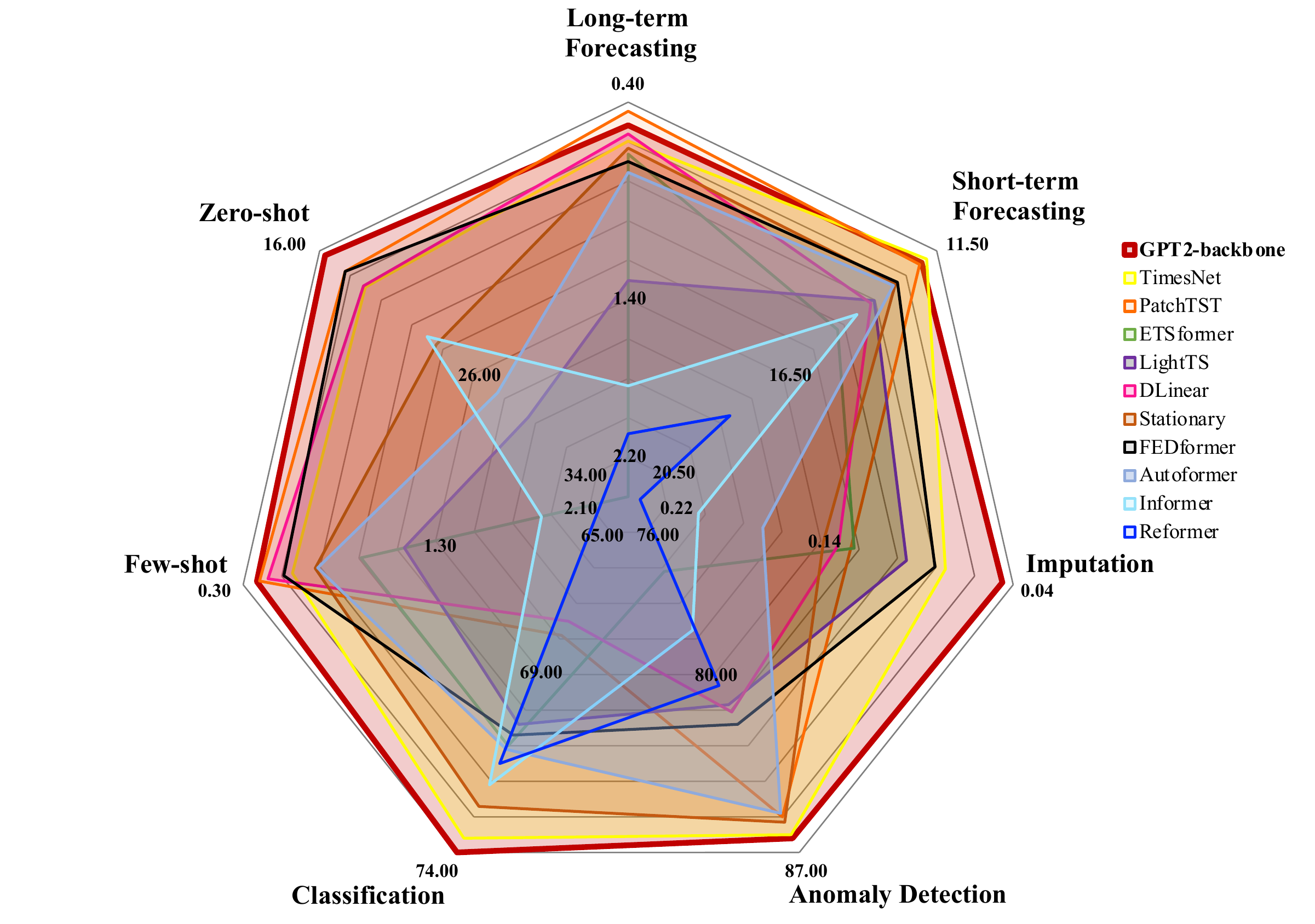}
    \captionsetup{font=small} 
    \caption{Model performance comparison on various tasks.}
    \label{fig:representation}
\end{figure}

\end{abstract}

\section{Introduction}

Time series analysis is a fundamental problem~\cite{hyndman:forecasting:3rd} that has played an important role in many real-world applications~\cite{wen2022robust}, such as retail sales forecasting~\cite{bose2017probabilistic,courty1999timing}
, imputation of missing data for economic time series ~\cite{Friedman1962}，anomaly detection for industrial maintenance~\cite{gao2020robusttad}, and classification of time series from various domain ~\cite{IsmailFawaz2018deep}. Numerous statistical and machine learning methods have been developed for time series analysis in the past. Inspired by its great success in natural language processing and computer vision~\cite{vaswani2017attention,Bert/NAACL/Jacob,Transformers-for-image-at-scale/iclr/DosovitskiyB0WZ21,DBLP:Global-filter-FNO-in-cv}, transformer has been introduced to various time series tasks with promising results~\cite{wen2022transformers}, especially for time series forecasting~\cite{lim2021temporal,zhou2022fedformer,zhou2021informer,wu2021autoformer,Patchformer}.

%numerous deep learning models have been developed recently shown their power in processing sequential data, especially in the area of natural language processing (NLP). Among these deep learning models, the recurrent neural network is probably the most well-examined~\cite {connor1994recurrent,hewamalage2021recurrent}. 

%healthcare~\cite{lim2018forecasting,zhang2018multi}, and engineering systems~\cite{zhang2019deep,gonzalez2019methodology}. 
% more intro to time series forecasting and methods SOTA. 
% Various deep models have been developed for such sequential forecasting, and among which recurrent neural network is probably mostly well examined~\cite{connor1994recurrent,hewamalage2021recurrent}. 
%Following the recent success in natural language process (NLP) and computer vision (CV) communities~\cite{vaswani2017attention,Bert/NAACL/Jacob,Transformers-for-image-at-scale/iclr/DosovitskiyB0WZ21,DBLP:Global-filter-FNO-in-cv}, 
% But such a paradigm is not as widely used in the field of time series, end-to-end learning algorithms still achieve SOTA performance in various cases. 

We have recently witnessed the rapid development of foundation models in NLP. The key idea is to pre-train a large language model from billions of tokens to facilitate model training for downstream tasks, particularly when we have a few, sometimes even zero, labeled instances. Another advantage of foundation models is that they provide a unified framework for handling diverse tasks, which contrasts conventional wisdom where each task requires a specially designed algorithm. However, so far, little progress has been made to exploit pre-trained or foundation models for time series analysis. One main challenge is the lack of the large amount of data to train a foundation model for time series analysis. The largest data sets for time series analysis is less than {\color{red}10GB}~\cite{godahewa2021monash}, which is much smaller than that for NLP. To address this challenge, we propose to leverage pre-trained language models for general time series analysis. Our approach provides a {\bf unified framework} for diverse time series tasks, such as classification, anomaly detection, forecasting, and few-shot or zero-shot learning. As shown in Figure~\ref{fig:representation}, using the same backbone, our approach performs either on-par or better than the state-of-the-art methods for all main time series analysis tasks. Besides extensive empirical studies, we also investigate why a transformer model pre-trained from the language domain can be adapted to time series analysis with almost no change. Our analysis indicates that the self-attention modules in the pre-trained transformer acquire the ability to perform certain non-data-dependent operations through training. These operations are closely linked to principal component analysis over the input patterns. We believe it is this generic function performed by the self-attention module that allows trained transformer models to be so-called universal compute engine~\cite{pretrained_transformer} or general computation calculator ~\cite{looped2023}. We support our claims by conducting an empirical investigation of the resemblance in model behaviors when self-attention is substituted with PCA, and by providing a theoretical analysis of their correlation.

Here we summarize our key contributions as follows:
\begin{enumerate}
    \item We propose a unified framework that uses a frozen pre-trained language model to achieve a SOTA or comparable performance in all major types of time series analysis tasks supported by thorough and extensive experiments, including time series classification, short/long-term forecasting, imputation, anomaly detection, few-shot and zero-sample forecasting. 
    % \item We demonstrate that our proposed algorithm can achieve a SOTA result in few-shot learning and zero-sample forecasting tasks in comparison with various recent baseline models. 
    \item We found, both theoretically and empirically, that self-attention performs a function similar to PCA, which helps explain the universality of transformer models. 
    \item We demonstrate the universality of our approach by exploring a pre-trained transformer model from another backbond model (BERT) or modality (computer vision) to power the time series forecasting. %decoder-only structure for video forecasting tasks,encoder-decoder structure for guided video forecasting tasks
    % \item We conduct extensive experiments over six benchmark datasets including two synthetic ones and four real-world datasets across multiple domains(precipitation, El Niño/Southern Oscillation (ENSO), and earth surface forecasting). Our empirical studies show that the proposed model achieves a state of art performance compared to various recent baselines. #这一段和introduction最后一段重复了,嗯嗯，我改写一下
\end{enumerate}
The remainder of this paper is structured as follows. Section 2 briefly summarizes the related work. Section 3 presents the proposed detailed model structure. In Section 4, we conduct a thorough and extensive evaluation of the performance of cross-modality time series analysis using our proposed method in seven main time series analysis tasks compared to various SOTA baseline models. Section 5 presents various ablation studies, and Section 6 demonstrates the universality of our proposed method using pre-trained models with another structure or pre-trained from another modality. In Section 7, we provide a theoretical explanation of the connection between self-attention and PCA. Finally, in Section 8, we discuss our results and future directions. Due to space limit, more extensive discussion of related work, experimental results, and theoretical analysis are provided in the Appendix.

\section{Related Work}

In this section, we provide short reviews of literature in the areas of time series analysis, in-modality transfer learning, and cross-modality knowledge transfer learning. We postpone the discussion of works for end-to-end time series analysis to Appendix \ref{section:related}, due to the limited space.

\paragraph{In-modality Transfer Learning through pre-trained models}
In recent years, a large number of research works have verified the effectiveness of the pre-trained model from NLP, CV to Vision-and-Language (VL).
Latest studies for NLP focus on learning contextual word embeddings for downstream tasks. With the increase of computing power, the very deep transformer models have shown powerful representation ability in various language tasks. Among them, BERT \cite{Bert/NAACL/Jacob} uses transformer encoders and employs masked language modeling task that aims to recover the random masked tokens within a text. OpenAI proposed GPT \cite{Radford2018ImprovingLU} that trains transformer decoders on a large language corpus and then fine-tunes on task-specific data. GPT2 \cite{gpt2-2019} is trained on larger datasets with much more parameters and can be transferred to various downstream tasks.
Since transformer models can adapt to various inputs, the idea of pre-training can also be well adapted to visual tasks. DEiT \cite{deit} proposed a teacher-student strategy for transformers with convolution neural networks (CNNs) as the teacher model and achieves competitive performance. BEiT \cite{bao2022beit} 
converts images as visual tokens and successfully uses the BERT model in CV. However, because of the \textbf{insufficient training sample}, there is little research on pre-trained models on general time series analysis that cover all major tasks like CV or NLP domain.

% MAE \cite{he2022mae} validates the feasibility of masked image encoding by simply masking out a high proportion of input patches and reconstructing the pixel values.
% Moreover, many pre-trained models have been designed for multi-modal VL representation \cite{radford2021clip, ALBEF, bao2021vlmo}. However, because of the \textbf{insufficient training sample}, there is little research work about pre-trained models on time series forecasting.

\paragraph{Cross-modality knowledge transfer}
Since transformers can handle different modal tasks through tokenizing the inputs to embeddings, it is also an interesting topic whether the transformers have universal representation ability and can be used for transferring between various domains.
The VL pre-trained model VLMo~\cite{bao2021vlmo} proposed a stagewise pre-training strategy that utilizes frozen attention blocks pre-trained by image-only data to train the language expert.
One of the most related works which transfer knowledge from a pre-trained language model to other domains is ~\cite{pretrained_transformer}, which studies the strong performance of a frozen pre-trained language model (LM) compared to an end-to-end transformer alternative learned from other domains' data. Another relative work for knowledge transfer to the time series is the Voice2series~\cite{yang2021voice2series}, which leverages a pre-trained speech processing model for time series classification and achieves superior performance. To the best of our knowledge, no previous research has investigated cross-modality knowledge transfer for the time series forecasting task, let alone general time series analysis.

% \begin{figure*}[!h]
%     \centering
%     \includegraphics[width=2\columnwidth]{graphs/model_structure.pdf}
%     \caption{Model architecture. NLP pretrained parameters are transferred to the time series forecasting tasks. Self-attention and Feedforward layers in the transformer blocks are frozen while only the embedding layer, normalization layers, and output layer require training.}
%     \label{fig:model_structure}
% \end{figure*}

\section{Methodology}

\subsection{Model Structure}

\begin{figure}[t]
    \centering
    \includegraphics[width=0.8\columnwidth]{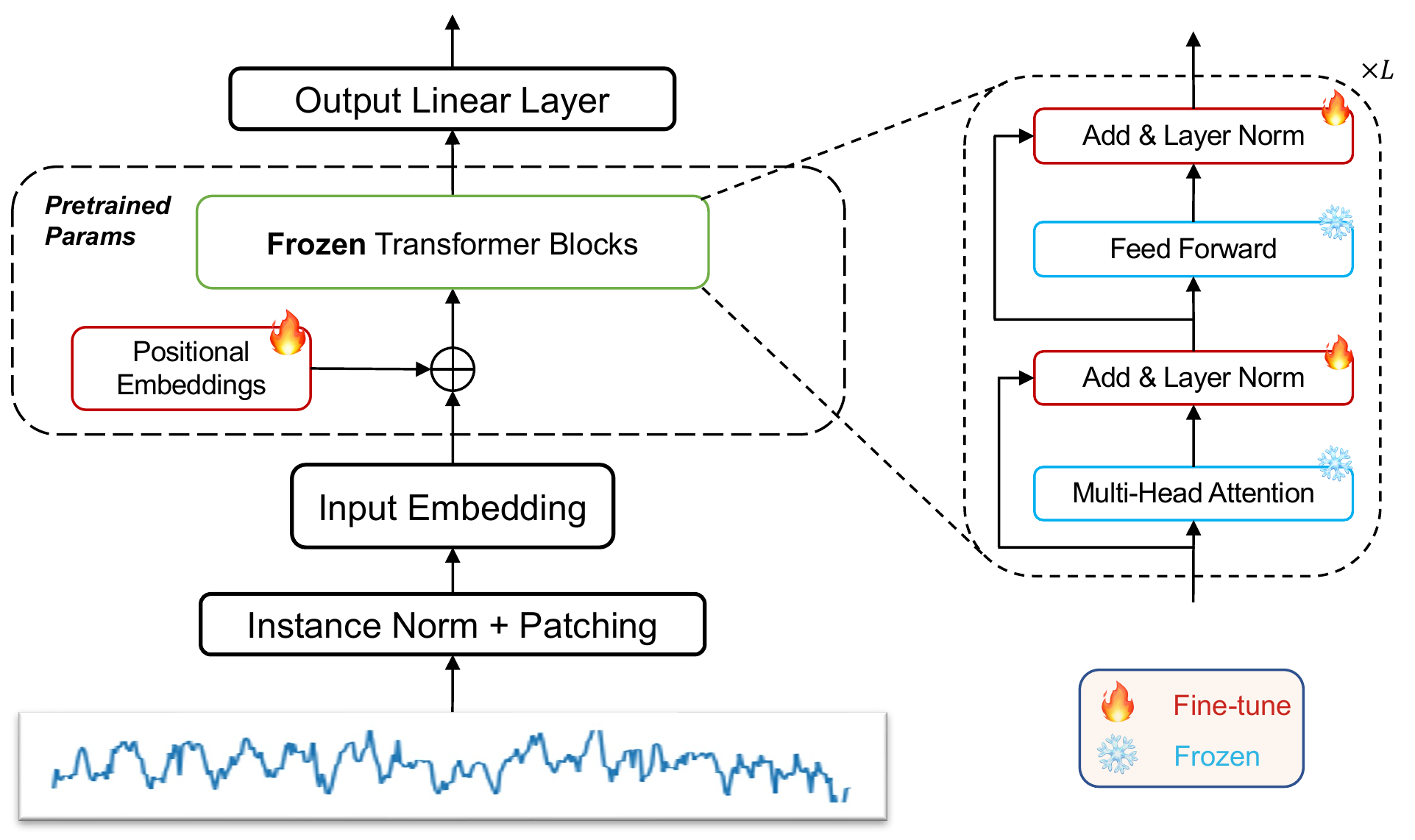}
    \captionsetup{font=small} 
    \caption{Model architecture. Pre-trained parameters are transferred to the time series forecasting tasks. Self-attention and Feedforward layers in the transformer blocks are frozen while only the embedding layer, normalization layers, and output layer require training.}
    \label{fig:structure_2}
    \vskip -0.25in
\end{figure}

The architecture we employ is depicted in Figure \ref{fig:structure_2}. We utilize parameters from NLP pretrained transformer models for time series analysis, with a focus on the GPT2 model \cite{gpt2-2019}. We also experiment with other models, such as BERT \cite{Bert/NAACL/Jacob} and BEiT \cite{bao2022beit}, to further demonstrate that the universal performance of cross-domain knowledge transfer exists in a wide range of pre-trained models.

\textbf{Frozen Pretrained Block}
Our architecture retains the positional embedding layers and self-attention blocks from the pre-trained models. As self-attention layers and FFN (Feedforward Neural Networks) contain the majority of learned knowledge from pre-trained language models, we opt to freeze the self-attention blocks while fine-tuning.

\textbf{Positional Embeddings and Layer Normalization}
To enhance downstream tasks with minimal effort, we fine-tune the positional embeddings and layer normalization layer, which is considered a standard practice\cite{pretrained_transformer,houlsby2019parameter}. As a result, we retrain these components during fine-tuning.

\textbf{Input Embedding}
Given our goal of applying the NLP pre-trained model to various tasks and a new modality, we must redesign and train the input embedding layer. This layer is responsible for projecting the time-series data to the required dimensions of the specific pre-trained model. To accomplish this, we use linear probing, which also reduces the number of parameters required for training.

\textbf{Normalization}
Data normalization is crucial for pre-trained models across various modalities. In addition to the layer norm utilized in pre-trained LM, we also incorporate a simple data normalization block, reverse instance norm \cite{kim2022reversible}, to further facilitate knowledge transfer. This normalization block simply normalizes the input time series using mean and variance, and then adds them back to the output.

\textbf{Patching}
To extract local semantic information, we utilize patching~\cite{Patchformer} by aggregating adjacent time steps to form a single patch-based token. Patching enables a significant increase in the input historical time horizon while maintaining the same token length and reducing information redundancy for transformer models. In our architecture, we apply patching after instance normalization.

\vspace{-.2cm}
\section{Main Time Series Analysis Tasks}
\vspace{-.2cm}

Our proposed method excels in various downstream time series analysis tasks through fine-tuning. To demonstrate the effectiveness of our approach, we conduct extensive experiments on major types of downstream tasks, including time series classification, anomaly detection, imputation, short/long-term forecasting and few-shot/zero-shot forecasting. To ensure a fair comparison, we use GPT2-backbone FPT and adhere to the experimental settings of TimesNet \cite{timesnet}. Due to the space limit, only the summarized results are presented below except zero-shot forecasting. Full experimental results of the other six downstream tasks can be found in Appendix ~\ref{appendix:full-data},~\ref{appendix:few-shot-learning},~\ref{app:zero-shot-full},~\ref{appendix:classification_full},~\ref{appendix:anomaly_full},~\ref{appendix:inputation_full},~\ref{appendix:short-term_full} respectively.

\textbf{Baselines}
We select representative baselines and cite their results from \cite{timesnet}, which includes the most recent and quite extensive empirical studies of time series. The baselines include CNN-based models: TimesNet~\cite{timesnet}; MLP-based models: LightTS~\cite{lightts} and DLinear~\cite{dlinear}; Transformer-based models: Reformer~\cite{reformer}, Informer~\cite{zhou2021informer}, Autoformer~\cite{wu2021autoformer}, FEDformer~\cite{zhou2022fedformer}, Non-stationary Transformer~\cite{non-stationary}, ETSformer~\cite{woo2022etsformer}, PatchTST~\cite{Patchformer}. Besides, N-HiTS~\cite{nhits} and N-BEATS~\cite{n-beats} are used for short-term forecasting. Anomaly Transformer~\cite{xu2021anomaly} is used for anomaly detection. XGBoost~\cite{xgboost}, Rocket~\cite{ROCKET}, LSTNet~\cite{lstnet}, LSSL~\cite{lssl}, Pyraformer \cite{pyraformer}, TCN~\cite{tcn} and Flowformer \cite{huang2022flowformer} are used for classification.

\subsection{Main Results}

Overall, as shown in Figure \ref{fig:representation}, GPT2-backbone FPT outperforms other models in most tasks, including long/short-term forecasting, classification, anomaly detection, imputation, and fow-shot/zero-short forecasting. This confirms that time series tasks can also take advantage of cross-modality transferred knowledge. In the following, we use GPT2(K) to represent GPT2-backbone with first K Layers.

\subsection{Imputation}

\noindent{\bf Setups} 
We conduct experiments on six popular real-world datasets, including 4 ETT datasets \cite{zhou2021informer} (ETTh1, ETTh2, ETTm1, ETTm2), Electricity\footnote{https://archive.ics.uci.edu/ml/datasets/ElectricityLoadDiagrams 20112014} and Weather\footnote{https://www.bgc-jena.mpg.de/wetter/}, where the data-missing is common. Following the settings of TimesNet, different random mask ratios (\{12.5\%, 25\%, 37.5\%, 50\%\}) of time points are selected for the evaluation on various proportions of missing data.

\noindent{\bf Results} 
The results are shown in Table \ref{tab:imputation} that GPT2(3) FPT achieves the best performance on most datasets. Particularly, compared to the previous SOTA TimesNet, GPT2(3) FPT yields a relative \textbf{11.5\%} MSE reduction on ETTh1,and a \textbf{4.1\%} MSE reduction on average on six benchmark datasets.
It verifies that the proposed method can also effectively mine temporal patterns of incomplete time series.

\begin{table}[h]
\captionsetup{font=small} 
\caption{Imputation task. We randomly mask \{12.5\%, 25\%, 37.5\%, 50\%\} time points of 96-length time series. The results are averaged from 4 different mask ratios. 
% GPT2(3) represent GPT2-backbone (3 Layers) FPT 
\textbf{Black}: best,  {\color{red}\textbf{Red}}: second best. Appendix \ref{appendix:inputation_full} shows the full results.}
\label{tab:imputation}
%\vskip 0.15in
\begin{center}
\begin{small}
\scalebox{0.7}{
\setlength\tabcolsep{3pt}
\begin{tabular}{c|cc|cc|cc|cc|cc|cc|cc|cc|cc|cc|cc}
\toprule

\multirow{2}{*}{Methods} 
&\multicolumn{2}{c|}{GPT2(3)} & \multicolumn{2}{c|}{TimesNet}&\multicolumn{2}{c|}
{PatchTST}&\multicolumn{2}{c|}
{ETSformer}&\multicolumn{2}{c|}{LightTS}&\multicolumn{2}{c|}{DLinear}&\multicolumn{2}{c|}{FEDformer}&\multicolumn{2}{c|}{Stationary}&\multicolumn{2}{c|}{Autoformer}&\multicolumn{2}{c|}{Informer}&\multicolumn{2}{c}{Reformer} \\
&MSE&MAE&MSE&MAE&MSE&MAE&MSE&MAE&MSE&MAE&MSE&MAE&MSE&MAE&MSE&MAE&MSE&MAE&MSE&MAE&MSE&MAE \\

\midrule

ETTm1&\color{red}\textbf{0.028}&\textbf{0.105}&\textbf{0.027} &\color{red}\textbf{0.107}&0.047 &0.140 & 0.120& 0.253& 0.104& 0.218& 0.093& 0.206& 0.062& 0.177& 0.036& 0.126&0.051& 0.150&  0.071& 0.188 & 0.055 & 0.166 \\
ETTm2&\textbf{0.021}&\textbf{0.084}&\color{red}\textbf{0.022}& \color{red}\textbf{0.088}&0.029 &0.102 & 0.208& 0.327& 0.046& 0.151& 0.096& 0.208& 0.101& 0.215& 0.026& 0.099 &0.029& 0.105& 0.156& 0.292 & 0.157 & 0.280\\
ETTh1&\textbf{0.069}&\textbf{0.173}&\color{red}\textbf{0.078}& \color{red}\textbf{0.187}&0.115 &0.224 & 0.202& 0.329& 0.284& 0.373& 0.201& 0.306& 0.117& 0.246& 0.094& 0.201 &0.103& 0.214& 0.161& 0.279&0.122& 0.245\\
ETTh2&\textbf{0.048}&\textbf{0.141}&\color{red}\textbf{0.049}& \color{red}\textbf{0.146}&0.065 &0.163 & 0.367& 0.436& 0.119& 0.250& 0.142& 0.259& 0.163& 0.279& 0.053& 0.152& 0.055& 0.156& 0.337& 0.452&0.234& 0.352\\
ECL&\color{red}\textbf{0.090}&\color{red}\textbf{0.207}&0.092&0.210&\textbf{0.072} &\textbf{0.183} & 0.214& 0.339& 0.131& 0.262& 0.132& 0.260& 0.130& 0.259& 0.100& 0.218 &0.101& 0.225& 0.222& 0.328&0.200& 0.313 \\
Weather&\color{red}\textbf{0.031}&\color{red}\textbf{0.056}&\textbf{0.030}& \textbf{0.054}&0.034 &0.055 & 0.076& 0.171& 0.055& 0.117& 0.052& 0.110& 0.099& 0.203& 0.032& 0.059 &0.031& 0.057& 0.045& 0.104&0.038& 0.087 \\

\midrule

Average & \textbf{0.047} & \textbf{0.127} & \color{red}\textbf{0.049} & \color{red}\textbf{0.132} & 0.060 & 0.144 & 0.197 & 0.309 & 0.123 & 0.228 & 0.119 & 0.224 & 0.112 & 0.229 & 0.056 & 0.142 & 0.061 & 0.151 & 0.165 & 0.273 & 0.134 & 0.240 \\

\bottomrule

\end{tabular}
}
\end{small}
\end{center}
\end{table}

\subsection{Time Series Classification}

\begin{wrapfigure}[15]{r}{0.4\textwidth}
    \centering
    \includegraphics[width=.4\textwidth]{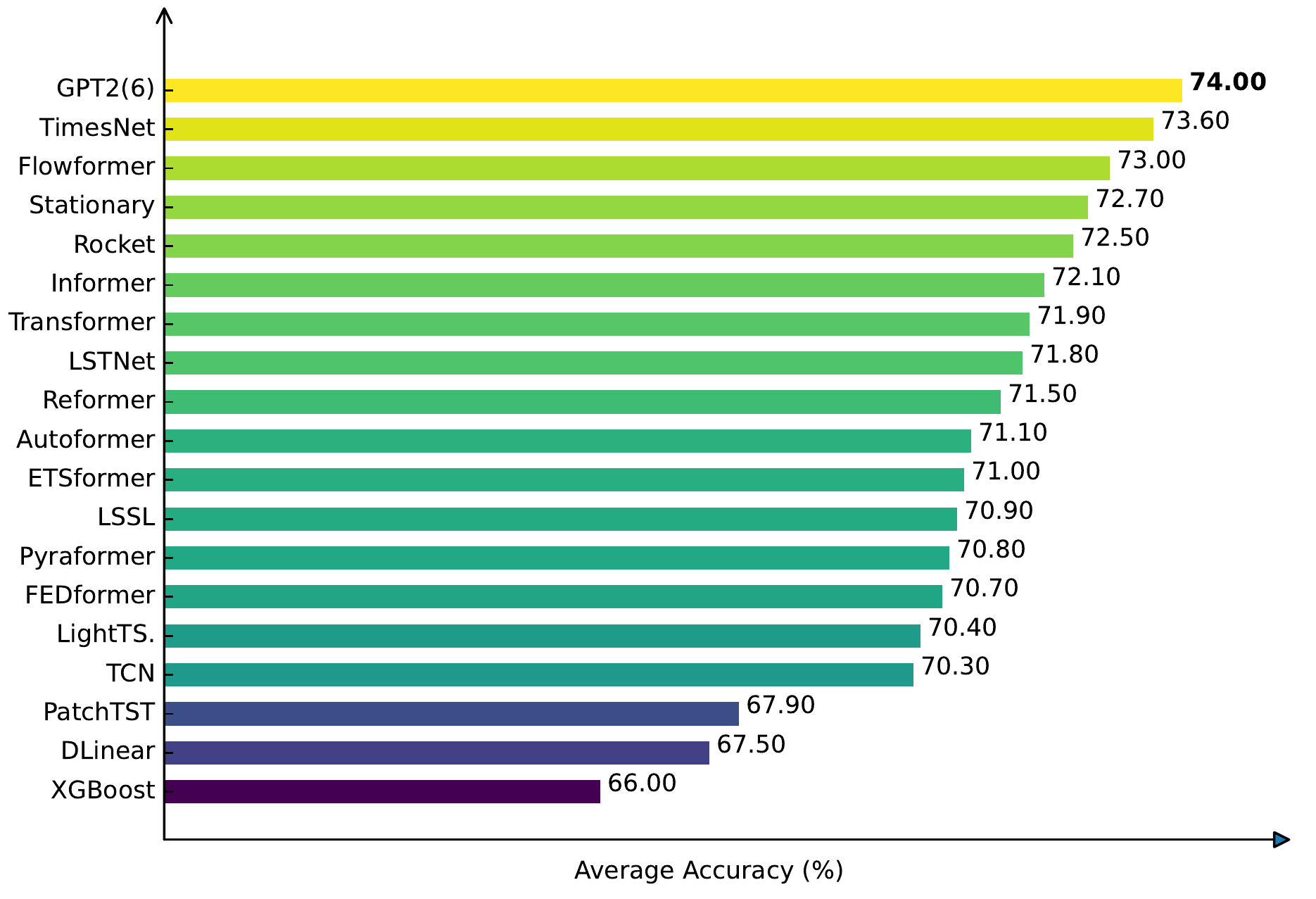}
    \caption{Model comparison in classification. The results are averaged from 10 subsets of UEA. Appendix \ref{appendix:classification_full} shows the full results.}
    \label{fig:classification}
\end{wrapfigure}
\textbf{Setups} 
% Time series classification has a wide range of applications, including recognition and medical diagnosis ~\cite{moody2011}. 
To evaluate the model's capacity for high-level representation learning, we employ sequence-level classification. Specifically, we follow the same setting as TimesNet: For classification, 10 multivariate UEA classification datasets \cite{UEA} are selected for evaluation, including gesture, action, audio recognition medical diagnosis and other practical tasks.

\noindent{\bf Results} As shown in Figure \ref{fig:classification}, GPT2(6) 
FPT achieves an average accuracy of 74.00\%, surpassing all baselines including TimesNet (73.60\%). Specifically, compared to recent published patch-transformer-based models \cite{Patchformer} , GPT2(6) FPT surpasses it by a large margin \textbf{9.0\%} which shows the prior NLP transfer knowledge can indeed help in time series representation. 
%Also, compared to similarly transformer-based models, GPT2(6) FPT with prior NLP transfered knowledge can obtain widely spread representation, effectively mapping to their respective clusters.

% \begin{figure}[h]
%     \centering
%     \includegraphics[width=0.65\columnwidth]{graphs/classification.pdf}
%     \caption{Model comparison in classification. $\ast$. in the Transformers indicates the name of $\ast$former. The results are averaged from 10 subsets of UEA. GPT2(6) represent GPT2-backbone 6 layers model. Table \ref{tab:classification} shows full results.}
%     \label{fig:classification}
% \end{figure}

\subsection{Time Series Anomaly Detection}

\noindent{\bf Setups} 
Detecting anomalies in time series is vital in industrial applications, ranging from health monitoring to space \& earth exploration. We compare models on five commonly used datasets, including SMD\cite{SMD}, MSL\cite{MSL_SMAP}, SMAP\cite{MSL_SMAP}, SWaT\cite{SWaT} and PSM\cite{PSM}. 
To perform a fair comparison, only the classical reconstruction error is used for all baseline models to the make the setting the same as TimesNet.

\noindent{\bf Results} 
Table \ref{tab:anomaly} demonstrates that GPT2(6) FPT also achieves the best performance with the averaged F1-score \textbf{86.72\%}, surpassing previous SOTA method TimesNet by \textbf{1.7\%}. Thus, in addition to its proficiency in representing complete sequences for classification purposes, GPT2(6) FPT is capable of detecting infrequent anomalies within time series.

\begin{table*}[h]
\vskip -0.1in
\captionsetup{font=small} 
\caption{Anomaly detection task. We calculate the F1-score (as \%) for each dataset. $\ast$. in the Transformers indicates the name of $\ast$former. 
% GPT2(6) represent GPT2-backbone (6 Layers) FPT. 
 \textbf{Black}: best,  {\color{red}\textbf{Red}}: second best. Appendix \ref{appendix:anomaly_full} shows the full results.}
\label{tab:anomaly}
%\vskip 0.15in
\begin{center}
\begin{small}
\scalebox{0.72}{
\setlength\tabcolsep{3pt}
% \begin{threeparttable}
\begin{threeparttable}[b]
\begin{tabular}{c|ccccccccccccccc}
\toprule

\multirow{2}{*}{Methods} 
& GPT2(6) & \multirow{2}{*}{TimesNet\tnote{*}} & \multirow{2}{*}{PatchTS.} & \multirow{2}{*}{ETS.} & \multirow{2}{*}{FED.} & \multirow{2}{*}{LightTS} & \multirow{2}{*}{DLinear} & \multirow{2}{*}{Stationary} & \multirow{2}{*}{Auto.} & \multirow{2}{*}{Pyra.} & \multirow{2}{*}{Anomaly.\tnote{**}} & \multirow{2}{*}{In.} & \multirow{2}{*}{Re.} & \multirow{2}{*}{LogTrans.} & \multirow{2}{*}{Trans.} \\
&Ours&&&&&&&&&&&&&& \\

\midrule
SMD &86.89 &84.61 &84.62&83.13& 85.08& 82.53& 77.10 &84.72 &85.11 &83.04& 85.49 &81.65 &75.32& 76.21& 79.56 \\
MSL & 82.45&81.84&78.70&85.03&78.57&78.95&84.88&77.50&79.05&84.86&83.31&84.06&84.40&79.57&78.68 \\
SMAP&72.88&69.39&68.82& 69.50& 70.76& 69.21& 69.26& 71.09& 71.12& 71.09& 71.18& 69.92& 70.40& 69.97& 69.70 \\
SWaT&94.23& 93.02&85.72 & 84.91& 93.19 &93.33& 87.52& 79.88& 92.74& 91.78& 83.10& 81.43& 82.80& 80.52& 80.37 \\
PSM&97.13&  97.34&96.08& 91.76& 97.23& 97.15& 93.55& 97.29& 93.29& 82.08& 79.40& 77.10& 73.61 &76.74 &76.07 \\
\midrule
Average &\textbf{86.72}&\color{red}\textbf{85.24}&82.79& 82.87& 84.97& 84.23& 82.46& 82.08& 84.26& 82.57 &80.50& 78.83& 77.31& 76.60& 76.88\\

\bottomrule
\end{tabular}

\begin{tablenotes}
 \item[*] We reproduce the results of TimesNet by \href{https://github.com/thuml/Time-Series-Library}{https://github.com/thuml/Time-Series-Library}. 
 \item[**] We replace the joint criterion in Anomaly Transformer with reconstruction error for fair comparison.
\end{tablenotes}
\end{threeparttable}

}
\end{small}
\end{center}

\vskip -0.1in

% \begin{tablenotes}
%      \item[1] tablefootnote 1
%      \item[2] tablefootnote 2
%    \end{tablenotes}
% \end{threeparttable}

\end{table*}

\subsection{Long-term Forecasting}

\noindent{\bf Setups}
Eight popular real-world benchmark datasets~\cite{timesnet}, including Weather, Traffic \footnote{http://pems.dot.ca.gov}, Electricity, ILI \footnote{https://gis.cdc.gov/grasp/fluview/fluportaldashboard.html}, and 4 ETT datasets (ETTh1, ETTh2, ETTm1, ETTm2), are used for long-term forecasting evaluation. 
Additional information regarding the discussion on the input length setting can be found in the appendix \ref{app:input_length}.

% For the long-term forecasting setting, we use eight popular real-world datasets ~\cite{timesnet}, including Weather, Traffic \footnote{http://pems.dot.ca.gov}, Electricity, ILI \footnote{https://gis.cdc.gov/grasp/fluview/fluportaldashboard.html}, and 4 ETT datasets (ETTh1, ETTh2, ETTm1, ETTm2).

\noindent{\bf Results} 
As shown in Table \ref{tab:100_percent_ett_main}, GPT2(6) FPT achieves comparable performance with PatchTST and outperforms other baselines. Specifically, compared with recent published SOTA method TimesNet, GPT2(6) FPT yields a relative \textbf{9.3\%} average MSE reduction.

\begin{table}[h]
\captionsetup{font=small} 
\caption{Long-term forecasting task. All the results are averaged from 4 different prediction lengths, that is \{24, 36, 48, 60\} for ILI and \{96, 192, 336, 720\} for the others. 
% GPT2(6) represent GPT2-backbone (6 Layers) FPT. 
\textbf{Black}: best,  {\color{red}\textbf{Red}}: second best. Appendix~\ref{appendix:full-data} shows the full results.}
\label{tab:100_percent_ett_main}
%\vskip 0.15in
% \begin{center}
\begin{small}
\scalebox{0.7}{
\setlength\tabcolsep{3pt}
\begin{tabular}{c|cc|cc|cc|cc|cc|cc|cc|cc|cc|cc|cc}
\toprule

\multirow{2}{*}{Methods} 
&\multicolumn{2}{c|}{GPT2(6)}  & \multicolumn{2}{c|}{TimesNet}&\multicolumn{2}{c|}{ETSformer}&\multicolumn{2}{c|}{LightTS}&\multicolumn{2}{c|}{DLinear}&\multicolumn{2}{c|}{FEDformer}&\multicolumn{2}{c|}{PatchTST}&\multicolumn{2}{c|}{Stationary}&\multicolumn{2}{c|}{Autoformer}&\multicolumn{2}{c|}{Informer}&\multicolumn{2}{c}{Reformer} \\
&MSE&MAE&MSE&MAE&MSE&MAE&MSE&MAE&MSE&MAE&MSE&MAE&MSE&MAE&MSE&MAE&MSE&MAE&MSE&MAE&MSE&MAE \\

\midrule

Weather &\color{red}\textbf{0.237} & \color{red}\textbf{0.270} & 0.259& 0.287& 0.271& 0.334& 0.261& 0.312&0.249 & 0.300 & 0.309& 0.360&\textbf{0.225} &\textbf{0.264} &0.288& 0.314& 0.338& 0.382& 0.634& 0.548&0.803&0.656\\
ETTh1 & \color{red}\textbf{0.427}& \textbf{0.426}& 0.458& 0.450& 0.542& 0.510& 0.491& 0.479&0.423 &0.437 & 0.440& 0.460&\textbf{0.413}&\color{red}\textbf{0.430} & 0.570& 0.537& 0.496& 0.487& 1.040& 0.795&1.029&0.915\\
ETTh2 & \color{red}\textbf{0.346}& \color{red}\textbf{0.394}& 0.414& 0.427& 0.439& 0.452& 0.602& 0.543&0.431 & 0.447& 0.437& 0.449&\textbf{0.330} &\textbf{0.379} &0.526& 0.516& 0.450& 0.459& 4.431& 1.729&6.736&2.191\\
ETTm1 &\color{red}\textbf{0.352} & \color{red}\textbf{0.383}& 0.400 &0.406 &0.429 &0.425 &0.435 &0.437 &0.357 &\textbf{0.378} &0.448 &0.452 & \textbf{0.351}&0.387 &0.481 &0.456& 0.588 &0.517 &0.961 &0.734&0.799&0.671\\
ETTm2 &\color{red}\textbf{0.266} & \color{red}\textbf{0.326}& 0.291& 0.333& 0.293& 0.342& 0.409& 0.436& 0.267& 0.334& 0.305& 0.349&\textbf{0.255} & \textbf{0.315}& 0.306& 0.347& 0.327& 0.371& 1.410& 0.810&1.479&0.915\\
ILI &\color{red}\textbf{1.925} &\color{red}\textbf{0.903} & 2.139& 0.931& 2.497& 1.004& 7.382& 2.003& 2.169&1.041 & 2.847& 1.144&\textbf{1.443} &\textbf{0.798} &2.077& 0.914& 3.006& 1.161& 5.137& 1.544&4.724&1.445\\
ECL &0.167 &\color{red}\textbf{0.263} & 0.192& 0.295& 0.208& 0.323& 0.229& 0.329&\color{red}\textbf{0.166} &\color{red}\textbf{0.263} & 0.214& 0.327&\textbf{0.161} &\textbf{0.253} &0.193& 0.296& 0.227& 0.338& 0.311& 0.397&0.338&0.422\\
Traffic &\color{red}\textbf{0.414} & \color{red}\textbf{0.294} & 0.620& 0.336& 0.621& 0.396& 0.622& 0.392& 0.434& 0.295& 0.610& 0.376&\textbf{0.390} &\textbf{0.264} &0.624& 0.340& 0.628& 0.379& 0.764& 0.416&0.741&0.422\\
% ETTm1&\color{red}\textbf{0.028}&\textbf{0.105}&\textbf{0.027} &\color{red}\textbf{0.107}& 0.120& 0.253& 0.104& 0.218& 0.093& 0.206& 0.062& 0.177& 0.036& 0.126&0.051& 0.150&  0.071& 0.188 & 0.055 & 0.166 \\
% ETTm2&\textbf{0.021}&\textbf{0.084}&\color{red}\textbf{0.022}& \color{red}\textbf{0.088}& 0.208& 0.327& 0.046& 0.151& 0.096& 0.208& 0.101& 0.215& 0.026& 0.099 &0.029& 0.105& 0.156& 0.292 & 0.157 & 0.280\\
% ETTh1&\textbf{0.069}&\textbf{0.173}&\color{red}\textbf{0.078}& \color{red}\textbf{0.187}& 0.202& 0.329& 0.284& 0.373& 0.201& 0.306& 0.117& 0.246& 0.094& 0.201 &0.103& 0.214& 0.161& 0.279&0.122& 0.245\\
% ETTh2&\textbf{0.048}&\textbf{0.141}&\color{red}\textbf{0.049}& \color{red}\textbf{0.146}& 0.367& 0.436& 0.119& 0.250& 0.142& 0.259& 0.163& 0.279& 0.053& 0.152& 0.055& 0.156& 0.337& 0.452&0.234& 0.352\\
% Electricity&\textbf{0.090}&\textbf{0.207}&\color{red}\textbf{0.092}& \color{red}\textbf{0.210}& 0.214& 0.339& 0.131& 0.262& 0.132& 0.260& 0.130& 0.259& 0.100& 0.218 &0.101& 0.225& 0.222& 0.328&0.200& 0.313 \\
% Weather&\color{red}\textbf{0.031}&\color{red}\textbf{0.056}&\textbf{0.030}& \textbf{0.054}& 0.076& 0.171& 0.055& 0.117& 0.052& 0.110& 0.099& 0.203& 0.032& 0.059 &\color{red}\textbf{0.031}& 0.057& 0.045& 0.104&0.038& 0.087 \\
\midrule

Average & \color{red}\textbf{0.516} & \color{red}\textbf{0.407} & 0.596 & 0.433 & 0.662 & 0.473 & 1.303 & 0.616 & 0.562 & 0.436 & 0.701 & 0.489 & \textbf{0.446} & \textbf{0.386} & 0.633 & 0.465 & 0.757 & 0.511 & 1.836 & 0.871 & 2.081 & 0.954 \\

\bottomrule

\end{tabular}
}
\end{small}
% \end{center}
\end{table}

\subsection{Short-term Forecasting}

\noindent{\bf Setups} 
To fully evaluate different algorithms in forecasting tasks, we also conduct short-term forecasting (with relatively short forecasting horizon) experiments on M4 \cite{makridakis2018m4}, contains marketing data of various frequencies.

\noindent{\bf Results} 
The results in Table \ref{tab:short_term} show that the performance of GPT2-backbone (6) FPT is superior to advanced Transformer-based and MLP-based models, and comparable to TimesNet and N-BEATS.

\begin{table*}[h]
\vskip -0.10in
\captionsetup{font=small} 
\caption{Short-term forecasting task on M4. The prediction lengths are in [6, 48] and results are weighted averaged from several datasets under different sample intervals. 
% GPT2(6) represent GPT2-backbone (6 Layers) FPT 
\textbf{Black}: best,  {\color{red}\textbf{Red}}: second best. Appendix \ref{appendix:short-term_full} shows the full results.}
\label{tab:short_term}
%\vskip 0.15in
\begin{center}
\begin{small}
\scalebox{0.7}{
\setlength\tabcolsep{3pt}
\begin{tabular}{c|ccccccccccccc}
\toprule

Methods&GPT2(6)&TimesNet&PatchTST&N-HiTS&N-BEATS& ETSformer& LightTS& DLinear &FEDformer &Stationary &Autoformer  &Informer&Reformer \\

% \multirow{2}{*}{Methods} 
% &\multicolumn{2}{c|}{GPT2(6)} & \multicolumn{2}{c|}{TimesNet}&\multicolumn{2}{c|}{ETSformer}&\multicolumn{2}{c|}{ETSformer}&\multicolumn{2}{c|}{LightTS}&\multicolumn{2}{c|}{DLinear}&\multicolumn{2}{c|}{FEDformer}&\multicolumn{2}{c|}{Stationary}&\multicolumn{2}{c|}{Autoformer}&\multicolumn{2}{c}{Informer}&\multicolumn{2}{c}{Reformer} \\

\midrule

SMAPE &11.991 & \textbf{11.829}&12.059& 11.927& \color{red}\textbf{11.851}& 14.718& 13.525& 13.639 &12.840 &12.780 &12.909 &14.086 &18.200 \\
MASE & 1.600 & \textbf{1.585}&1.623 & 1.613 & \color{red}\textbf{1.599} &2.408 &2.111 &2.095 &1.701 &1.756 &1.771  &2.718 &4.223\\
OWA &0.861 & \textbf{0.851}&0.869 &0.861 &\color{red}\textbf{0.855} &1.172 &1.051 &1.051 &0.918 &0.930 &0.939 & 1.230 & 1.775\\

\bottomrule

\end{tabular}
}
\end{small}
\end{center}
\vskip -0.1in
\end{table*}

\subsection{Few-shot Forecasting} 
The large language model (LLM) has demonstrated remarkable performance in both few-shot and zero-shot learning settings ~\cite{Brown2020LanguageMA,OpenAI2023GPT4TR}. It can be argued that few-shot and zero-shot learning also represent the ultimate tasks for a universal time series forecasting model. To extensively evaluate the representation power of the GPT2(6) for time series analysis, we conduct experiments under few-shot and zero-shot learning settings. 

Similar to traditional experimental settings, each time series is split into three parts: training data, validation data, and test data. For few-shot learning, only a certain percentage (10\%, 5\%) timesteps of training data are used.

The results of 10\% few-shot learning are shown in Table \ref{tab:few_shot_main}. Compared to TimesNet, DLinear, PatchTST and other methods, GPT2(6) FPT achieves the best performance. Traditionally, CNN-based and single MLP-based models are considered more data-efficient for training and suitable for few-shot learning methods. In comparison to convolution-based TimesNet and MLP-based DLinear models, GPT2(6) FPT demonstrates a relative average MSE reduction of \textbf{33.3\%} and \textbf{13.5\%} respectively. We add a comparison with traditional algorithms (ETS, ARIMA, NaiveDrift) in the Appendix \ref{app:classical_few_shot} as well, and GTP2(6)FPT also surpass all those traditional methods.

\begin{table}[h]
\captionsetup{font=small} 
\caption{Few-shot learning task on 10\% data. All the results are averaged from 4 different prediction lengths (\{96, 192, 336, 720\}). 
% GPT2(6) represent GPT2-backbone (6 layers) FPT 
\textbf{Black}: best,  {\color{red}\textbf{Red}}: second best. Appendix~\ref{appendix:few-shot-learning} shows the detailed results of 10\% and 5\% data.}
\label{tab:few_shot_main}
% \vskip 0.15in
% \begin{center}
\begin{small}
\scalebox{0.65}{
\setlength\tabcolsep{3pt}
\begin{tabular}{c|cc|cc|cc|cc|cc|cc|cc|cc|cc|cc|cc}
\toprule

\multirow{2}{*}{Methods} 
&\multicolumn{2}{c|}{GPT2(6)}  & \multicolumn{2}{c|}{TimesNet}&\multicolumn{2}{c|}{DLinear}&\multicolumn{2}{c|}{FEDformer}&\multicolumn{2}{c|}{PatchTST}&\multicolumn{2}{c|}{Autoformer}&\multicolumn{2}{c|}{Stationary}&\multicolumn{2}{c|}{ETSformer}&\multicolumn{2}{c|}{LightTS}&\multicolumn{2}{c|}{Informer}&\multicolumn{2}{c}{Reformer} \\
&MSE&MAE&MSE&MAE&MSE&MAE&MSE&MAE&MSE&MAE&MSE&MAE&MSE&MAE&MSE&MAE&MSE&MAE&MSE&MAE&MSE&MAE \\

\midrule

Weather &0.238&0.275&0.279&0.301&0.301&0.283&0.284&0.324&0.241&0.279&0.300&0.342&0.318&0.322& 0.317 & 0.359 & 0.289 & 0.322 & 0.597 & 0.494 & 0.545 & 0.469\\
ETTh1  &0.590&0.524&0.869&0.628&0.691&0.599&0.638&0.561&0.633&0.542&0.701&0.596&0.914&0.639& 1.179 & 0.833 & 1.375 & 0.877 & 1.199 & 0.808 &1.249&0.833\\
ETTh2  &0.397&0.421&0.479&0.465&0.608&0.538&0.466&0.475&0.415&0.431&0.488&0.499&0.461&0.454& 0.893 & 0.713 & 2.655 & 1.159 & 3.871 & 1.512 &3.485&1.485 \\
ETTm1  &0.464&0.441&0.676&0.537&0.411&0.429&0.721&0.605&0.501&0.466&0.802&0.628&0.797&0.577& 0.979 & 0.714 & 0.970 & 0.704 & 1.192 & 0.820 &1.425&0.856\\
ETTm2 &0.293&0.335&0.319&0.353&0.316&0.368&0.463&0.488&0.296&0.343&1.341&0.930&0.332&0.366& 0.447 & 0.487 & 0.987 & 0.755 & 3.369 & 1.439 & 3.977&1.586
\\
% ILI  &3.826&1.423&5.007&1.616&&&&&3.883&1.496&4.572&1.551&3.198&1.285&&&&&&&&\\
ECL  &0.176&0.269&0.323&0.392&0.180&0.280&0.346&0.428&0.180&0.269&0.431&0.478&0.443&0.479& 0.659 & 0.617 & 0.441 & 0.488 & 1.194 & 0.890 &0.965&0.768\\
Traffic  &0.440&0.309&0.951&0.535&0.496&0.371&0.663&0.425&0.430&0.305&0.749&0.446&1.453&0.815& 1.913 & 0.936 & 1.247 & 0.684 & 1.534 & 0.811 &1.550&0.821\\
% ETTm1&\color{red}\textbf{0.028}&\textbf{0.105}&\textbf{0.027} &\color{red}\textbf{0.107}& 0.120& 0.253& 0.104& 0.218& 0.093& 0.206& 0.062& 0.177& 0.036& 0.126&0.051& 0.150&  0.071& 0.188 & 0.055 & 0.166 \\
% ETTm2&\textbf{0.021}&\textbf{0.084}&\color{red}\textbf{0.022}& \color{red}\textbf{0.088}& 0.208& 0.327& 0.046& 0.151& 0.096& 0.208& 0.101& 0.215& 0.026& 0.099 &0.029& 0.105& 0.156& 0.292 & 0.157 & 0.280\\
% ETTh1&\textbf{0.069}&\textbf{0.173}&\color{red}\textbf{0.078}& \color{red}\textbf{0.187}& 0.202& 0.329& 0.284& 0.373& 0.201& 0.306& 0.117& 0.246& 0.094& 0.201 &0.103& 0.214& 0.161& 0.279&0.122& 0.245\\
% ETTh2&\textbf{0.048}&\textbf{0.141}&\color{red}\textbf{0.049}& \color{red}\textbf{0.146}& 0.367& 0.436& 0.119& 0.250& 0.142& 0.259& 0.163& 0.279& 0.053& 0.152& 0.055& 0.156& 0.337& 0.452&0.234& 0.352\\
% Electricity&\textbf{0.090}&\textbf{0.207}&\color{red}\textbf{0.092}& \color{red}\textbf{0.210}& 0.214& 0.339& 0.131& 0.262& 0.132& 0.260& 0.130& 0.259& 0.100& 0.218 &0.101& 0.225& 0.222& 0.328&0.200& 0.313 \\
% Weather&\color{red}\textbf{0.031}&\color{red}\textbf{0.056}&\textbf{0.030}& \textbf{0.054}& 0.076& 0.171& 0.055& 0.117& 0.052& 0.110& 0.099& 0.203& 0.032& 0.059 &\color{red}\textbf{0.031}& 0.057& 0.045& 0.104&0.038& 0.087 \\

\midrule

Average & \textbf{0.371} & \textbf{0.367} & 0.556 & 0.458 & 0.429 & 0.409 & 0.511 & 0.472 & \color{red}\textbf{0.385} & \color{red}\textbf{0.376} & 0.687 & 0.559 & 0.674 & 0.522 & 0.912 & 0.665 & 1.137 & 0.712 & 1.850 & 0.967 & 1.888 & 0.974 \\

\bottomrule

\end{tabular}
}
\end{small}
% \end{center}
\end{table}

\subsection{Zero-shot forecasting} 
This task is used to evaluate the cross datasets adaption ability of our proposed algorithm, i.e. how well a model is able to perform on dataset $A$ (without any training data from $A$) when it is trained from dataset $B$. 

The results are summarized in Table \ref{tab:monash-zero}. The GPT2(6) FPT model consistently outperforms all recent state-of-the-art transformer and MLP-based time series forecasting methods. Compared to recently published state-of-the-art MLP-based method Dlinear, convolution-based method Timesnet, and transformer-based method Patchtst, GPT2(6)FPT demonstrates a relative average metric reduction of \textbf{13.1\%},\textbf{13.6\%} and \textbf{7.3\%}, respectively. Also, the proposed method is comparable to N-BEATS without any meta-learning design and outperforms N-BEATS in the ELECTR dataset. 
We attribute this to the knowledge transfer capability from the FPT model.

\begin{table*}[h]
\vskip -0.10in
\captionsetup{font=small} 
\caption{Zero-shot learning results. Dataset-specific metrics aggregated over each dataset. A lower value indicates better performance. 
% GPT2(6) represents GPT2-backbone (6 Layers) FPT. 
The source dataset of M3, Tourism, Electricity are M4. For M4, the source data for N-BEATS is FRED, and M3 for other models. \textbf{Black}: best, {\color{red} \textbf{Red}}: second best, {\color{violet} \textbf{Violet}}: third best. Appendix \ref{app:zero-shot-full} shows full results.}
\label{tab:monash-zero}
%\vskip 0.1in
\begin{center}
\begin{small}
\scalebox{0.8}{
\begin{tabular}{c|ccccc}
\toprule

Methods & M4 & M3 & TOURISM & ELECTR & \multirow{2}{*}{Average}\\

Metric & sMAPE & sMAPE & MAPE & $ND \times 100$ & \\

\midrule

% DeepAR & 12.25 & 12.67 & 19.27 & 0.765 \\
% N-BEATS & 11.14 & 12.37 & 18.52 & 0.178 \\
% \midrule
N-BEATS & \textbf{11.70} & \textbf{12.44} & \textbf{18.82} & \color{red}\textbf{17.8} & \textbf{15.19} \\
\midrule
DLinear & 15.33 & 14.03 & 28.51 & 17.6 & 18.86\\
% DLinear-M3 & \color{violet} \textbf{15.33} & - & 28.45 & - \\
TimesNet & 13.55 & 14.17 & 28.84 & 19.3 & 18.96 \\
% TimesNet-M3 & 13.55 & - & & - \\
PatchTST & \color{violet}\textbf{13.22} & \color{red}\textbf{13.06} & 27.10 & \color{violet}\textbf{17.3} & \color{violet}\textbf{17.67} \\
ETSformer & 27.74 & 16.03 & 180.40 & 44.2 & 67.09 \\
LightTS & 13.62 & 17.90 & 66.99 & 19.6 & 29.52 \\
Stationary & 13.32 & 15.29 & 43.75 & 22.0 & 23.59 \\
FEDformer & 15.04 & 13.53 & 31.55 & 18.4 & 19.63 \\
Autoformer & 20.02 & 15.87 & 40.39 & 33.9 & 27.54 \\
Informer & 19.04 & 15.82 & 35.82 & 21.2 & 22.97 \\
Reformer & 14.09 & 13.37 & \color{violet}\textbf{25.48} & 21.6 & 18.63 \\
\midrule
GPT2(6) & \color{red}\textbf{13.12} & \color{red}\textbf{13.06} & \color{red}\textbf{22.14} & \textbf{17.2} & \color{red}\textbf{16.38} \\
% GPT2-FPT-M3 & \color{red} \textbf{13.12} & - & 24.99 & - \\

\bottomrule
\end{tabular}
}
\end{small}
\vskip -0.10in
\end{center}
\end{table*}

\section{Ablations}

In this section, we conduct several ablations on model selection and effectiveness of pre-training. The detailed results are shown in  Appendix \ref{appendix:ablations}.
We introduce several variants, GPT2(0) FPT, GPT2(6) without freezing and GPT2(6) without pre-training.

\noindent{\bf Model Selection} We separately analyze the number of GPT2 layers and the fine-tuning parameters selection. The results in Appendix \ref{appendix:ablations} show that GPT2 with 6-layers is a sound choice compared to full or few layers and partially freezing can avoid catastrophic forgetting, enabling fine-tuning without overfitting.

\noindent{\bf Effectiveness of Pre-training}
The results are shown in Table \ref{tab:ablation_main}, GPT2(6) FPT outperforms both GPT2(0) FPT and GPT2-random-initialized, suggesting that GPT2 with pre-training parameters can achieve improvement on times series tasks.
Besides, GPT2(6) FPT performs better than GPT2-unfrozen, demonstrating that partially freezing also helps. Also, results in Appendix \ref{app:nopretrain_freeze} show that random initialized GPT2(6) with freezing performs poorly and the pre-trained knowledge is instrumental for time series tasks.
% In the Appendix \ref{appendix:ablations}, we further conduct an comprehensive study to determine the crucial parameters within our pre-trained language model.
% and a noise mixing experiment to demonstrate the significance of the pre-trained attention weight in Figure \ref{fig:random_ratio}

\begin{table}[!h]
\captionsetup{font=small} 
\caption{Ablation study on 10\% data. All the results are averaged from 4 different prediction lengths. 
% GPT2(6) represents GPT2-backbone (6 Layers) FPT, GPT2(0) represents GPT2-backbone (0 Layer) FPT, 
\textbf{No Freeze} represents GPT2(6) without freezing, \textbf{No Pretrain} represents GPT2(6) without pre-training. \textbf{Black}: best.}
\label{tab:ablation_main}

% \vskip 0.1in
\begin{center}
\begin{small}
\scalebox{1.0}{
\begin{tabular}{c|cc|cc|cc|cc}
\toprule

\multirow{2}{*}{Methods}&\multicolumn{2}{c|}{GPT2(6)}&\multicolumn{2}{c|}{GPT2(0)}&\multicolumn{2}{c|}{No Freeze}&\multicolumn{2}{c}{No Pretrain}\\

&MSE&MAE&MSE&MAE&MSE&MAE&MSE&MAE \\
\midrule

Weather&\textbf{0.237}&\textbf{0.270}&0.263&0.297&0.273&0.302&0.277&0.305\\
% & 96 & \textbf{0} &	\textbf{0.215} &	0.168 &	0.221 &	0.175 &	0.229 \\
% & 192 & \textbf{0.210} &	\textbf{0.254} &	0.238 &	0.286 &	0.244 &	0.287 \\
% & 336 & \textbf{0.256} &	\textbf{0.292} &	0.289 &	0.318 &	0.301 &	0.325 \\
% & 720 & \textbf{0.321} &	\textbf{0.339} &	0.398 &	0.383 &	0.390 &	0.378 \\

ETTh1&\textbf{0.427}&\textbf{0.426}&0.874&0.647&0.753&0.596&1.326&0.743\\
% & 96 & \textbf{0.458} &	\textbf{0.456} &	0.605 &	0.532 &	0.680 &	0.560 \\
% & 192 & \textbf{0.570} &	\textbf{0.516} &	0.713 &	0.579 &	0.738 &	0.602 \\
% & 336 & \textbf{0.608} &	\textbf{0.535} &	0.747 &	0.586 &	0.893 & 0.641 \\
% & 720 & \textbf{0.725} &	\textbf{0.591} &	0.945 &	0.688 &	2.994 & 1.169 \\

% \multirow{4}{*}{\rotatebox{90}{$ETTh2$}}
ETTh2&\textbf{0.346}&\textbf{0.394}&0.666&0.559&0.447&0.451&0.502&0.479\\
% & 96 & \textbf{0.331} &	\textbf{0.374} &	0.369 &	0.394 &	0.422 &	0.433 \\
% & 192 & \textbf{0.402} &	\textbf{0.411} &	0.464 &	0.455 &	0.482 &	0.466 \\
% & 336 & \textbf{0.406} &	\textbf{0.433} &	0.420 &	0.439 &	0.540 &	0.496 \\
% & 720 & \textbf{0.449} &	\textbf{0.464} &	0.535 &	0.515 &	0.564 &	0.519 \\

\bottomrule
\end{tabular}
}
\label{tab:10_percent_ablation}
\end{small}
\end{center}
\end{table}

\section{Exploring Transfer Learning from others: The Unexceptional Nature of GPT2-based-FPT}
We also present experiments on BERT-backbond FPT \cite{Bert/NAACL/Jacob} model and the image-pretrained BEiT-backbone FPT model \cite{bao2022beit} to illustrate the generality of pre-trained models for cross-domain knowledge transferring. The results in Table \ref{tab:other_model_main} demonstrate that the ability of knowledge transfer is not exclusive to GPT2-based pre-trained language models. Subsequently, our theoretical analysis will shed light on the universality of this phenomenon.
% tab ett ablation 5percent

%\input{sections/7_Justification.tex}
\begin{table}[!h]
\captionsetup{font=small} 
\caption{Results of frozen pretrained transformer variants on 5\% ETTh2 and ETTm2. All the results are averaged from 4 different prediction lengths. \textbf{Black}: best. Appendix \ref{app:other_model} shows the full results. }
\label{tab:other_model_main}

% \vskip 0.1in
\begin{center}
\begin{small}
\scalebox{0.9}{
\begin{tabular}{c|cc|cc|cc|cc|cc|cc|cc}
\toprule

\multirow{2}{*}{Methods}&\multicolumn{2}{c|}{GPT2(6)}&\multicolumn{2}{c|}{BERT(6)}&\multicolumn{2}{c|}{BEiT(6)}&\multicolumn{2}{c|}{DLinear}&\multicolumn{2}{c|}{PatchTST}&\multicolumn{2}{c|}{FEDformer}&\multicolumn{2}{c}{Autoformer}\\

&MSE&MAE&MSE&MAE&MSE&MAE&MSE&MAE&MSE&MAE&MSE&MAE&MSE&MAE \\
\midrule

ETTh2&\textbf{0.400}&\textbf{0.433}&0.452&0.451&0.459&0.454&0.827&0.615&0.439&0.448&0.441&0.457&0.470&0.489\\
% & 96 & \textbf{0} &	\textbf{0.215} &	0.168 &	0.221 &	0.175 &	0.229 \\
% & 192 & \textbf{0.210} &	\textbf{0.254} &	0.238 &	0.286 &	0.244 &	0.287 \\
% & 336 & \textbf{0.256} &	\textbf{0.292} &	0.289 &	0.318 &	0.301 &	0.325 \\
% & 720 & \textbf{0.321} &	\textbf{0.339} &	0.398 &	0.383 &	0.390 &	0.378 \\

ETTm2&\textbf{0.308}&\textbf{0.346}&0.318&0.357&0.315&0.357&0.399&0.426&0.314&0.352&0.381&0.404&0.388&0.433\\
% & 96 & \textbf{0.458} &	\textbf{0.456} &	0.605 &	0.532 &	0.680 &	0.560 \\
% & 192 & \textbf{0.570} &	\textbf{0.516} &	0.713 &	0.579 &	0.738 &	0.602 \\
% & 336 & \textbf{0.608} &	\textbf{0.535} &	0.747 &	0.586 &	0.893 & 0.641 \\
% & 720 & \textbf{0.725} &	\textbf{0.591} &	0.945 &	0.688 &	2.994 & 1.169 \\

\bottomrule
\end{tabular}
}
\label{tab:10_percent_ablation}
\end{small}
\end{center}
\end{table}
\section{Training/Inferencing Cost}
\begin{table*}[h]
\vskip -0.1in
\captionsetup{font=small} 
\caption{Training parameters and Training/Inference Cost Comparison}
\label{tab:cost}
%\vskip 0.15in
\begin{center}
\begin{small}
\scalebox{0.9}{
\setlength\tabcolsep{3pt}
% \begin{threeparttable}
\begin{threeparttable}[b]
\begin{tabular}{c|cccc}
\toprule
Model& Training Params& Training Params Percentages& Training Time for 1 step(s) & Inference Time for 1 Batch(s)\\
\midrule
FEDformer-32 &44k &100 &0.889&0.170 \\
TimesNet-32 &2M &100 &0.747&0.302 \\
PatchTST-32&543K &100 &0.043&0.022  \\
\midrule
FEDformer-768&33M &100 &0.208&0.056  \\
TimesNet-768&42M &100 &5.723&2.162 \\
PatchTST-768 &20M &100 &0.457&0.123 \\
GPT-2(3)-768 &4M &6.12 &0.093&0.032 \\
GPT-2(6)-768 &4M &4.6 &0.104&0.054 \\
\bottomrule
\end{tabular}
\end{threeparttable}
}
\end{small}
\end{center}
\vskip -0.2in

\end{table*}
Analysis of computational cost is helpful for investigating the practicality of the LLM-based model. The results can be found in table \ref{tab:cost}. Each baseline model comes in two variants, featuring model hidden dimensions of 32 and 768, which align with GPT-2's specifications. Furthermore, the majority of the baseline models consist of three layers. We assessed the computational cost using a batch from ETTh2 (with a batch size of 128) on a 32G V100 GPU.

The results indicate that GPT-2(3) has substantially enhanced time efficiency and reduced parameter quantity compared to baselines with the same model dimension. This was a surprise since we initially anticipated that this large language model might be slower. However, we surmise that the efficient optimization of huggingface's GPT model implementation primarily accounts for such a significant improvement in time costs. Furthermore, GPT-2(3) and GPT-2(6) demonstrate a mere 6.12\% and 4.60\% proportion of learnable parameters among the overall parameter size, respectively.
\section{Towards Understanding the Universality of Transformer: Connecting Self-Attention with PCA}

 The observation, i.e. we can directly use a trained LM for time series forecasting without having to modify its model, makes us believe that the underlying model is doing something very generic and independent from texts despite it being trained from text data. Our analysis aims to show that part of this generic function can be related to PCA, as minimizing the gradient with respect to the self-attention layer seems to do something similar to PCA. In this section, we take the first step towards revealing the generality of self-attention by connecting the self-attention with principal component analysis (PCA).  Moreover, when coming the question of why fine-tuning is restricted to the embedding layer and layer norm, following our hypothesis that the pre-trained LM as a whole performs something generic, partially fine-tuning any of its components may break the generic function and lead to relatively poor performance for time series analysis.

For each layer, we calculate and perform statistical analysis of the pairwise token similarity values. Specifically, we denote each output feature map with shape of $(b, n, d)$, where $b$ is the batch size, $n$ is the number of tokens, and $d$ is the dimension of each token feature. We calculate the cosine similarity, and the resulting pairwise similarity matrix of shape $(b, n, n)$. Next we count the number of occurrences of similarity values within each interval as a simple statistical analysis.  

\begin{figure}[h]
    \centering
    \includegraphics[width=1\columnwidth]{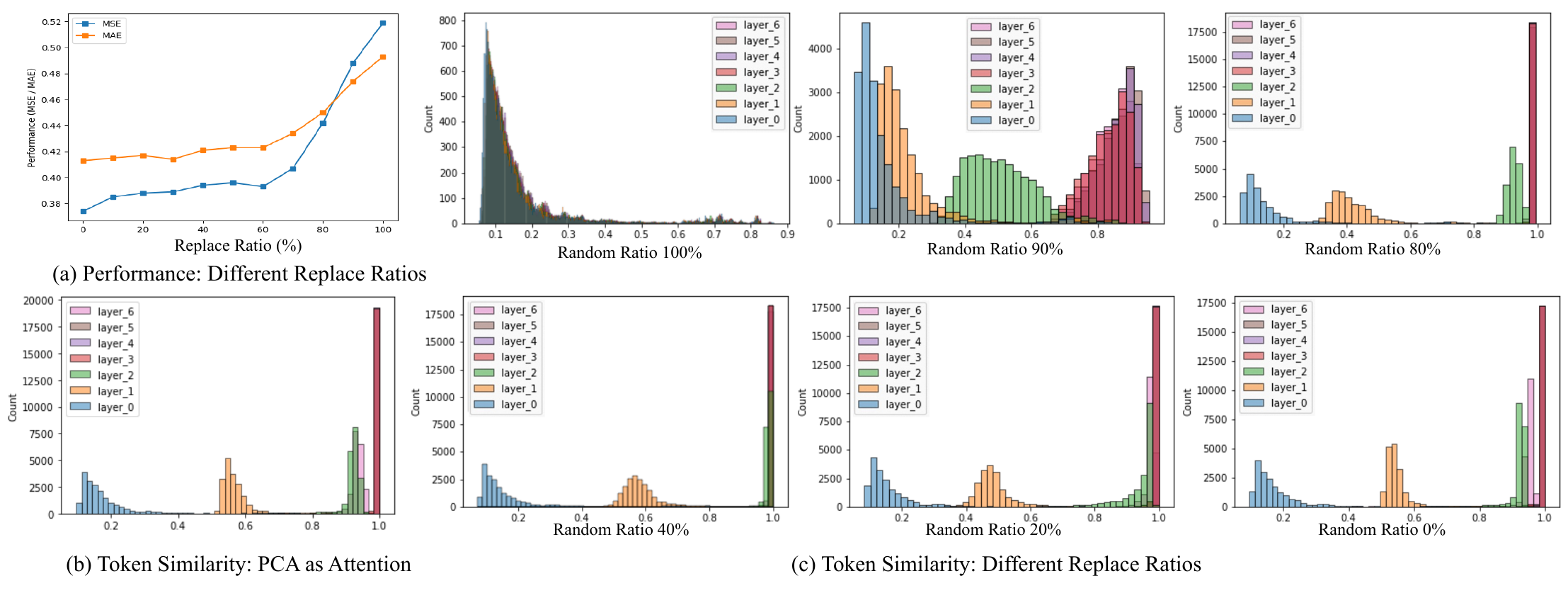}
    \captionsetup{font=small} 
    \caption{(a, c) The performance and token similarity within samples with respect to each layer with different random mixed ratio. Pre-trained parameters are mixed with random initial parameters according to certain proportions. (b) Token similarity within samples when replacing the attention with PCA.}
    \label{fig:random_ratio}
\end{figure}

Our analysis is motivated by the observation that the within-layer token similarity increases with deeper layers in transformer. We report the layer-wise average token cosine similarity on ETTh2 dataset in Figure~\ref{fig:random_ratio} (a, c), where we mix weights from pre-trained LM with weights randomly sampled from Gaussian distribution. Here we summarize our observations: a) in a randomly initialed GPT2 (6) model, the token similarity is low among all layers ($0.1-0.2$); b) when gradually switched to the pretrained GPT2 model, the token similarity significantly increases in the deep layers and eventually reaches more than 0.9 in the last layer. One potential explanation for the increasing token similarity is that all the token vectors are projected into the low-dimensional top eigenvector space of input patterns. To verify this idea, we further conduct experiments where we replace the self-attention module with PCA and find token similarity patterns remain unchanged according to Figure~\ref{fig:random_ratio} (b), which further justifies the potential connection between PCA and self-attention. 

%The ETTh2 dataset contains high volatility hourly information related to the electricity transformer temperature. In this situation, higher token similarity implies the high-frequency noise in the data is eased and only low-frequency information will be reserved. In another word, after going through the pretrained GPT2-FPT model, the signal-noise ratio is enhanced. We use the following theorem to characterize this behavior.

To build the theoretical connection between PCA and self-attention, we first analyze the gradient structure of self-attention. Let $X = (x_1, \ldots, x_N)^{\top} \in \R^{N\times D}$ be the input pattern, and let $f(X) = (f_1(X), \ldots, f_N(x))^{\top}:\R^{N\times D}\mapsto \R^{N\times D}$ be the function for self-attention, i.e., 
\scalebox{1.0}{
$\begin{array}{rcl}
f_i(X) = \mbox{softmax}(XAX^{\top})X
\end{array}$ 
}
where $A = W_QW_K^{\top} \in \R^{D\times D}$. 

\begin{lemma}\label{lemma:jacobian}
Let the Jacobian $J = \left[\frac{\partial f_i(X)}{\partial x_j}\right]_{i,j=1}^N$ represent the gradient $f(X)$ w.r.t the input pattern, then we have
% The lemma below shows an important structure of $J$. 
%\begin{lemma}
\scalebox{1.0}{
$\begin{array}{rcl}
|J|_2 \leq |A|_2\sum_{i=1}^N \left(P_{i,i} + \frac{1}{2}\right)\left|x_i - \sum_{j=1}^N P_{i,j}x_j\right|^2 + \Delta
\end{array}$ 
}
where
\scalebox{1.0}{
$\begin{array}{rcl}
\Delta = |A|_2 \sum_{i \neq j}^N P_{i,j}\left|x_j - \sum_{k = 1}^N P_{i,k}x_k\right|^2 + \frac{|A|_2}{2}\sum_{j=1}^N |x_i|^2
\end{array}$ 
}
and
\scalebox{1.0}{
$\begin{array}{rcl}
P_{i,j} = \frac{\exp(x_i^{\top}Ax_j)}{\sum_{k=1}^N \exp(x_i^{\top}Ax_k)}
\end{array}$.
}
\end{lemma}

This lemma reveals an important gradient structure of $J$. The proof of essentially follows the analysis in~\cite{kim2021lipschitz}, and we include it in Appendix~\ref{appendix:pca} for completeness. 

Using the gradient structure revealed in Lemma \ref{lemma:jacobian}, we can connect self-attention with PCA. In order to minimize the norm of gradient $|J|_2$, we essentially need to make $\sum_{i=1}^N |x_i - \sum_{j=1}^N P_{i,j}x_j|^2$ small. When $A$ is small and all the input patterns are centered at $0$ (i.e. $\sum_{i=1}^N x_i = 0$), we have $\sum_{i=1}^N|x_i - X^{\top}P_{i,:}|^2 \approx \sum_{i=1}^N |x_i - X^{\top}XAx_i|^2$. 

The theorem below shows that $A$ minimizing the objective $\sum_{i=1}^N |x_i - X^{\top}XAx_i|^2$ contains the largest $m$ eigenvectors of $X^{\top}X$ where $m$ is the rank of $A$.
\begin{thm}\label{thm:eigen}
Let $W_Q$ and $W_K$ be matrices of size $D\times m$. Let $\lambda_1 \geq \lambda_2\geq ...\geq \lambda_D$ be the eigenvalues of $X^{\top}X$ ranked in descending order, and let $v_i \in \R^D, i=1, \ldots, D$ be the corresponding eigenvectors. The optimal solution $A^*$ that minimizes $\sum_{i=1}^N |x_i - X^{\top}XAx_i|^2$ is given by
$A = \sum_{i=1}^m \frac{1}{\lambda_i}v_iv_i^{\top}$. 
\end{thm}
The proof of Theorem 1 can be found in Appendix~\ref{appendix:pca}. Following Theorem \ref{thm:eigen}, through the training of pushing gradient to zero, self-attention learns to perform a function closely related to PCA.

\section{Conclusions}
In this paper, we developed a foundation model for time series analysis, based on pre-trained model from NLP or CV, that can (a) facilitate the model training for downstream tasks, and (b) provide unified framework for diverse time series analysis tasks. Our empirical studies show that the proposed method performs on par or better than the state-of-the-art approaches on almost all time series tasks. We also examine the universality of transformer by connecting self-attention with PCA, an important step towards understanding how generative models work in practice. On the other hand, we do recognize some limitations of our work: the zero-shot performance of our approach is still behind N-beat on several datasets, and our analysis of the generality of transformer is still in the early stage. Moving forward, we plan to improve the performance of our approach by exploiting the parameter efficient fine-tuning approaches which usually introduce additional structures into the pre-trained model for better adaption. To better understand the universality of transformer, we also plan to examine it from the viewpoint of n-gram language model, an approach that is taken by~\cite{elhage2021mathematical,olsson2022context}. In Appendix~\ref{appendix:ngram}, we include our initial analysis along this direction. 

%In this paper, we have demonstrated that utilizing a pretrained language model can significantly enhance time series forecasting in few-shot learning, full data learning, and zero-shot learning scenarios. While transfer learning has been extensively studied in other domains, there has been limited work on transfer learning for time series data, primarily due to the complexity and diversity of such data across various domains. To the best of our knowledge, our work represents the first successful attempt to perform cross-domain knowledge transfer for time series forecasting, achieving state-of-the-art performance in various forecasting tasks. Our findings are supported by empirical and theoretical evidence.Moving forward, we aim to investigate more time series applications where cross-domain knowledge transfer can be beneficial, and further understand the underlying mechanism for cross-domain knowledge transfer in the context of time series data.
\section*{Acknowledgement}
We would like to express our sincere gratitude to Ziqing Ma, Qingsong Wen, Mengni Ye, and Tao Yao for their valuable suggestions and proofreading assistance throughout the development of this paper. Their insightful feedback and attention to detail greatly improved the quality and clarity of our work.

% % tab ett 10percent
% \input{tables/tab_ett_10percent.tex}
% % tab ett 5percent
% \input{tables/tab_ett_5percent.tex}
% % tab ett ablation 10percent
% \input{tables/tab_ett_ablation_10percent.tex}
% % tab ett ablation 5percent
% \input{tables/tab_ett_ablation_5percent.tex}

\bibliography{main}

\begin{thebibliography}{66}
\providecommand{\natexlab}[1]{#1}
\providecommand{\url}[1]{\texttt{#1}}
\expandafter\ifx\csname urlstyle\endcsname\relax
  \providecommand{\doi}[1]{doi: #1}\else
  \providecommand{\doi}{doi: \begingroup \urlstyle{rm}\Url}\fi

\bibitem[Abdulaal et~al.(2021)Abdulaal, Liu, and Lancewicki]{PSM}
Abdulaal, A., Liu, Z., and Lancewicki, T.
\newblock Practical approach to asynchronous multivariate time series anomaly
  detection and localization.
\newblock In \emph{Proceedings of the 27th ACM SIGKDD conference on knowledge
  discovery \& data mining}, pp.\  2485--2494, 2021.

\bibitem[Bagnall et~al.(2018)Bagnall, Dau, Lines, Flynn, Large, Bostrom,
  Southam, and Keogh]{UEA}
Bagnall, A., Dau, H.~A., Lines, J., Flynn, M., Large, J., Bostrom, A., Southam,
  P., and Keogh, E.
\newblock The uea multivariate time series classification archive, 2018.
\newblock \emph{arXiv preprint arXiv:1811.00075}, 2018.

\bibitem[Bao et~al.(2021)Bao, Wang, Dong, Liu, Mohammed, Aggarwal, Som, and
  Wei]{bao2021vlmo}
Bao, H., Wang, W., Dong, L., Liu, Q., Mohammed, O.~K., Aggarwal, K., Som, S.,
  and Wei, F.
\newblock Vlmo: Unified vision-language pre-training with
  mixture-of-modality-experts.
\newblock \emph{arXiv preprint arXiv:2111.02358}, 2021.

\bibitem[Bao et~al.(2022)Bao, Dong, Piao, and Wei]{bao2022beit}
Bao, H., Dong, L., Piao, S., and Wei, F.
\newblock {BE}it: {BERT} pre-training of image transformers.
\newblock In \emph{International Conference on Learning Representations}, 2022.

\bibitem[B{\"o}se et~al.(2017)B{\"o}se, Flunkert, Gasthaus, Januschowski,
  Lange, Salinas, Schelter, Seeger, and Wang]{bose2017probabilistic}
B{\"o}se, J.-H., Flunkert, V., Gasthaus, J., Januschowski, T., Lange, D.,
  Salinas, D., Schelter, S., Seeger, M., and Wang, Y.
\newblock Probabilistic demand forecasting at scale.
\newblock \emph{Proceedings of the VLDB Endowment}, 10\penalty0 (12):\penalty0
  1694--1705, 2017.

\bibitem[Box \& Jenkins(1968)Box and Jenkins]{arima_1}
Box, G.~E. and Jenkins, G.~M.
\newblock Some recent advances in forecasting and control.
\newblock \emph{Journal of the Royal Statistical Society. Series C (Applied
  Statistics)}, 17\penalty0 (2):\penalty0 91--109, 1968.

\bibitem[Box \& Pierce(1970)Box and Pierce]{arima_2}
Box, G.~E. and Pierce, D.~A.
\newblock Distribution of residual autocorrelations in
  autoregressive-integrated moving average time series models.
\newblock \emph{Journal of the American statistical Association}, 65\penalty0
  (332):\penalty0 1509--1526, 1970.

\bibitem[Brown et~al.(2020)Brown, Mann, Ryder, Subbiah, Kaplan, Dhariwal,
  Neelakantan, Shyam, Sastry, Askell, Agarwal, Herbert-Voss, Krueger, Henighan,
  Child, Ramesh, Ziegler, Wu, Winter, Hesse, Chen, Sigler, Litwin, Gray, Chess,
  Clark, Berner, McCandlish, Radford, Sutskever, and
  Amodei]{Brown2020LanguageMA}
Brown, T.~B., Mann, B., Ryder, N., Subbiah, M., Kaplan, J., Dhariwal, P.,
  Neelakantan, A., Shyam, P., Sastry, G., Askell, A., Agarwal, S.,
  Herbert-Voss, A., Krueger, G., Henighan, T.~J., Child, R., Ramesh, A.,
  Ziegler, D.~M., Wu, J., Winter, C., Hesse, C., Chen, M., Sigler, E., Litwin,
  M., Gray, S., Chess, B., Clark, J., Berner, C., McCandlish, S., Radford, A.,
  Sutskever, I., and Amodei, D.
\newblock Language models are few-shot learners.
\newblock \emph{ArXiv}, abs/2005.14165, 2020.

\bibitem[Challu et~al.(2022)Challu, Olivares, Oreshkin, Garza, Mergenthaler,
  and Dubrawski]{nhits}
Challu, C., Olivares, K.~G., Oreshkin, B.~N., Garza, F., Mergenthaler, M., and
  Dubrawski, A.
\newblock N-hits: Neural hierarchical interpolation for time series
  forecasting.
\newblock \emph{arXiv preprint arXiv:2201.12886}, 2022.

\bibitem[Chen \& Guestrin(2016)Chen and Guestrin]{xgboost}
Chen, T. and Guestrin, C.
\newblock Xgboost: A scalable tree boosting system.
\newblock KDD '16, 2016.

\bibitem[Chung et~al.(2014)Chung, Gulcehre, Cho, and
  Bengio]{chung2014empirical}
Chung, J., Gulcehre, C., Cho, K., and Bengio, Y.
\newblock Empirical evaluation of gated recurrent neural networks on sequence
  modeling.
\newblock \emph{arXiv preprint arXiv:1412.3555}, 2014.

\bibitem[Courty \& Li(1999)Courty and Li]{courty1999timing}
Courty, P. and Li, H.
\newblock Timing of seasonal sales.
\newblock \emph{The Journal of Business}, 72\penalty0 (4):\penalty0 545--572,
  1999.

\bibitem[Dempster et~al.(2020)Dempster, Petitjean, and Webb]{ROCKET}
Dempster, A., Petitjean, F., and Webb, G.~I.
\newblock {ROCKET}: Exceptionally fast and accurate time series classification
  using random convolutional kernels.
\newblock \emph{Data Mining and Knowledge Discovery}, 34\penalty0 (5):\penalty0
  1454--1495, 2020.

\bibitem[Devlin et~al.(2019)Devlin, Chang, Lee, and
  Toutanova]{Bert/NAACL/Jacob}
Devlin, J., Chang, M., Lee, K., and Toutanova, K.
\newblock {BERT:} pre-training of deep bidirectional transformers for language
  understanding.
\newblock In \emph{Proceedings of the 2019 Conference of the North American
  Chapter of the Association for Computational Linguistics: Human Language
  Technologies (NAACL-HLT), Minneapolis, MN, USA, June 2-7, 2019}, pp.\
  4171--4186, 2019.

\bibitem[Dosovitskiy et~al.(2021)Dosovitskiy, Beyer, Kolesnikov, Weissenborn,
  Zhai, Unterthiner, Dehghani, Minderer, Heigold, Gelly, Uszkoreit, and
  Houlsby]{Transformers-for-image-at-scale/iclr/DosovitskiyB0WZ21}
Dosovitskiy, A., Beyer, L., Kolesnikov, A., Weissenborn, D., Zhai, X.,
  Unterthiner, T., Dehghani, M., Minderer, M., Heigold, G., Gelly, S.,
  Uszkoreit, J., and Houlsby, N.
\newblock An image is worth 16x16 words: Transformers for image recognition at
  scale.
\newblock In \emph{9th International Conference on Learning Representations
  (ICLR), Austria, May 3-7, 2021}, 2021.

\bibitem[Elhage et~al.(2021)Elhage, Nanda, Olsson, Henighan, Joseph, Mann,
  Askell, Bai, Chen, Conerly, DasSarma, Drain, Ganguli, Hatfield-Dodds,
  Hernandez, Jones, Kernion, Lovitt, Ndousse, Amodei, Brown, Clark, Kaplan,
  McCandlish, and Olah]{elhage2021mathematical}
Elhage, N., Nanda, N., Olsson, C., Henighan, T., Joseph, N., Mann, B., Askell,
  A., Bai, Y., Chen, A., Conerly, T., DasSarma, N., Drain, D., Ganguli, D.,
  Hatfield-Dodds, Z., Hernandez, D., Jones, A., Kernion, J., Lovitt, L.,
  Ndousse, K., Amodei, D., Brown, T., Clark, J., Kaplan, J., McCandlish, S.,
  and Olah, C.
\newblock A mathematical framework for transformer circuits.
\newblock \emph{Transformer Circuits Thread}, 2021.
\newblock https://transformer-circuits.pub/2021/framework/index.html.

\bibitem[Franceschi et~al.(2019)Franceschi, Dieuleveut, and Jaggi]{tcn}
Franceschi, J.-Y., Dieuleveut, A., and Jaggi, M.
\newblock Unsupervised scalable representation learning for multivariate time
  series.
\newblock \emph{Advances in neural information processing systems}, 32, 2019.

\bibitem[Friedman(1962)]{Friedman1962}
Friedman, M.
\newblock The interpolation of time series by related series.
\newblock \emph{J. Amer. Statist. Assoc}, 1962.

\bibitem[Gao et~al.(2020)Gao, Song, Wen, Wang, Sun, and Xu]{gao2020robusttad}
Gao, J., Song, X., Wen, Q., Wang, P., Sun, L., and Xu, H.
\newblock {RobustTAD}: Robust time series anomaly detection via decomposition
  and convolutional neural networks.
\newblock \emph{KDD Workshop on Mining and Learning from Time Series
  (KDD-MileTS'20)}, 2020.

\bibitem[{Giannou} et~al.(2023){Giannou}, {Rajput}, {Sohn}, {Lee}, {Lee}, and
  {Papailiopoulos}]{looped2023}
{Giannou}, A., {Rajput}, S., {Sohn}, J.-y., {Lee}, K., {Lee}, J.~D., and
  {Papailiopoulos}, D.
\newblock {Looped Transformers as Programmable Computers}.
\newblock \emph{arXiv e-prints}, art. arXiv:2301.13196, January 2023.
\newblock \doi{10.48550/arXiv.2301.13196}.

\bibitem[Godahewa et~al.(2021)Godahewa, Bergmeir, Webb, Hyndman, and
  Montero-Manso]{godahewa2021monash}
Godahewa, R., Bergmeir, C., Webb, G.~I., Hyndman, R.~J., and Montero-Manso, P.
\newblock Monash time series forecasting archive.
\newblock In \emph{Neural Information Processing Systems Track on Datasets and
  Benchmarks}, 2021.

\bibitem[Gu et~al.(2021)Gu, Goel, and R{\'e}]{lssl}
Gu, A., Goel, K., and R{\'e}, C.
\newblock Efficiently modeling long sequences with structured state spaces.
\newblock \emph{arXiv preprint arXiv:2111.00396}, 2021.

\bibitem[Hochreiter \& Schmidhuber(1997)Hochreiter and
  Schmidhuber]{hochreiter1997long}
Hochreiter, S. and Schmidhuber, J.
\newblock Long short-term memory.
\newblock \emph{Neural computation}, 9\penalty0 (8):\penalty0 1735--1780, 1997.

\bibitem[Houlsby et~al.(2019)Houlsby, Giurgiu, Jastrzebski, Morrone,
  De~Laroussilhe, Gesmundo, Attariyan, and Gelly]{houlsby2019parameter}
Houlsby, N., Giurgiu, A., Jastrzebski, S., Morrone, B., De~Laroussilhe, Q.,
  Gesmundo, A., Attariyan, M., and Gelly, S.
\newblock Parameter-efficient transfer learning for nlp.
\newblock In \emph{International Conference on Machine Learning}, pp.\
  2790--2799. PMLR, 2019.

\bibitem[Huang et~al.(2022)Huang, Shi, Zhang, Wang, Cheung, Qin, Dai, and
  Li]{huang2022flowformer}
Huang, Z., Shi, X., Zhang, C., Wang, Q., Cheung, K.~C., Qin, H., Dai, J., and
  Li, H.
\newblock Flowformer: A transformer architecture for optical flow.
\newblock In \emph{Computer Vision--ECCV 2022: 17th European Conference, Tel
  Aviv, Israel, October 23--27, 2022, Proceedings, Part XVII}, pp.\  668--685.
  Springer, 2022.

\bibitem[Hundman et~al.(2018)Hundman, Constantinou, Laporte, Colwell, and
  Soderstrom]{MSL_SMAP}
Hundman, K., Constantinou, V., Laporte, C., Colwell, I., and Soderstrom, T.
\newblock Detecting spacecraft anomalies using lstms and nonparametric dynamic
  thresholding.
\newblock In \emph{Proceedings of the 24th ACM SIGKDD international conference
  on knowledge discovery \& data mining}, pp.\  387--395, 2018.

\bibitem[Hyndman \& Athanasopoulos(2021)Hyndman and
  Athanasopoulos]{hyndman:forecasting:3rd}
Hyndman, R. and Athanasopoulos, G.
\newblock \emph{Forecasting: Principles and Practice}.
\newblock OTexts, Australia, 3rd edition, 2021.

\bibitem[Ismail~Fawaz et~al.(2019)Ismail~Fawaz, Forestier, Weber, Idoumghar,
  and Muller]{IsmailFawaz2018deep}
Ismail~Fawaz, H., Forestier, G., Weber, J., Idoumghar, L., and Muller, P.-A.
\newblock Deep learning for time series classification: a review.
\newblock \emph{Data Mining and Knowledge Discovery}, 33\penalty0 (4):\penalty0
  917--963, 2019.

\bibitem[Kim et~al.(2021)Kim, Papamakarios, and Mnih]{kim2021lipschitz}
Kim, H., Papamakarios, G., and Mnih, A.
\newblock The lipschitz constant of self-attention.
\newblock In \emph{International Conference on Machine Learning}, pp.\
  5562--5571. PMLR, 2021.

\bibitem[Kim et~al.(2022)Kim, Kim, Tae, Park, Choi, and
  Choo]{kim2022reversible}
Kim, T., Kim, J., Tae, Y., Park, C., Choi, J.-H., and Choo, J.
\newblock Reversible instance normalization for accurate time-series
  forecasting against distribution shift.
\newblock In \emph{International Conference on Learning Representations}, 2022.

\bibitem[Kitaev et~al.(2020)Kitaev, Kaiser, and Levskaya]{reformer}
Kitaev, N., Kaiser, L., and Levskaya, A.
\newblock Reformer: The efficient transformer.
\newblock \emph{arXiv preprint arXiv:2001.04451}, 2020.

\bibitem[Lacoste-Julien et~al.(2012)Lacoste-Julien, Schmidt, and
  Bach]{lacoste2012simpler}
Lacoste-Julien, S., Schmidt, M., and Bach, F.
\newblock A simpler approach to obtaining an o (1/t) convergence rate for the
  projected stochastic subgradient method.
\newblock \emph{arXiv preprint arXiv:1212.2002}, 2012.

\bibitem[Lai et~al.(2018)Lai, Chang, Yang, and Liu]{lstnet}
Lai, G., Chang, W.-C., Yang, Y., and Liu, H.
\newblock Modeling long-and short-term temporal patterns with deep neural
  networks.
\newblock In \emph{The 41st international ACM SIGIR conference on research \&
  development in information retrieval}, pp.\  95--104, 2018.

\bibitem[Lim et~al.(2021)Lim, Ar{\i}k, Loeff, and Pfister]{lim2021temporal}
Lim, B., Ar{\i}k, S.~{\"O}., Loeff, N., and Pfister, T.
\newblock Temporal fusion transformers for interpretable multi-horizon time
  series forecasting.
\newblock \emph{International Journal of Forecasting}, 2021.

\bibitem[Liu et~al.(2021)Liu, Yu, Liao, Li, Lin, Liu, and Dustdar]{pyraformer}
Liu, S., Yu, H., Liao, C., Li, J., Lin, W., Liu, A.~X., and Dustdar, S.
\newblock Pyraformer: Low-complexity pyramidal attention for long-range time
  series modeling and forecasting.
\newblock In \emph{International conference on learning representations}, 2021.

\bibitem[Liu et~al.(2022)Liu, Wu, Wang, and Long]{non-stationary}
Liu, Y., Wu, H., Wang, J., and Long, M.
\newblock Non-stationary transformers: Exploring the stationarity in time
  series forecasting.
\newblock In \emph{Advances in Neural Information Processing Systems}, 2022.

\bibitem[Lu et~al.(2022)Lu, Grover, Abbeel, and
  Mordatch]{pretrained_transformer}
Lu, K., Grover, A., Abbeel, P., and Mordatch, I.
\newblock Frozen pretrained transformers as universal computation engines.
\newblock \emph{Proceedings of the AAAI Conference on Artificial Intelligence},
  36\penalty0 (7):\penalty0 7628--7636, Jun. 2022.

\bibitem[Makridakis et~al.(2018)Makridakis, Spiliotis, and
  Assimakopoulos]{makridakis2018m4}
Makridakis, S., Spiliotis, E., and Assimakopoulos, V.
\newblock The m4 competition: Results, findings, conclusion and way forward.
\newblock \emph{International Journal of Forecasting}, 34\penalty0
  (4):\penalty0 802--808, 2018.

\bibitem[Mathur \& Tippenhauer(2016)Mathur and Tippenhauer]{SWaT}
Mathur, A.~P. and Tippenhauer, N.~O.
\newblock Swat: A water treatment testbed for research and training on ics
  security.
\newblock In \emph{2016 international workshop on cyber-physical systems for
  smart water networks (CySWater)}, pp.\  31--36. IEEE, 2016.

\bibitem[Nie et~al.(2022)Nie, Nguyen, Sinthong, and Kalagnanam]{Patchformer}
Nie, Y., Nguyen, N.~H., Sinthong, P., and Kalagnanam, J.
\newblock A time series is worth 64 words: Long-term forecasting with
  transformers.
\newblock \emph{ArXiv}, abs/2211.14730, 2022.

\bibitem[Olsson et~al.(2022)Olsson, Elhage, Nanda, Joseph, DasSarma, Henighan,
  Mann, Askell, Bai, Chen, Conerly, Drain, Ganguli, Hatfield-Dodds, Hernandez,
  Johnston, Jones, Kernion, Lovitt, Ndousse, Amodei, Brown, Clark, Kaplan,
  McCandlish, and Olah]{olsson2022context}
Olsson, C., Elhage, N., Nanda, N., Joseph, N., DasSarma, N., Henighan, T.,
  Mann, B., Askell, A., Bai, Y., Chen, A., Conerly, T., Drain, D., Ganguli, D.,
  Hatfield-Dodds, Z., Hernandez, D., Johnston, S., Jones, A., Kernion, J.,
  Lovitt, L., Ndousse, K., Amodei, D., Brown, T., Clark, J., Kaplan, J.,
  McCandlish, S., and Olah, C.
\newblock In-context learning and induction heads.
\newblock \emph{Transformer Circuits Thread}, 2022.
\newblock
  https://transformer-circuits.pub/2022/in-context-learning-and-induction-heads/index.html.

\bibitem[OpenAI(2023)]{OpenAI2023GPT4TR}
OpenAI.
\newblock Gpt-4 technical report.
\newblock \emph{ArXiv}, abs/2303.08774, 2023.

\bibitem[Oreshkin et~al.(2019)Oreshkin, Carpov, Chapados, and Bengio]{n-beats}
Oreshkin, B.~N., Carpov, D., Chapados, N., and Bengio, Y.
\newblock N-beats: Neural basis expansion analysis for interpretable time
  series forecasting.
\newblock \emph{arXiv preprint arXiv:1905.10437}, 2019.

\bibitem[Oreshkin et~al.(2021)Oreshkin, Carpov, Chapados, and
  Bengio]{oreshkin2021meta}
Oreshkin, B.~N., Carpov, D., Chapados, N., and Bengio, Y.
\newblock Meta-learning framework with applications to zero-shot time-series
  forecasting.
\newblock In \emph{Proceedings of the AAAI Conference on Artificial
  Intelligence}, number~10, pp.\  9242--9250, 2021.

\bibitem[Radford \& Narasimhan(2018)Radford and
  Narasimhan]{Radford2018ImprovingLU}
Radford, A. and Narasimhan, K.
\newblock Improving language understanding by generative pre-training.
\newblock 2018.

\bibitem[Radford et~al.(2019)Radford, Wu, Child, Luan, Amodei, and
  Sutskever]{gpt2-2019}
Radford, A., Wu, J., Child, R., Luan, D., Amodei, D., and Sutskever, I.
\newblock Language models are unsupervised multitask learners.
\newblock 2019.

\bibitem[Rao et~al.(2021)Rao, Zhao, Zhu, Lu, and
  Zhou]{DBLP:Global-filter-FNO-in-cv}
Rao, Y., Zhao, W., Zhu, Z., Lu, J., and Zhou, J.
\newblock Global filter networks for image classification.
\newblock \emph{Advances in Neural Information Processing Systems (NeurIPS)},
  34, 2021.

\bibitem[Su et~al.(2019)Su, Zhao, Niu, Liu, Sun, and Pei]{SMD}
Su, Y., Zhao, Y., Niu, C., Liu, R., Sun, W., and Pei, D.
\newblock Robust anomaly detection for multivariate time series through
  stochastic recurrent neural network.
\newblock In \emph{Proceedings of the 25th ACM SIGKDD international conference
  on knowledge discovery \& data mining}, pp.\  2828--2837, 2019.

\bibitem[Touvron et~al.(2021)Touvron, Cord, Douze, Massa, Sablayrolles, and
  J{\'e}gou]{deit}
Touvron, H., Cord, M., Douze, M., Massa, F., Sablayrolles, A., and J{\'e}gou,
  H.
\newblock Training data-efficient image transformers \& distillation through
  attention.
\newblock In \emph{International Conference on Machine Learning}, pp.\
  10347--10357. PMLR, 2021.

\bibitem[Vardi et~al.(2021)Vardi, Yehudai, and Shamir]{vardi2021optimal}
Vardi, G., Yehudai, G., and Shamir, O.
\newblock On the optimal memorization power of relu neural networks.
\newblock \emph{arXiv preprint arXiv:2110.03187}, 2021.

\bibitem[Vaswani et~al.(2017)Vaswani, Shazeer, Parmar, Uszkoreit, Jones, Gomez,
  Kaiser，Lukasz, and Polosukhin]{vaswani2017attention}
Vaswani, A., Shazeer, N., Parmar, N., Uszkoreit, J., Jones, L., Gomez, A.~N.,
  Kaiser，Lukasz, and Polosukhin, I.
\newblock Attention is all you need.
\newblock \emph{arXiv preprint arXiv:1706.03762}, 2017.

\bibitem[Wang et~al.(2022)Wang, Wang, Wang, Lin, Chang, Li, and
  Jin]{wang2022kvt}
Wang, P., Wang, X., Wang, F., Lin, M., Chang, S., Li, H., and Jin, R.
\newblock Kvt: k-nn attention for boosting vision transformers.
\newblock In \emph{Computer Vision--ECCV 2022: 17th European Conference, Tel
  Aviv, Israel, October 23--27, 2022, Proceedings, Part XXIV}, pp.\  285--302.
  Springer, 2022.

\bibitem[Wang \& Isola(2020)Wang and Isola]{Wang2020UnderstandingCR}
Wang, T. and Isola, P.
\newblock Understanding contrastive representation learning through alignment
  and uniformity on the hypersphere.
\newblock \emph{ArXiv}, abs/2005.10242, 2020.

\bibitem[Wen et~al.(2022)Wen, Yang, Zhou, and Sun]{wen2022robust}
Wen, Q., Yang, L., Zhou, T., and Sun, L.
\newblock Robust time series analysis and applications: An industrial
  perspective.
\newblock In \emph{Proceedings of the 28th ACM SIGKDD Conference on Knowledge
  Discovery and Data Mining}, pp.\  4836--4837, 2022.

\bibitem[Wen et~al.(2023)Wen, Zhou, Zhang, Chen, Ma, Yan, and
  Sun]{wen2022transformers}
Wen, Q., Zhou, T., Zhang, C., Chen, W., Ma, Z., Yan, J., and Sun, L.
\newblock Transformers in time series: A survey.
\newblock In \emph{International Joint Conference on Artificial
  Intelligence(IJCAI)}, 2023.

\bibitem[Wolf et~al.(2020)Wolf, Debut, Sanh, Chaumond, Delangue, Moi, Cistac,
  Rault, Louf, Funtowicz, Davison, Shleifer, von Platen, Ma, Jernite, Plu, Xu,
  Scao, Gugger, Drame, Lhoest, and Rush]{huggingface}
Wolf, T., Debut, L., Sanh, V., Chaumond, J., Delangue, C., Moi, A., Cistac, P.,
  Rault, T., Louf, R., Funtowicz, M., Davison, J., Shleifer, S., von Platen,
  P., Ma, C., Jernite, Y., Plu, J., Xu, C., Scao, T.~L., Gugger, S., Drame, M.,
  Lhoest, Q., and Rush, A.~M.
\newblock Transformers: State-of-the-art natural language processing.
\newblock In \emph{Proceedings of the 2020 Conference on Empirical Methods in
  Natural Language Processing: System Demonstrations}, pp.\  38--45, Online,
  October 2020. Association for Computational Linguistics.
\newblock URL \url{https://www.aclweb.org/anthology/2020.emnlp-demos.6}.

\bibitem[Woo et~al.(2022)Woo, Liu, Sahoo, Kumar, and Hoi]{woo2022etsformer}
Woo, G., Liu, C., Sahoo, D., Kumar, A., and Hoi, S.
\newblock Etsformer: Exponential smoothing transformers for time-series
  forecasting.
\newblock \emph{arXiv preprint arXiv:2202.01381}, 2022.

\bibitem[Wu et~al.(2021)Wu, Xu, Wang, and Long]{wu2021autoformer}
Wu, H., Xu, J., Wang, J., and Long, M.
\newblock Autoformer: Decomposition transformers with auto-correlation for
  long-term series forecasting.
\newblock In \emph{Advances in Neural Information Processing Systems
  (NeurIPS)}, pp.\  101--112, 2021.

\bibitem[Wu et~al.(2023)Wu, Hu, Liu, Zhou, Wang, and Long]{timesnet}
Wu, H., Hu, T., Liu, Y., Zhou, H., Wang, J., and Long, M.
\newblock Timesnet: Temporal 2d-variation modeling for general time series
  analysis.
\newblock In \emph{The Eleventh International Conference on Learning
  Representations}, 2023.
\newblock URL \url{https://openreview.net/forum?id=ju_Uqw384Oq}.

\bibitem[Xu et~al.(2021)Xu, Wu, Wang, and Long]{xu2021anomaly}
Xu, J., Wu, H., Wang, J., and Long, M.
\newblock Anomaly transformer: Time series anomaly detection with association
  discrepancy.
\newblock \emph{arXiv preprint arXiv:2110.02642}, 2021.

\bibitem[Yang et~al.(2021)Yang, Tsai, and Chen]{yang2021voice2series}
Yang, C.-H.~H., Tsai, Y.-Y., and Chen, P.-Y.
\newblock Voice2series: Reprogramming acoustic models for time series
  classification.
\newblock In \emph{International Conference on Machine Learning}, pp.\
  11808--11819, 2021.

\bibitem[Yun et~al.(2020)Yun, Chang, Bhojanapalli, Rawat, Reddi, and
  Kumar]{yun2020n}
Yun, C., Chang, Y.-W., Bhojanapalli, S., Rawat, A.~S., Reddi, S., and Kumar, S.
\newblock O (n) connections are expressive enough: Universal approximability of
  sparse transformers.
\newblock \emph{Advances in Neural Information Processing Systems},
  33:\penalty0 13783--13794, 2020.

\bibitem[Zeng et~al.(2023)Zeng, Chen, Zhang, and Xu]{dlinear}
Zeng, A., Chen, M., Zhang, L., and Xu, Q.
\newblock Are transformers effective for time series forecasting?
\newblock 2023.

\bibitem[Zhang et~al.(2022)Zhang, Zhang, Cao, Bian, Yi, Zheng, and Li]{lightts}
Zhang, T., Zhang, Y., Cao, W., Bian, J., Yi, X., Zheng, S., and Li, J.
\newblock Less is more: Fast multivariate time series forecasting with light
  sampling-oriented mlp structures.
\newblock \emph{arXiv preprint arXiv:2207.01186}, 2022.

\bibitem[Zhou et~al.(2021)Zhou, Zhang, Peng, Zhang, Li, Xiong, and
  Zhang]{zhou2021informer}
Zhou, H., Zhang, S., Peng, J., Zhang, S., Li, J., Xiong, H., and Zhang, W.
\newblock Informer: Beyond efficient transformer for long sequence time-series
  forecasting.
\newblock In \emph{Proceedings of AAAI}, 2021.

\bibitem[Zhou et~al.(2022)Zhou, Ma, Wen, Wang, Sun, and Jin]{zhou2022fedformer}
Zhou, T., Ma, Z., Wen, Q., Wang, X., Sun, L., and Jin, R.
\newblock {FEDformer}: Frequency enhanced decomposed transformer for long-term
  series forecasting.
\newblock In \emph{Proc. 39th International Conference on Machine Learning
  (ICML 2022)}, 2022.

\end{thebibliography}
\bibliographystyle{icml2022}

%%%%%%%%%%%%%%%%%%%%%%%%%%%%%%%%%%%%%%%%%%%%%%%%%%%%%%%%%%%%%%%%%%%%%%%%%%%%%%%
%%%%%%%%%%%%%%%%%%%%%%%%%%%%%%%%%%%%%%%%%%%%%%%%%%%%%%%%%%%%%%%%%%%%%%%%%%%%%%%
% APPENDIX
%%%%%%%%%%%%%%%%%%%%%%%%%%%%%%%%%%%%%%%%%%%%%%%%%%%%%%%%%%%%%%%%%%%%%%%%%%%%%%%
%%%%%%%%%%%%%%%%%%%%%%%%%%%%%%%%%%%%%%%%%%%%%%%%%%%%%%%%%%%%%%%%%%%%%%%%%%%%%%%
\newpage
\appendix
\onecolumn
\section{Visualization}

\begin{figure}[h]
    \centering
    \includegraphics[width=1\columnwidth]{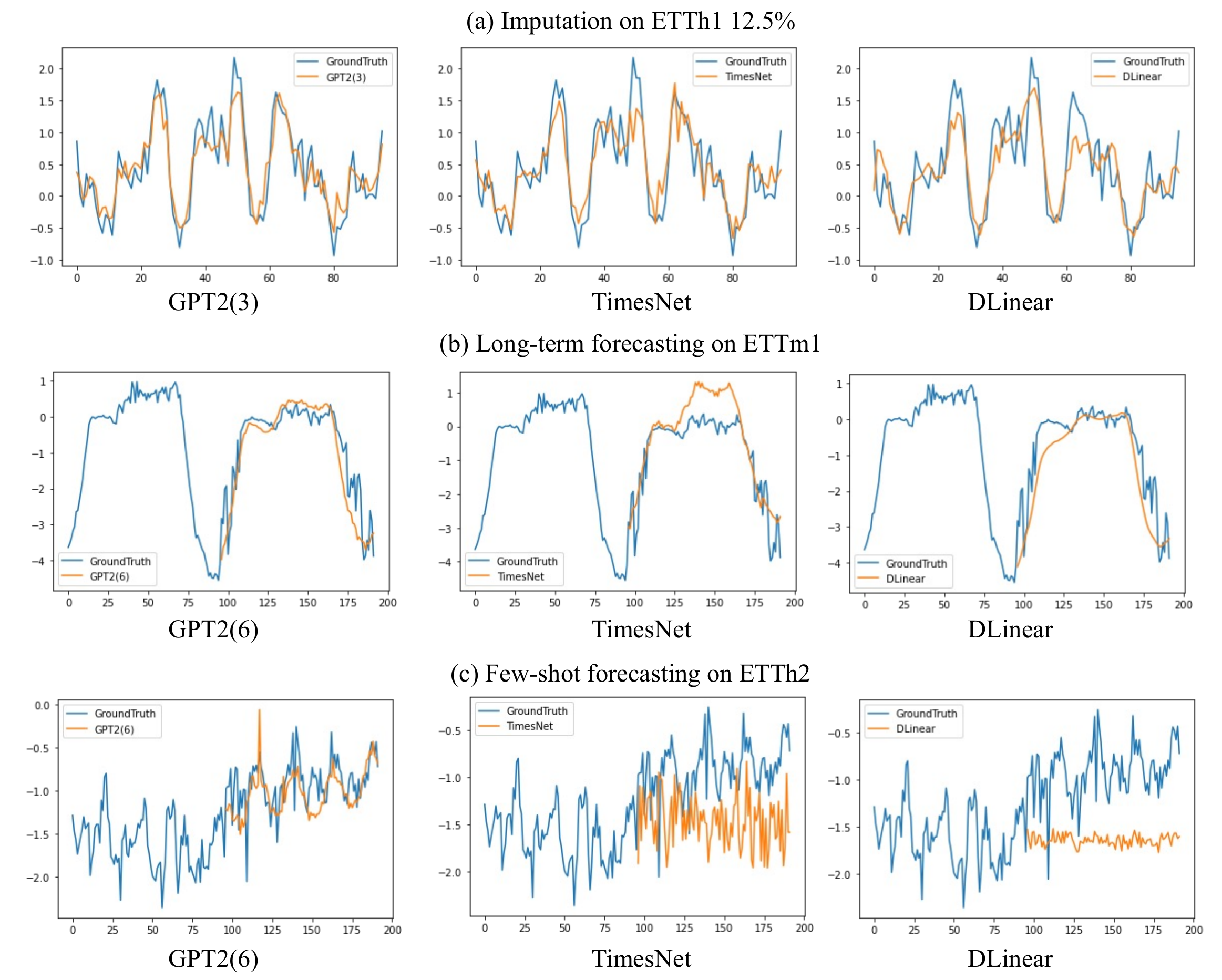}
    \caption{Visualization of imputation, long-term forecasting and few-shot forecasting.}
    \label{fig:visualization_plot}
\end{figure}

In order to clarify the representation ability more clearly, Figure \ref{fig:visualization_plot} provides showcases of imputation, long-term forecasting and few-shot forecasting. Especially for few-shot learning, GPT2(6) can accurately forecast, while TimesNet and DLinear fail in this task.

\section{Related Works}
\label{section:related}
%\paragraph{Time Series Forecasting}
We have presented a novel general time series analysis model in this paper, and to the best of our knowledge, there has been limited work on similar comprehensive methods for time series analysis. The most closely related field is time series forecasting, where transformer models have gained widespread popularity. Therefore, our focus in this related work will primarily be on introducing the end-to-end time series forecasting method.

Time series forecasting models can be roughly divided into three categories, ranging from the classic ARIMA models to the most recent transformer models. The first generation of well-discussed models can be dated back to auto-regressive family, such as ARIMA \cite{arima_1, arima_2} that follows the Markov process and recursively execute sequential forecasting. However, it is limited to stationary sequences while most time series is non-stationary. Additionally, with the bloom of deep neural networks, recurrent neural networks (RNNs), such as LSTM \cite{hochreiter1997long} and GRU \cite{chung2014empirical}, were designed for sequential tasks. Yet the recurrent model is inefficient for training and long-term dependencies are still under resolved.

Recently, transformer models have achieve great progress in NLP \cite{vaswani2017attention, Bert/NAACL/Jacob, gpt2-2019} and CV \cite{Transformers-for-image-at-scale/iclr/DosovitskiyB0WZ21, bao2022beit} tasks. Also, a large amount of transformer models are proposed to apply to time series forecasting~\cite{wen2022transformers}. In the following, we briefly introduce several representative algorithms. Informer \cite{zhou2021informer} proposes a probability sparse attention mechanism to deal with long-term dependencies. Autoformer \cite{wu2021autoformer} introduces a decomposition transformer architecture and replaces the attention module with an Auto-Correlation mechanism. FEDformer \cite{zhou2022fedformer} uses Fourier enhanced structure to improve computational efficiency and achieves linear complexity. Similar to 
 patching in ViT \cite{Transformers-for-image-at-scale/iclr/DosovitskiyB0WZ21}, PatchTST \cite{Patchformer} employs segmentation of time series that divide a sequence into patches to increase input length and reduce information redundancy. Besides, a simple MLP-based model DLinear \cite{dlinear} outperforms most transformer models and it validates channel-independence works well in time series forecasting. Recently, TimesNet~\cite{timesnet} has treated time series as a 2D signal and utilized a convolution-based inception net backbone to function as a comprehensive time series analysis model. This work is closely related to our tasks in this paper.

\section{Dataset Details}

In this section, we separately summarize dataset details long/short-term forecasting and few-shot/zero-shot forecasting.

\paragraph{Datasets of Long-term Forecasting and Few-shot Learning}

The details of datasets are shown as follows: 1) ETT datasets \cite{zhou2021informer} contain electricity load of various resolutions (ETTh \& ETTm) from two electricity stations. 2) Weather contains 21 meteorological indicators of Germany within 1 year; 3) Illness contains the influenza-like illness patients in
the United States; 4) Electricity dataset contains the electricity consumption; 5) Traffic dataset contains the occupation rate of freeway system across the State of California. Table \ref{tab:dataset} summarizes details of feature statistics.

Similar to PatchTST \cite{Patchformer}, Exchange is not contained. \cite{dlinear} shows that simply repeating the last value in the look-back window can outperform or be comparable to the best results.
Also, ILI is not used for few-shot learning for the limited quantity that is hard to follow the definition of few-shot.

\begin{table}[h]
\caption{Dataset details of few-shot learning.}
\label{tab:dataset}
\vskip 0.15in
\begin{center}
\begin{small}
\begin{tabular}{l|cccc}
\toprule
% Dataset & num & dim & freq \\
Dataset & Length & Dimension & Frequency \\
\midrule
ETTh & 17420 & 7 & 1 hour\\
ETTm & 69680 & 7 & 15 min\\
Weather & 52696 & 22 & 10 min & \\
ILI & 966 & 7 & 7 days\\
Electricity & 26304 & 321 & 1 hour \\
Traffic & 17544 & 862 & 1 hour \\
\bottomrule
\end{tabular}
\end{small}
\end{center}
\vskip -0.1in
\end{table}

\paragraph{Datasets of Short-term Forecasting and Zero-shot Learning}

The details of short-term forecasting and zero-shot learning datasets are shown as follows: 1) M4 is a large and diverse dataset that contains time series of various frequencies and fields, including business, financial and economic forecasting; 2) M3 is smaller than M4, but also contains time series from diverse domains and frequencies; 3) TOURISM is the dataset of tourism activities with different frequencies and contains a much higher fraction of erratic series compared with M4; 4) ELECTR represents the electricity usage monitoring of 370 customers over three years. Table \ref{tab:monash-zero} summarizes details of the datasets and zero-shot mapping between source and target.

\begin{table}[h]
\caption{Datasets and mapping details of zero-shot learning.}
\label{tab:zero_shot_dataset}
\vskip 0.15in
\begin{center}
\begin{small}
\begin{tabular}{l|cc|cc}
\toprule
% Dataset & num & dim & freq \\
& \multicolumn{2}{c|}{Dataset} & \multicolumn{2}{c}{Mapping} \\
& Length & Horizon & M4 & M3 \\
\midrule
M3 Yearly & 645 & 6 & Yearly & -\\
M3 Quarterly & 756 & 8 & Quarterly & -\\
M3 Monthly & 1428 & 18 & Monthly & -\\
M3 Others & 174 & 8 & Monthly & -\\
\midrule
M4 Yearly & 23000 & 6 & - & Yearly\\
M4 Quarterly & 24000 & 8 & - & Quarterly\\
M4 Monthly & 48000 & 18 & - & Monthly\\
M4 Weekly & 359 & 13 & - & Monthly\\
M4 Daily & 4227 & 14 & - & Monthly\\
M4 Hourly & 414 & 48 & - & Monthly\\
\midrule
TOURISM Yearly & 518 & 4 & Yearly & Yearly\\
TOURISM Quarterly & 427 & 8 & Quarterly & Quarterly\\
TOURISM Monthly & 366 & 24 & Monthly & Monthly\\
\midrule
ELECTR  & 1311 & 168 & Hourly & Monthly \\

\bottomrule
\end{tabular}
\end{small}
\end{center}
\vskip -0.1in
\end{table}

\section{Experimental Details}

All the deep learning networks are implemented in PyTorch and trained on NVIDIA V100 32GB GPUs. We use the pre-trained models from \cite{huggingface} for experiments. For few-shot learning, an early stopping counter is employed to stop the training process after three epochs if no loss degradation on the valid set is observed. Plus, we convert the multivariate data into univariate data. Specifically, we treat each feature of the sequence as a single time series. This is mainly for memory efficiency after patching of GPT2(6) and previous works, DLinear and PatchTST, have proved the effectiveness of channel-independence.

\subsection{Accuracy Metrics}
For long-term/short-term forecasting and few-shot forecasting, we use mean square error (MSE) and mean absolute error (MAE) as metrics. For zero-shot learning, mean absolute percentage error (MAPE) is used for TOURISM; symmetric MAPE (sMAPE) is used for M3 and M4; normalized deviation (ND) is used for ELECTR. All experiments are repeated 3 times and the mean of the metrics is used in the final results.

\subsection{Detailed Definition and Results for Few-shot and Long-term Forecasting}
\label{appendix:few-shot-learning}
\textbf{Task Definition}
Since \cite{dlinear} and \cite{Patchformer} have verified that channel-independence works well for time series datasets, we treat each multivariate series as multiple independent univariate series. Similar to traditional experimental settings, each time series is split into three parts: training data, validation data, and test data. For the few-shot forecasting task, only a certain percentage ($5\%$, $10\%$) timesteps of training data are used, and the other two parts remain unchanged. The evaluation metrics remain the same as for classic multivariate time series forecasting. We repeat this experiment 3 times and report the average metrics in the following experiments. 

\textbf{Detail Experiment Tables for Few-shot Time-Series Forecasting} in Table ~\ref{tab:5_percent_ett} and Table ~\ref{tab:10_percent_ett}

\begin{table*}[h!]
\caption{Few-shot learning results on 5\% data. We use prediction length $O \in \{96, 192, 336, 720\}$. A lower MSE indicates better performance, and the best results are highlighted in bold. '-' means that 5\% time series is not sufficient to constitute a training set.}
\label{tab:5_percent_ett}
% \vskip 0.15in
\begin{center}
\begin{small}
\scalebox{0.7}{
\setlength\tabcolsep{3pt}
\begin{tabular}{c|c|cc|cc|cc|cc|cc|cc|cc|cc|cc|cc|cc|cc}
\toprule

\multicolumn{2}{c|}{Methods}&\multicolumn{2}{c|}{GPT2(6)}&\multicolumn{2}{c|}{GPT2(0)}&\multicolumn{2}{c|}{DLinear}&\multicolumn{2}{c|}{PatchTST}&\multicolumn{2}{c|}{TimesNet}&\multicolumn{2}{c|}{FEDformer}&\multicolumn{2}{c|}{Autoformer}&\multicolumn{2}{c|}{Stationary}&\multicolumn{2}{c|}{ETSformer}&\multicolumn{2}{c|}{LightTS}&\multicolumn{2}{c|}{Informer}&\multicolumn{2}{c}{Reformer} \\

\midrule

\multicolumn{2}{c|}{Metric} & MSE  & MAE & MSE & MAE& MSE & MAE& MSE  & MAE& MSE  & MAE& MSE  & MAE& MSE  & MAE& MSE  & MAE& MSE  & MAE& MSE  & MAE& MSE  & MAE& MSE  & MAE\\

\midrule

\multirow{5}{*}{\rotatebox{90}{$Weather$}}
& 96  & 0.175 & 0.230 & 0.191 & 0.243 & 0.184 & 0.242 & 0.171 & 0.224 & 0.207& 0.253& 0.229 & 0.309 & 0.227 & 0.299 &0.215&0.252& 0.218 & 0.295 & 0.230 & 0.285 & 0.497 & 0.497 &0.406&0.435 \\
& 192 & 0.227 & 0.276 & 0.244 & 0.289 & 0.228 & 0.283 & 0.230 & 0.277 & 0.272& 0.307& 0.265 & 0.317 & 0.278 & 0.333 &0.290&0.307& 0.294 & 0.331 & 0.274 & 0.323 & 0.620 & 0.545 &0.446&0.450 \\
& 336 & 0.286 & 0.322 & 0.303 & 0.332 & 0.279 & 0.322 & 0.294 & 0.326 & 0.313& 0.328& 0.353 & 0.392 & 0.351 & 0.393 &0.353&0.348& 0.359 & 0.398 & 0.318 & 0.355 & 0.649 & 0.547 &0.465&0.459  \\
& 720 & 0.366 & 0.379 & 0.391 & 0.393 & 0.364 & 0.388 & 0.384 & 0.387 & 0.400& 0.385& 0.391 & 0.394 & 0.387 & 0.389 &0.452&0.407& 0.461 & 0.461 & 0.401 & 0.418 & 0.570 & 0.522  &0.471&0.468\\
&Avg.&\textbf{0.263}&\textbf{0.301}&0.282&0.314&\textbf{0.263}&0.308&0.269&0.303&0.298&0.318&0.309&0.353&0.310&0.353&0.327&0.328&0.333&0.371&0.305&0.345&0.584&0.527&0.447&0.453\\
\midrule

\multirow{5}{*}{\rotatebox{90}{$ETTh1$}}
& 96  & 0.543 & 0.506 & 0.825 & 0.638 & 0.547 & 0.503 & 0.557 & 0.519 & 0.892& 0.625& 0.593 & 0.529 & 0.681 & 0.570 &0.952&0.650& 1.169 & 0.832 & 1.483 & 0.91 & 1.225 & 0.812 &1.198&0.795 \\
& 192 & 0.748 & 0.580 & 1.220 & 0.778 & 0.720 & 0.604 & 0.711 & 0.570 &0.940 & 0.665& 0.652 & 0.563 & 0.725 & 0.602 &0.943&0.645& 1.221 & 0.853 & 1.525 & 0.93 & 1.249 & 0.828 &1.273&0.853\\
& 336 & 0.754 & 0.595 & 1.852 & 0.965 & 0.984 & 0.727 & 0.816 & 0.619 & 0.945&0.653 & 0.731 & 0.594 & 0.761 & 0.624 &0.935&0.644& 1.179 & 0.832 & 1.347 & 0.87 & 1.202 & 0.811 &1.254&0.857 \\
& 720 & - & - & - & - & - & - & - & - & - & - & - & - & - & - & - & - & - & - & - & - & - & - & - & - \\
&Avg.&0.681&\textbf{0.560}&1.299&0.793&0.750&0.611&0.694&0.569&0.925&0.647&\textbf{0.658}&0.562&0.722&0.598&0.943&0.646&1.189&0.839&1.451&0.903&1.225&0.817&1.241&0.835\\
\midrule

\multirow{5}{*}{\rotatebox{90}{$ETTh2$}}
& 96  & 0.376 & 0.421 & 0.551 & 0.507 & 0.442 & 0.456 & 0.401 & 0.421 & 0.409& 0.420& 0.390 & 0.424 & 0.428 & 0.468 &0.408&0.423 & 0.678 & 0.619 & 2.022 & 1.006 & 3.837 & 1.508 &3.753&1.518\\
& 192 & 0.418 & 0.441 & 0.765 & 0.610 & 0.617 & 0.542 & 0.452 & 0.455 & 0.483& 0.464& 0.457 & 0.465 & 0.496 & 0.504 &0.497&0.468 & 0.845 & 0.697 & 3.534 & 1.348 & 3.975 & 1.933 &3.516&1.473\\
& 336 & 0.408 & 0.439 & 0.767 & 0.614 & 1.424 & 0.849 & 0.464 & 0.469 & 0.499& 0.479& 0.477 & 0.483 & 0.486 & 0.496 &0.507&0.481 & 0.905 & 0.727 & 4.063 & 1.451 & 3.956 & 1.520 &3.312&1.427\\
& 720 & - & - & - & - & - & - & - & - & - & - & - & - & - & - & - & - & - & - & - & - & - & - & - & - \\
&Avg.&\textbf{0.400}&\textbf{0.433}&0.694&0.577&0.827&0.615&0.439&0.448&0.463&0.454&0.441&0.457&0.47&0.489&0.470&0.457&0.809&0.681&3.206&1.268&3.922&1.653&3.527&1.472\\
\midrule

\multirow{5}{*}{\rotatebox{90}{$ETTm1$}}
& 96  & 0.386 & 0.405 & 0.582 & 0.512 & 0.332 & 0.374 & 0.399 & 0.414 & 0.606& 0.518& 0.628 & 0.544 & 0.726 & 0.578 &0.823&0.587 & 1.031 & 0.747 & 1.048 & 0.733 & 1.130 & 0.775 &1.234&0.798\\
& 192 & 0.440 & 0.438 & 0.632 & 0.536 & 0.358 & 0.390 & 0.441 & 0.436 & 0.681& 0.539& 0.666 & 0.566 & 0.750 & 0.591 &0.844&0.591 & 1.087 & 0.766 & 1.097 & 0.756 & 1.150 & 0.788 &1.287&0.839\\
& 336 & 0.485 & 0.459 & 0.767 & 0.584 & 0.402 & 0.416 & 0.499 & 0.467 & 0.786& 0.597& 0.807 & 0.628 & 0.851 & 0.659 &0.870&0.603 & 1.138 & 0.787 & 1.147 & 0.775 & 1.198 & 0.809 &1.288&0.842\\
& 720 & 0.577 & 0.499 & 1.334 & 0.742 & 0.511 & 0.489 & 0.767 & 0.587 & 0.796& 0.593& 0.822 & 0.633 & 0.857 & 0.655 &0.893&0.611 & 1.245 & 0.831 & 1.200 & 0.799 & 1.175 & 0.794 &1.247&0.828\\
&Avg.&0.472&0.450&0.828&0.593&\textbf{0.400}&\textbf{0.417}&0.526&0.476&0.717&0.561&0.730&0.592&0.796&0.620&0.857&0.598&1.125&0.782&1.123&0.765&1.163&0.791&1.264&0.826\\
\midrule

\multirow{5}{*}{\rotatebox{90}{$ETTm2$}}
& 96  & 0.199 & 0.280 & 0.282 & 0.347 & 0.236 & 0.326 & 0.206 & 0.288 & 0.220& 0.299& 0.229 & 0.320 & 0.232 & 0.322 &0.238&0.316 & 0.404 & 0.485 & 1.108 & 0.772 & 3.599 & 1.478&3.883&1.545 \\
& 192 & 0.256 & 0.316 & 0.346 & 0.383 & 0.306 & 0.373 & 0.264 & 0.324 & 0.311& 0.361& 0.394 & 0.361 & 0.291 & 0.357 &0.298&0.349 & 0.479 & 0.521 & 1.317 & 0.850 & 3.578 & 1.475 &3.553&1.484\\
& 336 & 0.318 & 0.353 & 0.429 & 0.427 & 0.380 & 0.423 & 0.334 & 0.367 & 0.338& 0.366& 0.378 & 0.427 & 0.478 & 0.517 &0.353&0.380 & 0.552 & 0.555 & 1.415 & 0.879 & 3.561 & 1.473 &3.446&1.460\\
& 720 & 0.460 & 0.436 & 0.751 & 0.568 & 0.674 & 0.583 & 0.454 & 0.432 & 0.509& 0.465& 0.523 & 0.510 & 0.553 & 0.538 &0.475&0.445 & 0.701 & 0.627 & 1.822 & 0.984 & 3.896 & 1.533 &3.445&1.460\\
&Avg.&\textbf{0.308}&\textbf{0.346}&0.452&0.431&0.399&0.426&0.314&0.352&0.344&0.372&0.381&0.404&0.388&0.433&0.341&0.372&0.534&0.547&1.415&0.871&3.658&1.489&3.581&1.487\\
\midrule

% \multirow{4}{*}{\rotatebox{90}{$ILI$}}
% & 24  & 4.225 & 1.588 & 6.565 & 1.894 & 4.803 & 1.657 & \textbf{3.749} & \textbf{1.419} & 5.193 & 1.691 & 5.404 & 1.756 \\
% & 36 & - & - & - & - & - & - & - & - & - & - & - & -\\
% & 48 & - & - & - & - & - & - & - & - & - & - & - & -\\
% & 60 & - & - & - & - & - & - & - & - & - & - & - & -\\
% \midrule

\multirow{5}{*}{\rotatebox{90}{$ECL$}}
& 96  & 0.143 & 0.241 & 0.147 & 0.246 & 0.150 & 0.251 & 0.145 & 0.244 & 0.315& 0.389 & 0.235 & 0.322 & 0.297 & 0.367 &0.484&0.518 & 0.697 & 0.638 & 0.639 & 0.609 & 1.265 & 0.919 &1.414&0.855\\
& 192 & 0.159 & 0.255 & 0.163 & 0.260 & 0.163 & 0.263 & 0.163 & 0.260 & 0.318& 0.396& 0.247 & 0.341 & 0.308 & 0.375 &0.501&0.531 & 0.718 & 0.648 & 0.772 & 0.678 & 1.298 & 0.939 &1.240&0.919\\
& 336 & 0.179 & 0.274 & 0.182 & 0.278 & 0.175 & 0.278 & 0.183 & 0.281 & 0.340& 0.415& 0.267 & 0.356 & 0.354 & 0.411 &0.574&0.578 & 0.758 & 0.667 & 0.901 & 0.745 & 1.302 & 0.942 &1.253&0.921\\
& 720 & 0.233 & 0.323 & 0.239 & 0.329 & 0.219 & 0.311 & 0.233 & 0.323 & 0.635&0.613 & 0.318 & 0.394 & 0.426 & 0.466 &0.952&0.786 & 1.028 & 0.788 & 1.200 & 0.871 & 1.259 & 0.919 &1.249&0.921\\
&Avg.&0.178&\textbf{0.273}&0.182&0.278&\textbf{0.176}&0.275&0.181&0.277&0.402&0.453&0.266&0.353&0.346&0.404&0.627&0.603&0.800&0.685&0.878&0.725&1.281&0.929&1.289&0.904\\
\midrule

\multirow{5}{*}{\rotatebox{90}{$Traffic$}}
& 96  & 0.419 & 0.298 & 0.468 & 0.354 & 0.427 & 0.304 & 0.404 & 0.286 & 0.854& 0.492& 0.670 & 0.421 & 0.795 & 0.481 &1.468&0.821& 1.643 & 0.855 & 1.157 & 0.636 & 1.557 & 0.821  &1.586&0.841\\
& 192 & 0.434 & 0.305 & 0.479 & 0.352 & 0.447 & 0.315 & 0.412 & 0.294 & 0.894& 0.517& 0.653 & 0.405 & 0.837 & 0.503 &1.509&0.838 & 1.856 & 0.928 & 1.688 & 0.848 & 1.596 & 0.834 &1.602&0.844\\
& 336 & 0.449 & 0.313 & 0.477 & 0.345 & 0.478 & 0.333 & 0.439 & 0.310 & 0.853& 0.471& 0.707 & 0.445 & 0.867 & 0.523 &1.602&0.860 & 2.080 & 0.999 & 1.826 & 0.903 & 1.621 & 0.841&1.668&0.868 \\
& 720 & - & - & - & - & - & - & - & - & - & - & - & - & - & - & - & - & - & - & - & - & - & - & - & - \\
&Avg.&0.434&0.305&0.474&0.350&0.450&0.317&\textbf{0.418}&\textbf{0.296}&0.867&0.493&0.676&0.423&0.833&0.502&1.526&0.839&1.859&0.927&1.557&0.795&1.591&0.832&1.618&0.851\\
\midrule
\multicolumn{2}{c|}{Average} &\textbf{0.377}&\textbf{0.375}&0.575&0.465&0.441&0.413&\color{red}\textbf{0.392}&\color{red}\textbf{0.383}&0.552&0.464&0.483&0.445&0.537&0.480&0.697&0.537&0.909&0.675&1.341&0.789&1.878&0.994&1.819&0.966\\

% \midrule
% \multicolumn{2}{c|}{Count} &11&13&0&0&7&7&5&4&&&2&2&0&0\\

\bottomrule
\end{tabular}
}
\end{small}
\end{center}
\vskip -0.1in
\end{table*}

\clearpage

\begin{table*}[h!]
\caption{Few-shot learning results on 10\% data. We use prediction length $O \in \{96, 192, 336, 720\}$. A lower MSE indicates better performance, and the best results are highlighted in bold. '-' means that 10\% time series is not sufficient to constitute a training set.}
\label{tab:10_percent_ett}
% \vskip 0.15in
\begin{center}
\begin{small}
\scalebox{0.70}{
\setlength\tabcolsep{3pt}
\begin{tabular}{c|c|cc|cc|cc|cc|cc|cc|cc|cc|cc|cc|cc|cc}
\toprule

\multicolumn{2}{c|}{Methods}&\multicolumn{2}{c|}{GPT2(6)}&\multicolumn{2}{c|}{GPT2(0)}&\multicolumn{2}{c|}{DLinear}&\multicolumn{2}{c|}{PatchTST}&\multicolumn{2}{c|}{TimesNet}&\multicolumn{2}{c|}{FEDformer}&\multicolumn{2}{c|}{Autoformer}&\multicolumn{2}{c|}{Stationary}&\multicolumn{2}{c|}{ETSformer}&\multicolumn{2}{c|}{LightTS}&\multicolumn{2}{c|}{Informer}&\multicolumn{2}{c}{Reformer} \\

\midrule

\multicolumn{2}{c|}{Metric} & MSE  & MAE & MSE & MAE& MSE & MAE& MSE  & MAE& MSE  & MAE& MSE  & MAE& MSE  & MAE& MSE  & MAE& MSE  & MAE& MSE  & MAE& MSE  & MAE& MSE  & MAE\\
\midrule

\multirow{5}{*}{\rotatebox{90}{$Weather$}}
& 96  & 0.163 & 0.215 & 0.190 & 0.240 & 0.171 & 0.224 & 0.165 & 0.215 & 0.184& 0.230& 0.188 & 0.253 & 0.221 & 0.297 &0.192&0.234& 0.199 & 0.272 & 0.217 & 0.269 & 0.374 & 0.401&0.335&0.380 \\
& 192 & 0.210 & 0.254 & 0.243 & 0.284 & 0.215 & 0.263 & 0.210 & 0.257 & 0.245& 0.283& 0.250 & 0.304 & 0.270 & 0.322 &0.269&0.295& 0.279 & 0.332 & 0.259 & 0.304 & 0.552 & 0.478 &0.522&0.462\\
& 336 & 0.256 & 0.292 & 0.270 & 0.305 & 0.258 & 0.299 & 0.259 & 0.297 & 0.305& 0.321& 0.312 & 0.346 & 0.320 & 0.351 &0.370&0.357& 0.356 & 0.386 & 0.303 & 0.334 & 0.724 & 0.541&0.715&0.535 \\
& 720 & 0.321 & 0.339 & 0.348 & 0.359 & 0.320 & 0.346 & 0.332 & 0.346 & 0.381& 0.371& 0.387 & 0.393 & 0.390 & 0.396 &0.441&0.405& 0.437 & 0.448 & 0.377 & 0.382 & 0.739 & 0.558 &0.611&0.500\\
&Avg.&\textbf{0.238} &\textbf{0.275} &0.263 &0.297 &0.241 &0.283 &0.242 &0.279 &0.279 &0.301 &0.284 &0.324 &0.300 &0.342 &0.318 &0.323 &0.318 &0.360 &0.289 &0.322 &0.597 &0.495 &0.546 &0.469\\
\midrule

\multirow{5}{*}{\rotatebox{90}{$ETTh1$}}
& 96  & 0.458 & 0.456 & 0.601 & 0.536 & 0.492 & 0.495 & 0.516 & 0.485 & 0.861& 0.628& 0.512 & 0.499 & 0.613 & 0.552 &0.918&0.639& 1.112 & 0.806 & 1.298 & 0.838 & 1.179 & 0.792 &1.184&0.790\\
& 192 & 0.570 & 0.516 & 0.709 & 0.587 & 0.565 & 0.538 & 0.598 & 0.524 & 0.797& 0.593& 0.624 & 0.555 & 0.722 & 0.598 &0.915&0.629& 1.155 & 0.823 & 1.322 & 0.854 & 1.199 & 0.806 &1.295&0.850\\
& 336 & 0.608 & 0.535 & 0.801 & 0.635 & 0.721 & 0.622 & 0.657 & 0.550 & 0.941& 0.648& 0.691 & 0.574 & 0.750 & 0.619 &0.939&0.644& 1.179 & 0.832 & 1.347 & 0.870 & 1.202 & 0.811 &1.294&0.854\\
& 720 & 0.725 & 0.591 & 1.385 & 0.831 & 0.986 & 0.743 & 0.762 & 0.610 &0.877 &0.641 & 0.728 & 0.614 & 0.721 & 0.616 &0.887&0.645& 1.273 & 0.874 & 1.534 & 0.947 & 1.217 & 0.825 &1.223&0.838\\
&Avg.&\textbf{0.590} &\textbf{0.525} &0.874 &0.647 &0.691 &0.600 &0.633 &0.542 &0.869 &0.628 &0.639 &0.561 &0.702 &0.596 &0.915 &0.639 &1.180 &0.834 &1.375 &0.877 &1.199 &0.809 &1.249 &0.833\\
\midrule

\multirow{5}{*}{\rotatebox{90}{$ETTh2$}}
& 96  & 0.331 & 0.374 & 0.539 & 0.495 & 0.357 & 0.411 & 0.353 & 0.389 & 0.378& 0.409& 0.382 & 0.416 & 0.413 & 0.451 &0.389&0.411& 0.678 & 0.619 & 2.022 & 1.006 & 3.837 & 1.508&3.788&1.533\\
& 192 & 0.402 & 0.411 & 0.675 & 0.555 & 0.569 & 0.519 & 0.403 & 0.414 & 0.490& 0.467& 0.478 & 0.474 & 0.474 & 0.477 &0.473&0.455& 0.785 & 0.666 & 2.329 & 1.104 & 3.856 & 1.513 &3.552&1.483\\
& 336 & 0.406 & 0.433 & 0.718 & 0.580 & 0.671 & 0.572 & 0.426 & 0.441 & 0.537& 0.494& 0.504 & 0.501 & 0.547 & 0.543 &0.507&0.480& 0.839 & 0.694 & 2.453 & 1.122 & 3.952 & 1.526 &3.395&1.526\\
& 720 & 0.449 & 0.464 & 0.732 & 0.605 & 0.824 & 0.648 & 0.477 & 0.480 & 0.510& 0.491& 0.499 & 0.509 & 0.516 & 0.523 &0.477&0.472& 1.273 & 0.874 & 3.816 & 1.407 & 3.842 & 1.503 &3.205&1.401\\
&Avg.&\textbf{0.397} &\textbf{0.421} &0.666 &0.559 &0.605 &0.538 &0.415 &0.431 &0.479 &0.465 &0.466 &0.475 &0.488 &0.499 &0.462 &0.455 &0.894 &0.713 &2.655 &1.160 &3.872 &1.513 &3.485 &1.486\\
\midrule

\multirow{5}{*}{\rotatebox{90}{$ETTm1$}}
& 96  & 0.390 & 0.404 & 0.610 & 0.508 & 0.352 & 0.392 & 0.410 & 0.419 & 0.583& 0.501& 0.578 & 0.518 & 0.774 & 0.614 &0.761&0.568& 0.911 & 0.688 & 0.921 & 0.682 & 1.162 & 0.785&1.442&0.847\\
& 192 & 0.429 & 0.423 & 0.666 & 0.540 & 0.382 & 0.412 & 0.437 & 0.434 & 0.630& 0.528& 0.617 & 0.546 & 0.754 & 0.592 &0.781&0.574& 0.955 & 0.703 & 0.957 & 0.701 & 1.172 & 0.793 &1.444&0.862\\
& 336 & 0.469 & 0.439 & 0.895 & 0.615 & 0.419 & 0.434 & 0.476 & 0.454 & 0.725& 0.568& 0.998 & 0.775 & 0.869 & 0.677 &0.803&0.587& 0.991 & 0.719 & 0.998 & 0.716 & 1.227 & 0.908&1.450&0.866 \\
& 720 & 0.569 & 0.498 & 0.916 & 0.646 & 0.490 & 0.477 & 0.681 & 0.556 & 0.769& 0.549& 0.693 & 0.579 & 0.810 & 0.630 &0.844&0.581& 1.062 & 0.747 & 1.007 & 0.719 & 1.207 & 0.797&1.366&0.850 \\
&Avg.&0.464 &0.441 &0.772 &0.577 &\textbf{0.411} &\textbf{0.429} &0.501 &0.466 &0.677 &0.537 &0.722 &0.605 &0.802 &0.628 &0.797 &0.578 &0.980 &0.714 &0.971 &0.705 &1.192 &0.821 &1.426 &0.856\\
\midrule

\multirow{5}{*}{\rotatebox{90}{$ETTm2$}}
& 96  & 0.188 & 0.269 & 0.283 & 0.344 & 0.213 & 0.303 & 0.191 & 0.274 & 0.212& 0.285& 0.291 & 0.399 & 0.352 & 0.454 &0.229&0.308& 0.331 & 0.430 & 0.813 & 0.688 & 3.203 & 1.407 &4.195&1.628\\
& 192 & 0.251 & 0.309 & 0.353 & 0.384 & 0.278 & 0.345 & 0.252 & 0.317 & 0.270& 0.323& 0.307 & 0.379 & 0.694 & 0.691 &0.291&0.343& 0.400 & 0.464 & 1.008 & 0.768 & 3.112 & 1.387&4.042&1.601 \\
& 336 & 0.307 & 0.346 & 0.420 & 0.422 & 0.338 & 0.385 & 0.306 & 0.353 & 0.323& 0.353& 0.543 & 0.559 & 2.408 & 1.407 &0.348&0.376& 0.469 & 0.498 & 1.031 & 0.775 & 3.255 & 1.421&3.963&1.585 \\
& 720 & 0.426 & 0.417 & 0.553 & 0.491 & 0.436 & 0.440 & 0.433 & 0.427 & 0.474& 0.449& 0.712 & 0.614 & 1.913 & 1.166 &0.461&0.438& 0.589 & 0.557 & 1.096 & 0.791 & 3.909 & 1.543&3.711&1.532 \\
&Avg.&\textbf{0.293} &\textbf{0.335} &0.402 &0.410 &0.316 &0.368 &0.296 &0.343 &0.320 &0.353 &0.463 &0.488 &1.342 &0.930 &0.332 &0.366 &0.447 &0.487 &0.987 &0.756 &3.370 &1.440 &3.978 &1.587\\
\midrule

% \multirow{4}{*}{\rotatebox{90}{$ILI$}}
% & 24 & \textbf{3.022} & \textbf{1.247} & 3.521 & 1.274 & 3.969 & 1.536 & 3.144 & 1.261 &5.526 &1.700 & 4.338 & 1.525 & 3.733 & 1.380 \\
% & 36 & 3.854 & 1.453 & 3.466 & 1.340 & 3.700 & 1.467 & \textbf{2.950} & \textbf{1.237} &5.092 &1.611 & 5.166 & 1.658 & 4.342 & 1.507 \\
% & 48 & 4.603 & 1.571 & 4.435 & 1.554 & 3.980 & 1.485 & 3.501 & 1.358 &4.729 &1.580 & 4.404 & 1.523 & 4.267 & 1.493 \\
% & 60 & - & - & - & - & - & - & - & - &4.680 &1.573 & 4.380 & \textbf{1.498} & \textbf{4.429} & 1.527 \\
% \midrule

\multirow{5}{*}{\rotatebox{90}{$ECL$}}
& 96  & 0.139 & 0.237 & 0.142 & 0.240 & 0.150 & 0.253 & 0.140 & 0.238 & 0.299& 0.373& 0.231 & 0.323 & 0.261 & 0.348 &0.420&0.466& 0.599 & 0.587 & 0.350 & 0.425 & 1.259 & 0.919 &0.993&0.784\\
& 192 & 0.156 & 0.252 & 0.158 & 0.254 & 0.164 & 0.264 & 0.160 & 0.255 & 0.305& 0.379& 0.261 & 0.356 & 0.338 & 0.406 &0.411&0.459& 0.620 & 0.598 & 0.376 & 0.448 & 1.160 & 0.873 &0.938&0.753\\
& 336 & 0.175 & 0.270 & 0.175 & 0.271 & 0.181 & 0.282 & 0.180 & 0.276 &0.319 &0.391 & 0.360 & 0.445 & 0.410 & 0.474 &0.434&0.473& 0.662 & 0.619 & 0.428 & 0.485 & 1.157 & 0.872&0.925&0.745 \\
& 720 & 0.233 & 0.317 & 0.230 & 0.315 & 0.223 & 0.321 & 0.241 & 0.323 & 0.369& 0.426& 0.530 & 0.585 & 0.715 & 0.685 &0.510&0.521& 0.757 & 0.664 & 0.611 & 0.597 & 1.203 & 0.898&1.004&0.790 \\
&Avg.&\textbf{0.176} &\textbf{0.269} &\textbf{0.176} &0.270 &0.180 &0.280 &0.180 &0.273 &0.323 &0.392 &0.346 &0.427 &0.431 &0.478 &0.444 &0.480 &0.660 &0.617 &0.441 &0.489 &1.195 &0.891 &0.965 &0.768\\
\midrule

\multirow{5}{*}{\rotatebox{90}{$Traffic$}}
& 96  & 0.414 & 0.297 & 0.478 & 0.368 & 0.419 & 0.298 & 0.403 & 0.289 & 0.719& 0.416& 0.639 & 0.400 & 0.672 & 0.405 &1.412&0.802& 1.643 & 0.855 & 1.157 & 0.636 & 1.557 & 0.821 &1.527&0.815\\
& 192 & 0.426 & 0.301 & 0.481 & 0.363 & 0.434 & 0.305 & 0.415 & 0.296 & 0.748& 0.428& 0.637 & 0.416 & 0.727 & 0.424 &1.419&0.806& 1.641 & 0.854 & 1.207 & 0.661 & 1.454 & 0.765 &1.538&0.817\\
& 336 & 0.434 & 0.303 & 0.488 & 0.365 & 0.449 & 0.313 & 0.426 & 0.304 & 0.853& 0.471& 0.655 & 0.427 & 0.749 & 0.454 &1.443&0.815& 1.711 & 0.878 & 1.334 & 0.713 & 1.521 & 0.812&1.550&0.819 \\
& 720 & 0.487 & 0.337 & 0.537 & 0.386 & 0.484 & 0.336 & 0.474 & 0.331 & 1.485& 0.825& 0.722 & 0.456 & 0.847 & 0.499 &1.539&0.837& 2.660 & 1.157 & 1.292 & 0.726 & 1.605 & 0.846 &1.588&0.833\\
&Avg.&0.440 &0.310 &0.496 &0.371 &0.447 &0.313 &\textbf{0.430} &\textbf{0.305} &0.951 &0.535 &0.663 &0.425 &0.749 &0.446 &1.453 &0.815 &1.914 &0.936 &1.248 &0.684 &1.534 &0.811 &1.551 &0.821\\
\midrule
\multicolumn{2}{c|}{Average}& \textbf{0.371} &\textbf{0.367} &0.521 &0.447 &0.413 &0.401 &\color{red}\textbf{0.385} &\color{red}\textbf{0.376} &0.556 &0.458 &0.511 &0.472 &0.687 &0.559 &0.674 &0.522 &0.912 &0.665 &1.137 &0.712 &1.850 &0.967 &1.888 &0.974  \\

\bottomrule
\end{tabular}
}
\end{small}
\end{center}
\vskip -0.1in
\end{table*}

\clearpage

\subsection{Long-term Time-series Forecasting}
\label{appendix:full-data}
Here we investigate whether our architecture performs consistently well with more training data. Thus, we follow the classical experiment settings of \cite{Patchformer} and conduct experiments on full data. The results are shown in Table \ref{tab:100_percent_ett_full}.
Overall, GPT2(6) FPT achieves comparable performance to PatchTST, Dlinear and outperforms other baselines by a large margin.
Compared with the second best transformer-based baseline method FEDformer, GPT2(6) FPT yields an overall \textbf{18.7\%} relatively MSE reduction.
It verifies the effectiveness of NLP pretrained model in time series forecasting, not limited to the few-shot setting.

\textbf{Detail Experiment Table for Long-term Time-Series Forecasting} in table ~\ref{tab:100_percent_ett_full}

\begin{table*}[h]
\caption{Full results on full data. We use prediction length $O \in \{96, 192, 336, 720\}$ for ILI and $O \in \{24, 36, 48, 60\}$ for others. A lower MSE indicates better performance. \textbf{Black}: best, {\color{red} \textbf{Red}}: second best.}
\label{tab:100_percent_ett_full}
\vskip 0.15in
\begin{center}
\begin{small}
\scalebox{0.70}{
\setlength\tabcolsep{3pt}
\begin{tabular}{c|c|cc|cc|cc|cc|cc|cc|cc|cc|cc|cc|cc|cc}
\toprule

\multicolumn{2}{c|}{Methods}&\multicolumn{2}{c|}{GPT2(6)}&\multicolumn{2}{c|}{GPT2(0)}&\multicolumn{2}{c|}{DLinear}&\multicolumn{2}{c|}{PatchTST}&\multicolumn{2}{c|}{TimesNet}&\multicolumn{2}{c|}{FEDformer}&\multicolumn{2}{c|}{Autoformer}&\multicolumn{2}{c|}{Stationary}&\multicolumn{2}{c|}{ETSformer}&\multicolumn{2}{c|}{LightTS}&\multicolumn{2}{c|}{Informer}&\multicolumn{2}{c}{Reformer} \\

\midrule

\multicolumn{2}{c|}{Metric} & MSE  & MAE & MSE & MAE& MSE & MAE& MSE  & MAE& MSE  & MAE& MSE  & MAE& MSE  & MAE& MSE  & MAE& MSE  & MAE& MSE  & MAE& MSE  & MAE& MSE  & MAE\\
\midrule

\multirow{5}{*}{\rotatebox{90}{$Weather$}}
& 96  & 0.162 & 0.212 & 0.181 & 0.232 & 0.176 & 0.237 & 0.149 & 0.198 &0.172&0.220&0.217&0.296&0.266&0.336&0.173&0.223&0.197&0.281&0.182&0.242&0.300&0.384&0.689&0.596\\
& 192 & 0.204 & 0.248 & 0.222 & 0.266 & 0.220 & 0.282 & 0.194 & 0.241 &0.219&0.261&0.276&0.336&0.307&0.367&0.245&0.285&0.237&0.312&0.227&0.287&0.598&0.544&0.752&0.638\\
& 336 & 0.254 & 0.286 & 0.270 & 0.299 & 0.265 & 0.319 & 0.245 & 0.282&0.280&0.306&0.339&0.380&0.359&0.395&0.321&0.338&0.298&0.353&0.282&0.334&0.578&0.523&0.639&0.596 \\
& 720 & 0.326 & 0.337 & 0.338 & 0.345 & 0.333 & 0.362 & 0.314 & 0.334&0.365&0.359&0.403&0.428&0.419&0.428&0.414&0.410&0.352&0.288&0.352&0.386&1.059&0.741&1.130&0.792\\
& Avg &\color{red}\textbf{0.237}&\color{red}\textbf{0.270}&0.252&0.285&0.248&0.300&\textbf{0.225}&\textbf{0.264}&0.259&0.287&0.309&0.360&0.338&0.382&0.288&0.314&0.271&0.334&0.261&0.312&0.634&0.548&0.803&0.656\\
\midrule

\multirow{5}{*}{\rotatebox{90}{$ETTh1$}}
& 96  & 0.376 & 0.397 & 0.422 & 0.428 & 0.375 & 0.399 & 0.370 & 0.399 &0.384&0.402&0.376&0.419&0.449&0.459&0.513&0.491&0.494&0.479&0.424&0.432&0.865&0.713&0.837&0.728\\
& 192 & 0.416 & 0.418 & 0.466 & 0.450 & 0.405 & 0.416 & 0.413 & 0.421&0.436&0.429&0.420&0.448&0.500&0.482&0.534&0.504&0.538&0.504&0.475&0.462&1.008&0.792&0.923&0.766\\
& 336 & 0.442 & 0.433 & 0.488 & 0.464 &0.439 & 0.443 & 0.422 & 0.436 &0.491&0.469&0.459&0.465&0.521&0.496&0.588&0.535&0.574&0.521&0.518&0.488&1.107&0.809&1.097&0.835\\
& 720 & 0.477 & 0.456 & 0.485 & 0.478 & 0.472 & 0.490 & 0.447 & 0.466 &0.521&0.500&0.506&0.507&0.514&0.512&0.643&0.616&0.562&0.535&0.547&0.533&1.181&0.865&1.257&0.889\\
& Avg &\color{red}\textbf{0.427}&\textbf{0.426}&0.465&0.455&0.422&0.437&\textbf{0.413}&\color{red}\textbf{0.430}&0.458&0.450&0.440&0.460&0.496&0.487&0.570&0.537&0.542&0.510&0.491&0.479&1.040&0.795&1.029&0.805\\
\midrule

\multirow{5}{*}{\rotatebox{90}{$ETTh2$}}
& 96  & 0.285 & 0.342 & 0.318 & 0.368 & 0.289 & 0.353 & 0.274 & 0.336&0.340&0.374&0.358&0.397&0.346&0.388&0.476&0.458&0.340&0.391&0.397&0.437&3.755&1.525&2.626&1.317 \\
& 192 & 0.354 & 0.389 & 0.383 & 0.407 & 0.383 & 0.418 & 0.339 &  0.379 &0.402&0.414&0.429&0.439&0.456&0.452&0.512&0.493&0.430&0.439&0.520&0.504&5.602&1.931&11.12&2.979\\
& 336 & 0.373 & 0.407 & 0.406 & 0.427 & 0.448 & 0.465 & 0.329 & 0.380&0.452&0.452&0.496&0.487&0.482&0.486&0.552&0.551&0.485&0.479&0.626&0.559&4.721&1.835&9.323&2.769 \\
& 720 & 0.406 & 0.441 & 0.420 & 0.446 & 0.605 & 0.551 & 0.379 & 0.422&0.462&0.468&0.463&0.474&0.515&0.511&0.562&0.560&0.500&0.497&0.863&0.672&3.647&1.625&3.874&1.697\\
& Avg &\color{red}\textbf{0.354}&\color{red}\textbf{0.394}&0.381&0.412&0.431&0.446&\textbf{0.330}&\textbf{0.379}&0.414&0.427&0.437&0.449&0.450&0.459&0.526&0.516&0.439&0.452&0.602&0.543&4.431&1.729&6.736&2.191\\
\midrule

\multirow{5}{*}{\rotatebox{90}{$ETTm1$}}
& 96  & 0.292 & 0.346 & 0.330 & 0.372 & 0.299 & 0.343 & 0.290 & 0.342 &0.338&0.375&0.379&0.419&0.505&0.475&0.386&0.398&0.375&0.398&0.374&0.400&0.672&0.571&0.538&0.528 \\
& 192 & 0.332 & 0.372 & 0.371 & 0.394 & 0.335 & 0.365 & 0.332 & 0.369&0.374&0.387&0.426&0.441&0.553&0.496&0.459&0.444&0.408&0.410&0.400&0.407&0.795&0.669&0.658&0.592\\
& 336 & 0.366 & 0.394 & 0.398 & 0.409 & 0.369 & 0.386 & 0.366 & 0.392&0.410&0.411&0.445&0.459&0.621&0.537&0.495&0.464&0.435&0.428&0.438&0.438&1.212&0.871&0.898&0.721\\
& 720 & 0.417 & 0.421 & 0.454 & 0.440 & 0.425 & 0.421 & 0.416 & 0.420&0.478&0.450&0.543&0.490&0.671&0.561&0.585&0.516&0.499&0.462&0.527&0.502&1.166&0.823&1.102&0.841\\
& Avg&\color{red}\textbf{0.352}&\color{red}\textbf{0.383}&0.388&0.403&0.357&\textbf{0.378}&\textbf{0.351}&0.380&0.400&0.406&0.448&0.452&0.588&0.517&0.481&0.456&0.429&0.425&0.435&0.437&0.961&0.734&0.799&0.671 \\
\midrule

\multirow{5}{*}{\rotatebox{90}{$ETTm2$}}
& 96  & 0.173 & 0.262 & 0.192 & 0.281 & 0.167 & 0.269 & 0.165 & 0.255&0.187&0.267&0.203&0.287&0.255&0.339&0.192&0.274&0.189&0.280&0.209&0.308&0.365&0.453&0.658&0.619\\
& 192 & 0.229 & 0.301 & 0.245 & 0.317 & 0.224 & 0.303 & 0.220 & 0.292&0.249&0.309&0.269&0.328&0.281&0.340&0.280&0.339&0.253&0.319&0.311&0.382&0.533&0.563&1.078&0.827\\
& 336 & 0.286 & 0.341 & 0.302 & 0.352 & 0.281 & 0.342 & 0.274 & 0.329 &0.321&0.351&0.325&0.366&0.339&0.372&0.334&0.361&0.314&0.357&0.442&0.466&1.363&0.887&1.549&0.972\\
& 720 & 0.378 & 0.401 & 0.399 & 0.408 & 0.397 & 0.421 & 0.362 & 0.385&0.408&0.403&0.421&0.415&0.433&0.432&0.417&0.413&0.414&0.413&0.675&0.587&3.379&1.338&2.631&1.242 \\
& Avg&\color{red}\textbf{0.266}&\color{red}\textbf{0.326}&0.284&0.339&0.267&0.333&\textbf{0.255}&\textbf{0.315}&0.291&0.333&0.305&0.349&0.327&0.371&0.306&0.347&0.293&0.342&0.409&0.436&1.410&0.810&1.479&0.915 \\
\midrule

\multirow{5}{*}{\rotatebox{90}{$ILI$}}
& 24 &2.063 & 0.881 & 2.723 & 1.099 & 2.215 & 1.081 & 1.319 &0.754&2.317&0.934&3.228&1.260&3.483&1.287&2.294&0.945&2.527&1.020&8.313&2.144&5.764&1.677&4.400&1.382\\
& 36 & 1.868 & 0.892 & 2.027 & 0.966 & 1.963 & 0.963 & 1.430 & 0.834&1.972&0.920&2.679&1.080&3.103&1.148&1.825&0.848&2.615&1.007&6.631&1.902&4.755&1.467&4.783&1.448 \\
& 48 & 1.790 & 0.884 & 2.206 & 1.022 & 2.130 & 1.024 & 1.553 & 0.815 &2.238&0.940&2.622&1.078&2.669&1.085&2.010&0.900&2.359&0.972&7.299&1.982&4.763&1.469&4.832&1.465 \\
& 60 & 1.979 &0.957 &1.976 & 0.983 & 2.368 & 1.096 & 1.470 & 0.788&2.027&0.928&2.857&1.157&2.770&1.125&2.178&0.963&2.487&1.016&7.283&1.985&5.264&1.564&4.882&1.483 \\
& Avg &\color{red}\textbf{1.925}&\color{red}\textbf{0.903}&2.233&1.017&2.169&1.041&\textbf{1.443}&\textbf{0.797}&2.139&0.931&2.847&1.144&3.006&1.161&2.077&0.914&2.497&1.004&7.382&2.003&5.137&1.544&4.724&1.445\\
\midrule

\multirow{5}{*}{\rotatebox{90}{$ECL$}}
& 96  & 0.139 & 0.238 &0.138 & 0.234 & 0.140 & 0.237 & 0.129 & 0.222&0.168&0.272&0.193&0.308&0.201&0.317&0.169&0.273&0.187&0.304&0.207&0.307&0.274&0.368&0.312&0.402 \\
& 192 & 0.153 & 0.251 & 0.152 & 0.247 & 0.153 & 0.249 & 0.157 &0.240&0.184&0.289&0.201&0.315&0.222&0.334&0.182&0.286&0.199&0.315&0.213&0.316&0.296&0.386&0.348&0.433\\
& 336 & 0.169 & 0.266 & 0.168 & 0.263 & 0.169 & 0.267 & 0.163 & 0.259&0.198&0.300&0.214&0.329&0.231&0.338&0.200&0.304&0.212&0.329&0.230&0.333&0.300&0.394&0.350&0.433\\
& 720 & 0.206 & 0.297 & 0.207 & 0.295 & 0.203 & 0.301 & 0.197 & 0.290&0.220&0.320&0.246&0.355&0.254&0.361&0.222&0.321&0.233&0.345&0.265&0.360&0.373&0.439&0.340&0.420\\
& Avg &0.167&0.263&\color{red}\textbf{0.166}&\color{red}\textbf{0.259}&\color{red}\textbf{0.166}&0.263&\textbf{0.161}&\textbf{0.252}&0.192&0.295&0.214&0.327&0.227&0.338&0.193&0.296&0.208&0.323&0.229&0.329&0.311&0.397&0.338&0.422\\
\midrule

\multirow{5}{*}{\rotatebox{90}{$Traffic$}}
& 96  & 0.388 & 0.282 & 0.390 & 0.272 & 0.410 & 0.282 & 0.360 & 0.249&0.593&0.321&0.587&0.366&0.613&0.388&0.612&0.338&0.607&0.392&0.615&0.391&0.719&0.391&0.732&0.423\\
& 192 & 0.407 & 0.290 &0.403 & 0.276 & 0.423 & 0.287 &0.379 & 0.256&0.617&0.336&0.604&0.373&0.616&0.382&0.613&0.340&0.621&0.399&0.601&0.382&0.696&0.379&0.733&0.420\\
& 336 & 0.412 & 0.294 & 0.413 & 0.280 & 0.436 & 0.296 & 0.392 & 0.264&0.629&0.336&0.621&0.383&0.622&0.337&0.618&0.328&0.622&0.396&0.613&0.386&0.777&0.420&0.742&0.420 \\
& 720 & 0.450 & 0.312 & 0.447 & 0.298 & 0.466 & 0.315 & 0.432 & 0.286&0.640&0.350&0.626&0.382&0.660&0.408&0.653&0.355&0.632&0.396&0.658&0.407&0.864&0.472&0.755&0.423\\
& Avg &0.414&0.294&\color{red}\textbf{0.413}&\color{red}\textbf{0.281}&0.433&0.295&\textbf{0.390}&\textbf{0.263}&0.620&0.336&0.610&0.376&0.628&0.379&0.624&0.340&0.621&0.396&0.622&0.392&0.764&0.416&0.741&0.422\\
\midrule
\multicolumn{2}{c|}{Average}&\color{red}\textbf{0.516}&\color{red}\textbf{0.407}&0.573&0.0.431&0.562&0.436&\textbf{0.446}&\textbf{0.386}&0.596&0.433&0.701&0.489&0.757&0.511&0.633&0.465&0.662&0.473&1.303&0.616&1.836&0.871&2.081&0.954\\

\bottomrule
\end{tabular}
}
\end{small}
\end{center}
\vskip -0.1in
\end{table*}
\clearpage

\subsection{Mean and STD for Few-shot Learning}
\label{app:detailed_res_few-shot}

Table \ref{tab:std_results} lists both mean and STD for GPT2(6), DLinear and PatchTST with 3 runs on 5\%  ETTh2 and ETTm2. The results show a small variance in performance of GPT2(6) that represents the stability of GPT2(6).
\begin{table}[h]
\caption{A subset of results showing both Mean and STD on 5\% datasets.}
\label{tab:std_results}
\vskip 0.15in
\begin{center}
\begin{small}
\scalebox{0.8}{
\begin{tabular}{c|c|cc}
\toprule

\multicolumn{2}{c|}{Methods}&\multicolumn{2}{c}{GPT2-backbone(6 Layers)} \\ % &\multicolumn{2}{c|}{DLinear}&\multicolumn{2}{c}{PatchTST}\\

% \midrule

\multicolumn{2}{c|}{Metric} & MSE  & MAE \\ % & MSE & MAE& MSE  & MAE\\
\midrule

\multirow{4}{*}{\rotatebox{90}{$ETTh2$}}
& 96  & 0.376 $\pm$ 0.0072 & 0.421 $\pm$ 0.0054 \\ % & 0.442 $\pm$ 0.0089 & 0.456 $\pm$ 0.0071 & 0.401 $\pm$ 0.0044 & 0.421 $\pm$ 0.0031 \\
& 192 & 0.418 $\pm$ 0.0013 & 0.441 $\pm$ 0.0014 \\ % & 0.617 $\pm$ 0.0045 & 0.542 $\pm$ 0.0019 & 0.452 $\pm$ 0.0028 & 0.455 $\pm$ 0.0017 \\
& 336 & 0.408 $\pm$ 0.0006 & 0.439 $\pm$ 0.0002 \\ % & 1.426 $\pm$ 0.0442 & 0.849 $\pm$ 0.0175 & 0.464 $\pm$ 0.0018 & 0.469 $\pm$ 0.0011 \\
& 720 & - & - \\ % & - & - & - & - \\
\midrule

\multirow{4}{*}{\rotatebox{90}{$ETTm2$}}
& 96  & 0.199 $\pm$ 0.0040 & 0.280 $\pm$ 0.0042 \\ % & 0.236 $\pm$ 0.0024 & 0.326 $\pm$ 0.0025 & 0.206 $\pm$ 0.0011 & 0.288 $\pm$ 0.0009 \\
& 192 & 0.256 $\pm$ 0.0030 & 0.316 $\pm$ 0.0017 \\ % & 0.306 $\pm$ 0.0049 & 0.373 $\pm$ 0.0047 & 0.264 $\pm$ 0.0013 & 0.324 $\pm$ 0.0011 \\
& 336 & 0.318 $\pm$ 0.0046 & 0.353 $\pm$ 0.0032 \\ % & 0.380 $\pm$ 0.0025 & 0.423 $\pm$ 0.0017 & 0.334 $\pm$ 0.0029 & 0.367 $\pm$ 0.0017 \\
& 720 & 0.460 $\pm$ 0.0132 & 0.436 $\pm$ 0.0066 \\ % & 0.674 $\pm$ 0.0005 & 0.583 $\pm$ 0.0003 & 0.454 $\pm$ 0.0030 & 0.432 $\pm$ 0.0015 \\

\bottomrule
\end{tabular}
}
\end{small}
\end{center}
\vskip -0.1in
\end{table}

\subsection{Comparison with Traditional Methods on Few-shot Learning}
\label{app:classical_few_shot}
Since deep learning methods are more advantageous than traditional methods when applied to large datasets. For few-shot learning, traditional methods should also consider. The results are shown in Table \ref{tab:classical_methods} that GPT2(6) also achieves best performance.
\begin{table}[h]
\caption{Comparison with traditional methods.}
\label{tab:classical_methods}
\begin{center}
\begin{small}
\scalebox{0.8}{
\begin{tabular}{c|c|cc|cc|cc|cc|cc}
\toprule

\multicolumn{2}{c|}{Methods}&\multicolumn{2}{c|}{GPT2(6) 5\%}&\multicolumn{2}{c|}{GPT2(6) 10\%}&\multicolumn{2}{c|}{ETS}&\multicolumn{2}{c|}{ARIMA}&\multicolumn{2}{c}{NaiveDrift}\\

\multicolumn{2}{c|}{Metric} & MSE  & MAE & MSE & MAE& MSE  & MAE& MSE  & MAE& MSE  & MAE\\

\midrule

\multirow{2}{*}{\rotatebox{90}{\tiny $ETTh2$}}
& 96  & 0.376 & 0.421 & 0.331 & 0.374 & 2.954  & 0.742 & 0.481 & 0.443 & 0.764 & 0.561 \\
& 192 & 0.418 & 0.441 & 0.402 & 0.411 & 10.226 & 1.212 & 0.585 & 0.495 & 1.560 & 0.785 \\
\midrule

\multirow{2}{*}{\rotatebox{90}{\tiny $ETTm1$}}
& 96  & 0.386 & 0.405 & 0.390 & 0.404 & 52.237 & 2.689 & 0.693 & 0.547 & 1.539  & 0.913 \\
& 192 & 0.440 & 0.438 & 0.429 & 0.423 & 186.445 & 4.654 & 0.710 & 0.557 & 2.869 & 1.215 \\

% \multirow{4}{*}{\rotatebox{90}{$Traffic$}}
% & 96  & 0.419 & 0.298 & 0.468 & 0.354 & 0.427 & 0.304 & \textbf{0.404} & \textbf{0.286} & 0.670 & 0.421 & 0.795 & 0.481 \\
% & 192 & 0.434 & 0.305 & 0.479 & 0.352 & 0.447 & 0.315 & \textbf{0.412} & \textbf{0.294} & 0.653 & 0.405 & 0.837 & 0.503 \\
% & 336 & 0.449 & 0.313 & 0.477 & 0.345 & 0.478 & 0.333 & \textbf{0.439} & \textbf{0.310} & 0.707 & 0.445 & 0.867 & 0.523 \\
% & 720 & - & - & - & - & - & - & - & - & - & - & - & - \\
% \hline
\bottomrule
\end{tabular}
}
\end{small}
\end{center}
\end{table}

% \subsection{Zero-shot Transfer Learning}

% We investigate the representation ability of GPT-backbone (6 Layers) FPT in zero-shot transfer learning tasks and follow the experiments settings of meta-learning based N-BEATS \cite{oreshkin2021meta}. We conduct experiments on several univariate datasets, including M4 \cite{makridakis2020m4}, M3 \cite{makridakis2000m3}, Tourism \cite{athanasopoulos2011tourism}, Electricity \cite{asuncion2007uci}. All these datasets are obtained from \cite{godahewa2021monash}.

\subsection{Baselines with Instance Normalization}
Instance normalization \cite{kim2022reversible} is a plug-in for time series for distribution shift. Most baselines, such as Autoformer and FEDformer are not equipped with instance normalization. Thus, for a fair comparison, we add the experiment, as in Table \ref{tab:revin_baseline}, for baselines w/o instance normalization and GPT(6) can also perform superior.

\begin{table}[h]
\caption{Comparison on 5\% data. Autoformer and FEDformer are equiped with instance normalization.}
\label{tab:revin_baseline}
\begin{center}
\begin{small}
\scalebox{0.7}{
\begin{tabular}{c|c|cc|cc|cc|cc|cc|cc|cc}
\toprule

\multicolumn{2}{c|}{Methods}&\multicolumn{2}{c|}{GPT2(6)}&\multicolumn{2}{c|}{PatchTST}&\multicolumn{2}{c|}{DLinear}&\multicolumn{2}{c|}{Autoformer}&\multicolumn{2}{c|}{Autoformer(Revin)}&\multicolumn{2}{c|}{FEDformer}&\multicolumn{2}{c}{FEDformer(Revin)}\\

\multicolumn{2}{c|}{Metric} & MSE  & MAE & MSE & MAE& MSE  & MAE& MSE  & MAE& MSE  & MAE& MSE  & MAE& MSE  & MAE\\

\midrule

\multirow{2}{*}{\rotatebox{90}{\tiny $ETTm2$}}
& 96  & 0.199	&0.280	&0.206&	0.288&	0.236	&0.326&	0.232	&0.322	&0.224	&0.300	&0.229&	0.320&	0.223	&0.298 \\
& 192 & 0.256	&0.316	&0.264&	0.324&	0.306&	0.373&	0.291	&0.357&	0.296&	0.343&	0.294&	0.361	&0.288	&0.336 \\

% \multirow{4}{*}{\rotatebox{90}{$Traffic$}}
% & 96  & 0.419 & 0.298 & 0.468 & 0.354 & 0.427 & 0.304 & \textbf{0.404} & \textbf{0.286} & 0.670 & 0.421 & 0.795 & 0.481 \\
% & 192 & 0.434 & 0.305 & 0.479 & 0.352 & 0.447 & 0.315 & \textbf{0.412} & \textbf{0.294} & 0.653 & 0.405 & 0.837 & 0.503 \\
% & 336 & 0.449 & 0.313 & 0.477 & 0.345 & 0.478 & 0.333 & \textbf{0.439} & \textbf{0.310} & 0.707 & 0.445 & 0.867 & 0.523 \\
% & 720 & - & - & - & - & - & - & - & - & - & - & - & - \\
% \hline
\bottomrule
\end{tabular}
}
\end{small}
\end{center}
\end{table}

\subsection{Detailed Definition and Results of Zero-shot Learning}
\label{app:zero-shot-full}
\textbf{Task Definition}
Each experiment contains two distinct datasets, source, and target datasets. The source dataset is used to train the model and then forecasts without fine-tuning in the target dataset. The target dataset is split into non-overlapping historical and test sequences. We use the historical sequence as input to the model, and the obtained output is used to calculate errors with the test sequences. Besides meta-learning-based models like N-BEATS, evaluated models' parameters are not allowed any adjustment using the forecasting phase. Also, same as \cite{oreshkin2021meta}, each data set adopts a specific metric (M4: sMAPE; M3: sMAPE; TOURISM: MAPE; ELECTR: ND)

\textbf{Detailed Results}
\label{app:detailed_res_zero-shot}
Here, we list detailed performance of zero-shot learning in Table \ref{tab:zero_m4}, Table \ref{tab:zero_m3} and Table \ref{tab:zero_tourism}. For each dataset, we separately list the performance of models under diverse frequency. Compared to the most recent published method DLinear, GPT2(6) performs superior in most situations. Also, GPT2(6) does not use any information from the test data, but achieves a comparable performance of meta-leaning based N-BEATS.

% tab moash zeroshot on m4
\begin{table*}[h]
\caption{Zero-shot performance on M4 (sMAPE).}
\vskip 0.15in
\label{tab:zero_m4}
\begin{center}
\begin{small}
\scalebox{1}{
\begin{tabular}{c|ccccc}
\toprule

 & Yearly & Quarterly & Monthly & Others & Average \\
 & (23k) & (24k) & (48k) & (5k) & (100k) \\

\midrule

% DeepAR & 12.362 & 10.822 & 13.705 & 4.668 & 12.253 \\
% N-BEATS & 12.913 & 9.213 & 12.024 & 3.643 & 11.135 \\
% \midrule
N-BEATS-FR & 13.267 & 9.596 & 12.676 & 4.696 & 11.675 \\
\midrule
DLinear-M3 & 14.193 & 18.856 & 14.765 & 9.194 & 15.337 \\
TimesNet-M3 & 15.655 & 11.877 & 16.165 & 6.863 & 14.553 \\
PatchTST-M3 & 13.966 & 10.929 & 14.664 & 7.087 & 13.228 \\
ETSformer-M3 & 27.846 & 36.134 & 25.114 & 12.338 & 27.748 \\
LightTS-M3 & 13.787 & 11.289 & 15.181 & 9.117 & 13.623 \\
Stationary-M3 & 14.988 & 11.686 & 16.098 & 6.977 & 14.327 \\
FEDformer-M3 & 13.887 & 11.513 & 18.154 & 7.529 & 15.047 \\
Autoformer-M3 & 14.552 & 17.341 & 25.063 & 9.666 & 20.022 \\
Informer-M3 & 18.542 & 16.907 & 23.454 & 7.348 & 19.047 \\
Reformer-M3 & 15.652 & 11.051 & 15.604 & 7.001 & 14.092 \\
\midrule
GPT(6)-M3 & 13.740 & 10.787 & 14.630 & 7.081 & 13.125 \\

\bottomrule
\end{tabular}
}
\end{small}
\end{center}
\vskip -0.1in
\end{table*}

% tab moash zeroshot on m3
\begin{table*}[h]
\caption{Zero-shot performance on M3 (sMAPE).}
\vskip 0.15in
\label{tab:zero_m3}
\begin{center}
\begin{small}
\scalebox{1}{
\begin{tabular}{c|ccccc}
\toprule

 & Yearly & Quarterly & Monthly & Others & Average \\
 & (645) & (756) & (1428) & (174) & (3003) \\

\midrule

% DeepAR & 13.33 & 9.07 & 13.72 & 7.11 & 12.67 \\
% N-BEATS & 15.93 & 8.84 & 13.11 & 4.24 & 12.37 \\
% \midrule
% DeepAR-M4 & 14.76 & 9.28 & 16.15 & 13.09 & 14.76 \\
N-BEATS-M4 & 15.07 & 9.07 & 13.19 & 4.29 & 12.38 \\
N-BEATS-FR & 16.43 & 9.05 & 13.30 & 4.51 & 12.61 \\
\midrule
DLinear-M4 & 17.43 & 9.74 & 15.65 & 6.81 & 14.03 \\
TimesNet-M4 & 18.75 & 12.26 & 14.01 & 6.88 & 14.17 \\
PatchTST-M4 &  15.99 & 9.62 & 14.71 & 9.44 & 13.39 \\
ETSformer-M4 & 20.56 & 11.65 & 16.97 & 10.57 & 16.03 \\
LightTS-M4 & 15.63 & 9.40 & 24.60 & 8.28 & 17.90 \\
Stationary-M4 & 17.05 & 12.56 & 16.82 & 8.13 & 15.29 \\
FEDformer-M4 & 16.00 & 9.48 & 15.12 & 8.94 & 13.53 \\
Autoformer-M4 & 16.18 & 13.92 & 16.91 & 14.68 & 15.87 \\
Informer-M4 & 19.70 & 13.00 & 15.91 & 13.03 & 15.82 \\
Reformer-M4 & 16.03 & 9.76 & 14.80 & 7.53 & 13.37 \\
\midrule
GPT2(6)-M4 & 16.42 & 10.13 & 14.10 & 4.81 & 13.06 \\

\bottomrule
\end{tabular}
}
\end{small}
\end{center}
\vskip -0.1in
\end{table*}

% tab moash zeroshot on tourism
\begin{table*}[!h]
\caption{Zero-shot performance on Tourism (MAPE).}
\vskip 0.15in
\label{tab:zero_tourism}
\begin{center}
\begin{small}

\scalebox{1}{
\begin{tabular}{c|ccccc}
\toprule

 & Yearly & Quarterly & Monthly & Average \\
 & (518) & (427) & (366) & (1311) \\

\midrule

% DeepAR & 21.14 & 15.82 & 20.18 & 19.27 \\
% N-BEATS & 21.44 & 14.78 & 19.29 & 18.52 \\
% \midrule
% DeepAR-M4 & 21.51 & 22.01 & 26.64 & 24.79 \\
N-BEATS-M4 & 23.57 & 14.66 & 19.32 & 18.82 \\
N-BEATS-FR & 23.43 & 14.45 & 20.47 & 19.46 \\
\midrule
DLinear-M4 & 39.59 & 18.30 & 24.76 & 28.51\\
% DLinear-M3 & 38.20 & 17.21 & 27.76 & 28.45 \\
TimesNet-M4 & 35.59 & 19.22 & 30.54 & 28.84 \\
PatchTST-M4 & 33.23 & 19.27 & 27.57 & 27.10 \\
ETSformer-M4 & 391.60 & 35.56 & 50.47 & 180.40 \\
LightTS-M4 & 138.22 & 16.28 & 25.34 & 66.99 \\
Stationary-M4 & 35.42 & 35.15 & 65.58 & 43.75 \\
FEDformer-M4 & 43.41 & 19.88 & 28.39 & 31.55 \\
Autoformer-M4 & 51.19 & 34.95 & 31.47 & 40.39 \\
Informer-M4 & 41.16 & 30.98 & 33.92 & 35.82 \\
Reformer-M4 & 33.86 & 16.85 & 23.71 & 25.48 \\
\midrule
GPT2(6)-M4 & 27.17 & 16.21 & 21.92 & 22.14\\
% GPT2-FPT-M3 & 27.46 & 17.66 & 30.03 & 24.99 \\

\bottomrule
\end{tabular}
}
\end{small}
\end{center}
\vskip -0.1in
\end{table*}

% results on 10% monash data
% \input{tables/tab_monash_10percent.tex}

% results on 5% monash data
% \input{tables/tab_monash_5percent.tex}

% \section{Supplemental Experiments}

% \subsection{Ablation Study on Finetune Parameters}
% \label{subsection:appendix_ablation}
% Ablation study on which part of the parameters to finetune is shown in Table \ref{tab:ablation}. We successively add parameters of various blocks and conduct experiments on 5\% ETTm1 and ETTm2. We observe that the benefits of finetune layer normalization and positional embeddings are relatively small, especially for shorter prediction length. Thus, although freeze all parameters of GPT2 also works, we simply follow the standard finetuning protocol of pretrained model that we still retrain the layer normalization and postional embeddings.

% \input{tables/tab_ett_ablation_ln_pos.tex}

\section{Proof}\label{sec:proof}

\begin{figure}[t]
    \centering
    \includegraphics[width=1\columnwidth]{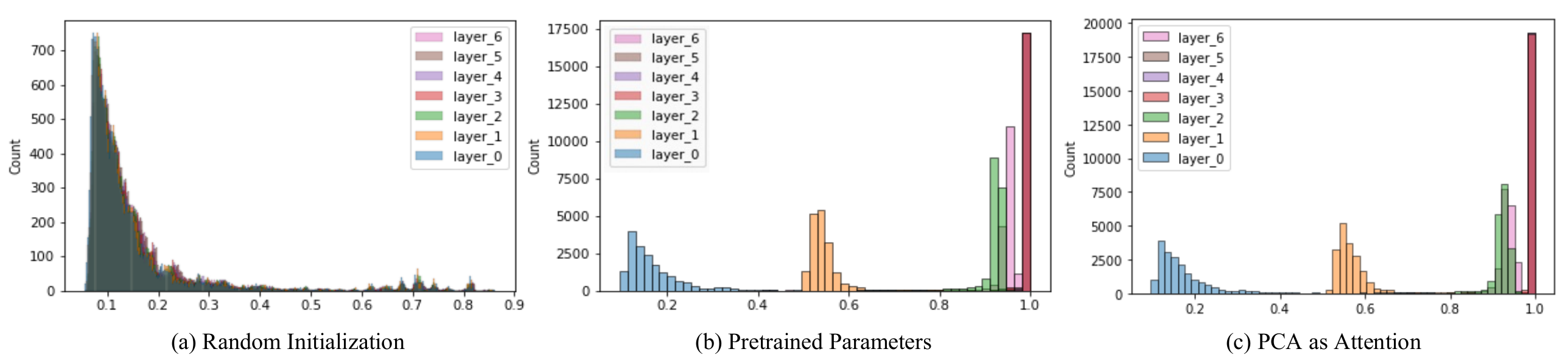}
    \caption{The token similarity within samples with respect to each layer. (a) GPT2-noPretrain-model; (b) GPT2-Pretrained-model; (c) Pretrained attention is replaced by PCA.}
    \label{fig:pca}
\end{figure}

In our numerical experiments, we obtain two interesting observations. First, the token similarity within a sample is larger in pretrained LM. We report the layer-wise average token cosine similarity in ETTh2 experiment in Figure~\ref{fig:pca}. In particular, Figure~\ref{fig:pca} (a) shows that in a fine-tuned random initialed GPT2(6) model, the token similarity is around 0.1-0.2 among different layers. When switching to the frozen pre-trained GPT2-FPT model, the token similarity significantly increases in the deep layers and eventually reaches more than 0.9 in the last layer. The ETTh2 dataset contains high volatility hourly information related to the electricity transformer temperature. In this situation, higher token similarity implies the high-frequency noise in the data is eased and only low-frequency information will be reserved. In other words, after going through the pretrained GPT2-FPT model, the signal-noise ratio is enhanced. We use the following theorem to characterize this behavior.

\subsection{Theorem \ref{thm:1}}

\begin{theorem}[informal]\label{thm:1}
We consider the self-attention for $l$-th query token. Let’s
assume the input token $\bm{x}_i$ are bounded with mean $\bm{\mu}$ for $i = 1, 2, ..., n$. Under mild conditions, with high probability, the output value token $\bm{V}_l$ converges to $\bm{\mu}W_v$ on the order of $\mathcal{O}(n^{-1/2})$, where $W_v$ is the parameter matrix to compute the value token.
\end{theorem}
The Theorem \ref{thm:1} describes the self-attention structure can efficiently make output value token $\bm{V}_l$ converge its mean value $\bm{\mu}W_v$. In the time series forecasting task, each token represents several adjacent points in a time series. When the time series has some periodical or translation invariant structures, by comparing a given token with other tokens, one could have a higher chance to figure out those invariant structures. This phenomenon is especially important in few-shot forecasting tasks. Without enough token noise distillation ability, the model will more likely tend to overfit due to insufficient training data.

We denote $x_i$ as $i$-th element of vector $\bm{x}$, $\bm{W}_{ij}$ as the element at $i$-th row and $j$-th column of matrix $\bm{W}$, and $\bm{W}_j$ as the $j$-th row of matrix $\bm{W}$. Moreover, we denote $\bm{x}_i$ as the $i$-th patch (token) of the inputs with $\bm{x}_i = \bm{X}_i$.\\

Before given the formal statement of the Theorem \ref{thm:1}, we first show the assumptions.

% {\bf Assumption 1}
\begin{enumerate}
    \item The token $\bm{x}_i$ is the sub-gaussian random vector with mean $\bm{\mu}_i$ and variance $(\sigma^2/d) I$ for $i=1,2,...,n$.% We assume there exists $x_{\max}>0$ such that $\|\bm{x}\|\le x_{\max}$.
    \item $\bm{\mu}$ follows a discrete distribution with finite values $\bm{\mu}\in \mathcal{V}$. Moreover, there exist $0<\nu_{1},0<\nu_{2}<\nu_4$ such that a) $\|\bm{\mu}_i\| = \nu_1$, and b) $\bm{\mu}_i\bm{W}_{\bm{Q}}\bm{W}_{\bm{K}}^T\bm{\mu}_i\in [\nu_2,\nu_4]$  for all $i$ and $| \bm{\mu}_i\bm{W}_{\bm{Q}}\bm{W}_{\bm{K}}^{\top}\bm{\mu}_j^{\top}|\le \nu_2$ for all $\bm{\mu}_{i}\ne \bm{\mu}_{j}\in\mathcal{V}$. \item $\bm{W}_V$ and $\bm{W}_{\bm{Q}}\bm{W}_{\bm{K}}^{\top}$ are element-wise bounded with $\nu_5$ and $\nu_6$ respectively, that is, $|\bm{W}_V^{(ij)}|\le \nu_5$ and $|(\bm{W}_{\bm{Q}}\bm{W}_{\bm{K}}^{\top})^{(ij)}|\le \nu_6$, for all $i,j$ from 1 to $d$.
\end{enumerate}
In the above assumptions, we ensure that for a given query patch, the difference between the clustering center and noises are large enough to be distinguished. 

% {\bf Assumption 4.2} 

% {\bf Assumption 4.3} 

% Let us consider the self-attention for query $i$ and we assume all token $\bm{x}_i$ are $\sigma^2_3$-subguassian random variable with $\bm{\mu}_i$. When there are $n_1$ tokens out of all $n$ tokens follows the same distribution of $i$-th token.  Under mild conditions, with probability $1-5\delta$, we have

\begin{theorem}[formal statement of Theorem \ref{thm:1}]
Let patch $\bm{x}_i$ be $\sigma^2$-subgaussian random variable with mean $\bm{\mu}_i$ and  all $n$ patches follow the same clustering center of query $l$. Per Assumptions aforementioned, when $\sqrt{d}\ge 3(\psi(\delta,d)+\nu_2+\nu_4)$, then with probability $1-5\delta$, we have

\begin{align}
&\left\|\frac{\sum_{i=1}^n\exp\left(\frac{1}{\sqrt{d}}\bm{x}_l\bm{W}_{\bm{Q}}\bm{W}^{\top}_k\bm{x}_i\right)\bm{x}_i\bm{W}_V}{\sum_{j=1}^n\exp\left(\frac{1}{\sqrt{d}}\bm{x}_l\bm{W}_{\bm{Q}}\bm{W}_{\bm{K}}^{\top}\bm{x}_j\right)}-\bm{\mu}_l\bm{W}_V\right\|_{\infty}\le 4\exp\left(\frac{\psi(\delta,d)}{\sqrt{d}}\right)\sigma\nu_5 \sqrt{\frac{2}{dn}\log\left(\frac{2d}{\delta}\right)}\notag\\
%&+\exp\left(\frac{\psi(\delta,d)}{\sqrt{d}}\right)\left[ 8\left[\exp\left(\frac{\nu_2-\nu_4}{\sqrt{d}}\right)-\frac{k_1}{k}\exp\left(-\frac{\psi(\delta,d)}{\sqrt{d}}\right)\right]+\left[1-\exp\left(\frac{\nu_2-\nu_4}{\sqrt{d}}\right)\right]\frac{k_1}{k}\right]\|\bm{\mu}_1\bm{W}_V\|_{\infty},
&+  7\left[\exp\left(\frac{\nu_2-\nu_4+\psi(\delta,d)}{\sqrt{d}}\right)-1\right]\|\bm{\mu}_l\bm{W}_V\|_{\infty}\notag,
%&+4\exp\left(\frac{\psi(\delta,d)}{\sqrt{d}}\right)\sigma\nu_5 \sqrt{\frac{2}{dk}\log\left(\frac{2d}{\delta}\right)},
\end{align}
where $\psi(\delta,d) = 2\sigma\nu_1\nu_6\sqrt{2\log\left(\frac{1}{\delta}\right)}+ 2\sigma^2\nu_6\log\left(\frac{d}{\delta}\right)$.\\
\end{theorem}
\begin{proof}
See the proof of Lemma 2 in \citet{wang2022kvt} with $k_1 = k = n$.
\end{proof}

\subsection{Theorem \ref{thm:2}}

We first give the formal statement of Theorem \ref{thm:2}.

\begin{theorem}[formal statement of Theorem \ref{thm:2}]
Let $\bm{g}_i\in\mathbbm{R}^{d}$ and $\bm{y}_i\in\mathbbm{R}^{T}$ be the feature map vector and forecasting targets for the sample $i=1,2,...,N$ respectively, and we assume $\frac{1}{N}\sum_{i=1}^N\bm{g}_i\bm{g}_i^{\top}\succeq\sigma I$ for some $\sigma >0$. We want to learn a matrix $\bm{W}\in\mathbbm{R}^{d\times T}$ from the following optimization problem:
\begin{align}
    \bm{W} = \arg\min \frac{1}{2N}\sum_{i=1}^N\|\bm{W}\bm{g}_i-\bm{y}_i\|_2^2.\label{eq:xue:1}
\end{align}
If we apply stochastic gradient descent with diminishing step sizes $\eta_t = \frac{1}{\sigma t}$ at step $t$, we will need $t = \tilde{\mathcal{O}}(\epsilon^{-1}\sigma^{-1})$ steps to reach
\begin{align}
  \frac{1}{t}\sum_{j=1}^t\left(\frac{1}{2N}\sum_{i=1}^N\|\bm{W}_{j}\bm{g}_i-\bm{y}_i\|_2^2 \right) -  \frac{1}{2N}\sum_{i=1}^N\|\bm{W}^{*}\bm{g}_i-\bm{y}_i\|_2^2 \le \epsilon,
\end{align}
where $\bm{W}^{*}$ is the optimal solution and $\bm{W}_{j}$ is the $j$ step's solution and $\tilde{\mathcal{O}}$ we suppress the logarithmic dependence.

% the total step to reach some optimization tolerance $\epsilon>0$ is on the order of $\mathcal{O}(\sigma^{-1}\epsilon^{-2})$.
\end{theorem}

\begin{proof}
As we assume $\frac{1}{N}\sum_{i=1}^T\bm{g_i}\bm{g}_i^{\top}\succeq \sigma I$, the hessian of optimization problem in \eqref{eq:xue:1} is also positive definite, which is equivalent to the optimization problem in \eqref{eq:xue:1} is strongly convex with parameter proportional to $\sigma$. Then via standard stochastic gradient decent analysis (e.g., section 3.1 in \citet{lacoste2012simpler}), we obtain:
\begin{align}
 \frac{1}{t}\sum_{j=1}^t\left(\frac{1}{2N}\sum_{i=1}^N\|\bm{W}_{j}\bm{g}_i-\bm{y}_i\|_2^2 \right) -  \frac{1}{2N}\sum_{i=1}^N\|\bm{W}^{*}\bm{g}_i-\bm{y}_i\|_2^2  \le \mathcal{O}\left(\frac{\log t}{\sigma t}\right) = \tilde{O}(\sigma^{-1}t^{-1}).
\end{align}
Therefore, to reach $\epsilon$ optimization gap, we just need to set $t = \tilde{\mathcal{O}}(\sigma^{-1}\epsilon^{-1})$.
\end{proof}

\begin{figure}[t]
    \centering
    \includegraphics[width=1\columnwidth]{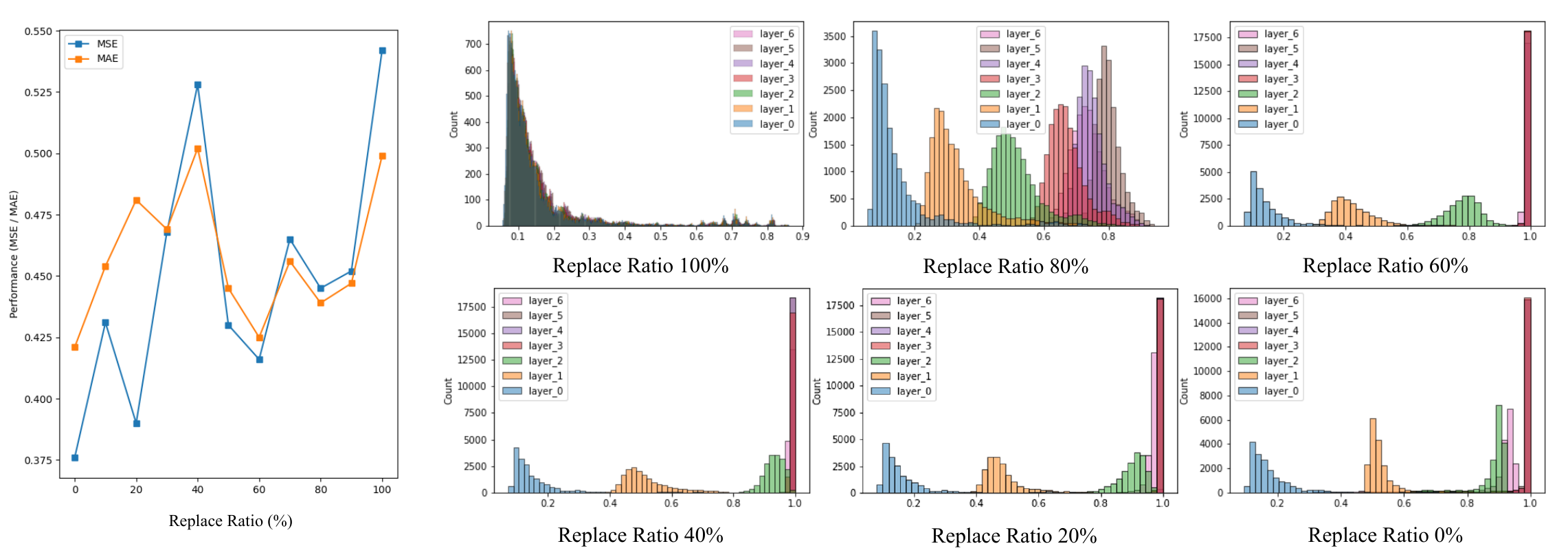}
    \caption{The performance and token similarity within samples with respect to each layer with different random replace ratios. Pretrained parameters are replaced by random initial parameters according to certain proportions.}
    \label{fig:replace_ratio}
\end{figure}

\begin{figure}[t]
    \centering
    \includegraphics[width=1\columnwidth]{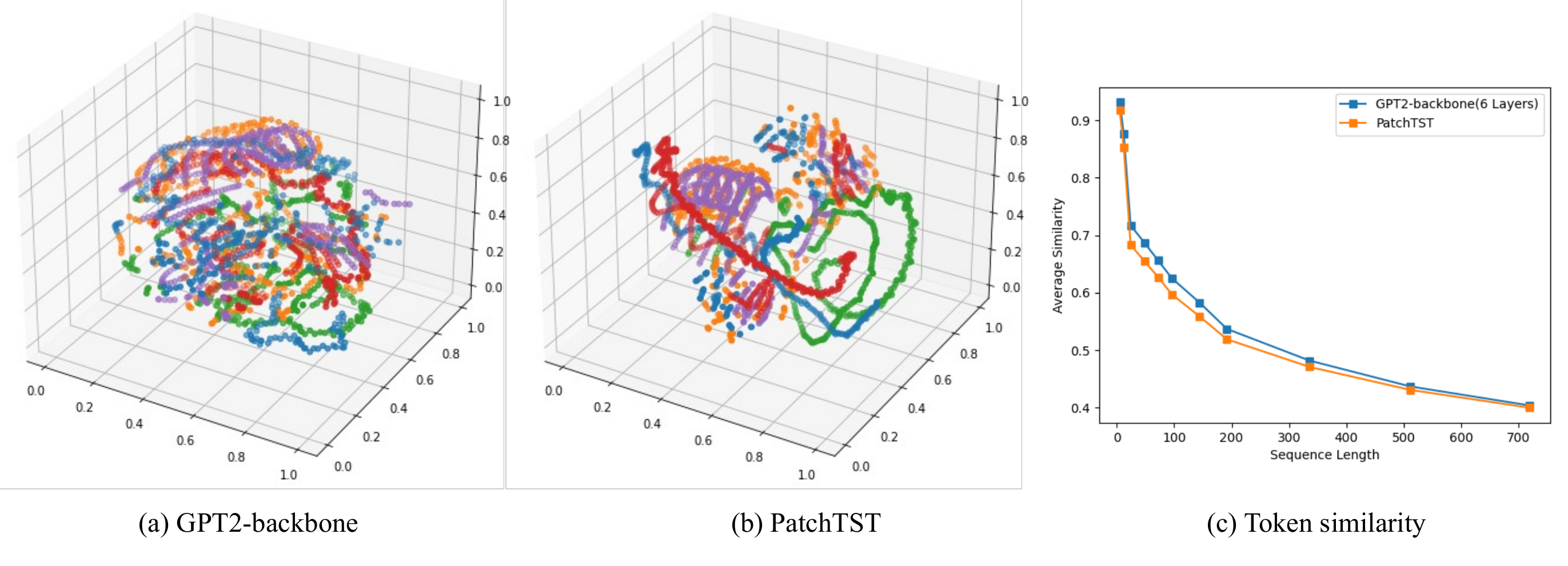}
    \caption{The t-SNE visualization of sample feature maps for (a) GPT-backbone, (b) end-to-end-PatchTST-model. (c) The token similarity within samples within different continuous sequence lengths.}
    \label{fig:tsne_metric}
\end{figure}

The second observation is that for the pretrained GPT2-FPT model, the last transformer layer's outputs, i.e., feature maps, are spread widely throughout the feature space. We report the t-SNE visualization of the feature maps for GPT2-FPT and an end-to-end model PatchTST in Figure~\ref{fig:tsne_metric}. In Figure~\ref{fig:tsne_metric} (a) and (b), we color the samples chunked from the one single time series into the same color and the same configuration of the T-SNE is applied. One may observe that the feature maps of GPT2-FPT has less concentration compared to PatchTST.
It implies the GPT2-FPT's feature maps corresponding to different samples are more distinctive which eventually facilitates the learning ability of the last MLP layer. Researchers \cite{Wang2020UnderstandingCR}  have found that contrastive learning-based representation learning may result in a uniform distribution of training data, and such behavior plays an important role in its good downstream task performance. We use the following theorem to justify it.
\begin{theorem}[informal]\label{thm:2}
Let $\bm{g}_i$ and $\bm{y}_i$ be the feature map vector and forecasting targets for the sample $i=1,2,...,N$ respectively, and we assume $\frac{1}{N}\sum_{i=1}^N\bm{g}_i\bm{g}_i^{\top}\succeq\sigma I$ for some $\sigma >0$. Under mild conditions, if we train an MLP layer that maps feature maps to forecasting targets via the stochastic gradient descent, the total step to reach some optimization tolerance is on the order of $\mathcal{O}(\sigma^{-1})$.
\end{theorem}

The Theorem~\ref{thm:2} considers the covariate matrix of feature maps being positive definite that indicates the set of all feature maps $\{\bm{g}_i\}$ spans the whole feature spaces, and the higher  spread level gives a larger $\sigma$. In this case, if we only want to learn an MLP layer, the problem reduces to a well-conditioned least-squared regression problem. Then the  fast convergence rate is achieved.

Efficiently learning the last MLP layer plays a very important role in time series forecasting and can substantially impact the prediction performance. In \citet{dlinear}, the authors show that learning a single MLP layer can also bring very promising performance. In few-shot forecasting, the pre-trained GPT2 model may still preserve highly diverse feature maps than end-to-end type models and eventually leads to fast learning speed on the last MLP layer.

Another possible benefit of wide spared feature maps is enhancing the model memorization ability when using a multi-layer decoder structure. In  the literature  on network memorization ability (e.g., \citet{vardi2021optimal, yun2020n}), the deep learning model tends to have better memorization ability when feature maps are well separated. In forecasting tasks, capturing extreme or rare behavior is very important. The pretrained GPT gains more capacity in the decoder to correctly forecast uncommon time series.

\section{N-gram Explanation for Universality}
\label{appendix:ngram}
Why does the proposed pretrained-frozen-model work so effectively? We have achieved state-of-the-art performance in time series analysis using a language model that is mostly trained on natural language data. The answer lies in the universality of the frozen structure, which includes attention layers and Feed Forward layers.
We can represent images and time series forecasting tasks as an n-gram estimation problem, akin to text analysis, by employing a patching approach. This method treats subsequences of time series or image patches as individual tokens. Central to sequential prediction is the $n$-order Markov process, and a simple way to capture the $n$-order Markov process is $n$-gram language model. To predict next token $w_0$, we need to compute $p(w_0|w_1, \ldots, w_{n-1})$, which can be further computed as $p(w_0w_1\ldots w_{n-1})/p(w_1\ldots w_{n-1})$. Hence, the core of $n$-gram language model is to estimate the probability of observing a sequence of $n$ tokens. When $n$ is large, most of $n$ token sequences will not be observed from data, leading to the sparse data problem, a common challenge faced by $n$-gram language model. As a result, a large body of research in $n$-gram language model is focused on how to effectively estimate probability of having $n$-token sequences even when they are NOT observed from data. We hypothesize that the transformer model pretrained by GPT-2 essentially allows us to estimate $p(w_0w_1\ldots w_{n-1})$ from observations of significantly shorter token sequences. In this section, we will show that the function of estimating probabilities of longer sequences from observation of shorter sequences is universal and is independent from domain as long as data exhibit a skew distribution (e.g., follows a power law). We note that our work is closely related to the discussion presented in ~\cite{elhage2021mathematical,olsson2022context}, where the authors also connect the function of transformer to compute of $n$-grams. We however note that our key result is to show the universality in computing probability of longer sequences from observations of shorter sequences, which can't be found in any existing studies. Although the discussion is restricted to discrete tokens, it should be generalized to continuous signals as we can always quantize continuous signals into a finite number of discrete tokens, similar to what BEiT ~\cite{bao2022beit} did. 

To gain a better understanding, let's start by examining a "zero-layer" Transformer model. This model operates by taking a token, embedding it, and transforming it back to produce logits that predict the subsequent token. Because it cannot transfer information from other tokens, it relies solely on the current token to predict the next one. Consequently, the optimal behavior of this model is to closely resemble the \textbf{bigram} log-likelihood.

Then we move on to the so-called "attention-only" transformer, which doesn't have MLP layers. As discussed in a recent work \cite{elhage2021mathematical}, one-layer attention-only Transformers can be comprehended as a combination of a \textbf{bigram} model and multiple \textbf{"skip-trigram"} models (impacting the probabilities of sequences "A… BC"). This can be intuitively understood as each attention head having the ability to selectively attend from the current token ("B") to a previous token ("A") and transfer relevant information to fine-tune the probability of potential subsequent tokens ("C"). \cite{olsson2022context} further discusses a multi-layer transformer can do more complex n-gram estimation using an induction heads mechanism. To be more precise, induction heads employ a straightforward principle: the '[A][B] ... [A] → [B]' rule, which elevates the likelihood of generating the subsequent token 'B' given the current token 'A' if there is a fuzzy match of the AB bigram in the historical context. This rule seems to largely decouple A and B, which means they do not memorize a fixed table of n-gram statistics. The rule [A][B] … [A] → [B] applies regardless of what A and B are, which can abstract to new patterns.

Building upon these discussions, we are now prepared to substantiate the following argument: \textbf{For sequential data following a power law, there is a potentially universal solution to the final estimation of n-gram probabilities}. That's the reason behind the universality of pretrained LM's performance in cross-domain tasks. For simplicity, we assume that $n$ is so large that we are unable to observe any occurrence of $n$-gram from data, and we only observe the occurrence of $n'$-grams with $n' < n$. We denote by $s^n_i$ the $i$th unique $n$-gram, and by the notation $s^{n'}_j \in s^n_i$ if $n'$-gram $s^{n'}_j$ appears in $s^n_i$, the $i$th $n$-gram. Let $m_n$ be the number of unique $n$-grams. According to the maximum entropy model, our estimation of n-gram probabilities can be cast into the following optimization problem:

\scalebox{1.0}{
$\begin{array}{rcl}
\min\ \sum_{i=1}^{m_n} p(s^n_i)\log p(s^n_i)\quad \mbox{s. t.}\sum_{i: s^{n'}_j \in s^n_i} p(s^n_i) = \ph (s^{n'}_j)  \notag
\end{array}$ 
}
where $\ph(s^{n'}_j)$ represents the probability of observing pattern $s^{n'}_j$ from the data and $j\in [m_{n'}], n' \in [n-1]$.

For each constraint for $\ph(s_j^{n'})$, we introduce a Lagrangian dual variable $\lambda_{j}^{n'}$, and rewrite the optimization problem as follows:

\scalebox{1.0}{
$\begin{array}{rcl}
\min_{\lambda}\ \log\left(\sum_{i=1}^{m_n} \exp\left(\sum_{(n',j): s^{n'}_j \in s^n_i} \lambda_j^{n'}\right)\right) - \sum_{n'=1}^{n-1}\sum_{j=1}^{m_{n'}} \lambda_{j}^{n'} \ph(s^{n'}_j),\notag
\end{array}$ 
}

where n-gram probability $p(s^n_j)$ is given as $p(s^n_j) = \frac{1}{Z(\lambda)}\exp\left(\sum_{(n',j):s_j^{n'} \in s_i^n} \lambda^{n'}_j\right)$ and $Z(\lambda) = \sum_{i=1}^{m_n} \exp(\sum_{(n',j):s_j^{n'} \in s_i^n} \lambda^{n'}_j)$

In the case that all n-grams follow a power law, for each $n' \in [n-1]$, we divide $n'$-gram into two groups: the group $\V_{n'}$ includes the high frequency $n'$-gram and the group $\U_{n'}$ including the low frequency of $n'$-gram. For simplicity, we assume that the probability for all the high frequency $n'$-grams are roughly $\alpha_{n'} \in [0, 1]$ and the probability for all the low frequency $n'$-grams are roughly $\beta_{n'} \in [0, 1]$. By assuming that all the patterns in $\V_{n'}$ and $\U_{n'}$ share similar appearance frequency, we simplify the optimization problem by only introducing two dual variables for each $n'$-gram, i.e. $\lambda_a^{n'}$ for high-frequency patterns and $\lambda_b^{n'}$ for low-frequency patterns as follow
Using these notations, we have the optimization problem simplified as

\scalebox{1.0}{
$\begin{array}{rcl}
\min_{\lambda} \quad \log(\sum_{i=1}^{m_n} \exp(\sum_{n'=1}^{n-1}\sum_{j: s^{n'}_j \in s^n_i} \lambda_a^{n'}I(s_j^{n'} \in \V_{n'})\notag\\
+ \lambda_b^{n'} I(s_j^{n'} \in \U_{n'}))) - \sum_{n'=1}^{n-1}\left(\lambda_a^{n'} g_{n'} + \lambda_b^{n'} h_{n'}\right)\notag,
\end{array}$ 
}

where $g_{n'} = \sum_{s^{n'}_j \in \V_{n'}} \ph(s^{n'}_j)$ and $ \quad h_{n'} = \sum_{s^{n'}_j \in \U_{n'}} \ph(s^{n'}_j)$.

Furthermore, let $q^{n'}_a$ be the probability to observe a high frequency $n'$-gram appearing in any $n$-gram, and $q^{n'}_b$ be the probability to observe a low frequency $n'$-gram appearing in any $n$-gram, we have

\scalebox{1.0}{
$\begin{array}{rcl}
&\sum_{i=1}^{m_n} \exp(\sum_{n'=1}^{n-1}\sum_{j: s^{n'}_j \in s^n_i} \lambda_a^{n'}I(s_j^{n'} \in \V_{n'}) + \lambda_b^{n'} I(s_j^{n'} \in \U_{n'})) \notag\\
&= m_n\prod_{n'=1}^{n-1}(1 + q_a^{n'}\exp(\lambda_a^{n'}))(1 + q_b^{n'}\exp(\lambda_b^{n'})) + \mathcal{O}\left(\sqrt{m_n}\right)\notag.
\end{array}$ 
}

By skipping the term $\mathcal{O}(\sqrt{m_n})$, we further simplify the optimization problem as

\scalebox{1.0}{
$\begin{array}{rcl}
\min_{\lambda} \quad& \sum_{n'=1}^{n-1}\log\left(1 
+ q_a^{n'} \exp(\lambda_a^{n'})\right) - %\lambda_{a}^{n'} g_{n'}\notag\\
+& \sum_{n'=1}^{n-1}\log\left(1 + q_b^{n'}\exp(\lambda_b^{n'}\right) - \lambda_b^{n'} h_{n'}\notag,    
\end{array}$ 
}

which is equivalent to

\scalebox{1.0}{
$\begin{array}{rcl}
\lambda_{n'}^a &= \min_{\lambda} \log\left(1 + q_a^{n'}\exp(\lambda)\right) - \lambda g_n'\notag\\ 
\lambda_{n'}^b &= \min_{\lambda}\log\left(1 + q_b^{n'}\exp(\lambda)\right) - \lambda h_n'.\notag 
\end{array}$ 
}
As illustrated by the above analysis, dual variables $\lambda_a^{n'}$ and $\lambda_b^{n'}$ will only depend on statistics $q_a^{n'}$, $q_b^{n'}$, $g_{n'}$ and $h_{n'}$. They are independent from the detailed statistics $\ph(s^{n'}_j)$ and how each $n'$-gram appears in different $n$-gram. Thus, this simple analysis does indicate, to some degree, that the solution obtained from the maximum entropy model can be universal, as long as $n$-grams follow skewed distributions like power law. 

We informally demonstrate that transformer models utilize attention mechanisms to perform a sophisticated form of n-gram estimation, and the generation rule for such n-gram distributions could be universal. This is how universality is achieved in our proposed cross-domain knowledge transfer. However, we currently lack a concrete metric to evaluate the performance of knowledge transfer between different domains, which requires further investigation. Nonetheless, in our experimental study, we demonstrate that a transformer model (beit) \cite{bao2022beit} trained on images can perform well on cross-domain time series forecasting tasks.
\section{Connection between self-attention and Principle component analysis}
\label{appendix:pca}
\noindent{\bf Understand the Gradient Structure of Self-Attention}
% In ``The Lipschitz Constant of Self-Attention'', the authors analyzed the structure of gradient for self-attention. 

Let $X = (x_1, \ldots, x_N)^{\top} \in \R^{N\times D}$ be the input pattern, and let $f(X) = (f_1(X), \ldots, f_N(x))^{\top}:\R^{N\times D}\mapsto \R^{N\times D}$ be the function for self-attention, i.e. 

\scalebox{1.0}{
    $\begin{array}{rcl}
f_i(X) = \mbox{softmax}(XAX^{\top})X
\end{array}$ }

where $A = W_QW_K^{\top} \in \R^{D\times D}$. Let the Jacobian $J = \left[\frac{\partial f_i(X)}{\partial x_j}\right]_{i,j=1}^N$ represent the gradient $f(X)$ with respect to input pattern. The lemma below shows an important structure of $J$. 

\begin{lemma}
\scalebox{1.0}{
    $\begin{array}{rcl}
|J|_2 \leq |A|_2\sum_{i=1}^N \left(P_{i,i} + \frac{1}{2}\right)\left|x_i - \sum_{j=1}^N P_{i,j}x_j\right|^2 + \Delta
\end{array}$ }

where
\scalebox{1.0}{
    $\begin{array}{rcl}
\Delta = |A|_2 \sum_{i \neq j}^N P_{i,j}\left|x_j - \sum_{k = 1}^N P_{i,k}x_k\right|^2 + \frac{|A|_2}{2}\sum_{j=1}^N |x_i|^2
\end{array}$ } and
\scalebox{1.0}{
    $\begin{array}{rcl}
P_{i,j} = \frac{\exp(x_i^{\top}Ax_j)}{\sum_{k=1}^N \exp(x_i^{\top}Ax_k)}
\end{array}$ }
\end{lemma}

\begin{proof}
According to the analysis from the work, we have the gradient $J_{i,j} = \frac{\partial f_i(X)}{x_j}$ is given by
\scalebox{1.0}{
    $\begin{array}{rcl}
J_{i,j} = P_{i,j} I + X^{\top}Q^i\left(XA\delta_{i,j} + E_{j,i}XA^{\top}\right)
\end{array}$ }
where
\scalebox{1.0}{
    $\begin{array}{rcl}
Q^i = \mbox{diag}(P_{i,:}) - P_{i,:}P_{i,:}^{\top}
\end{array}$ }
Here $P_{i,:} \in \R^N_+$ represents the $i$-th row of matrix $P$. We thus have
% \begin{eqnarray*}
%     \scalebox{0.7}{
%         $\begin{array}{rcl}
% |J|_2 & \leq & \sum_{i,j=1}^N |J_{i,j}|_2 \\
% & \leq & \sum_{i,j = 1}^N P_{i,j} + \sum_{i=1}^N |X^{\top}Q^i X|_2|A|_2 + \sum_{i,j=1}^N |X^{\top}Q^iE_{j,i}X|_2|A|_2 \\
% & \leq & N + |A|_2\sum_{i=1}^N \left(\sum_{j=1}^N P_{i,j}|x_j|^2 - \left|\sum_{j=1}^NP_{i,j}x_j\right|^2\right) + |A|_2\sum_{i,j=1}^N |X^{\top}Q^ie_jx_i^{\top}| \\
% & \leq & N + |A|_2\sum_{i=1}^N \sum_{j=1}^N P_{i,j}\left|x_j - \sum_{k=1}^N P_{i,k}x_k\right|^2 + |A|_2\sum_{i,j=1}^N P_{i,j}\left|x_i^{\top}\left(x_j - X^{\top}P_{i,:}\right)\right| \\
% & \leq & |A|_2\sum_{i=1}^N \left(P_{i,i}+\frac{1}{2}\right)\left|x_i - X^{\top}P_{i,:}\right|^2 + \underbrace{N + |A|_2\sum_{i\neq j}^N P_{i,j}\left|x_j - X^{\top}P_{i,:}\right|^2 + \frac{|A|_2}{2}\sum_{j=1}^N|x_i|^2}_{:=\Delta}
%     \end{array}$}
% \end{eqnarray*}
% \begin{figure*}
%\begin{eqnarray*}

    \scalebox{0.7}{
        $\begin{array}{rcl}
        |J|_2 & \leq & \sum_{i,j=1}^N |J_{i,j}|_2 \nonumber \\
        & \leq & \sum_{i,j = 1}^N P_{i,j} + \sum_{i=1}^N |X^{\top}Q^i X|_2|A|_2 + \sum_{i,j=1}^N |X^{\top}Q^iE_{j,i}X|_2|A|_2 \nonumber \\
        & \leq & N + |A|_2\sum_{i=1}^N \left(\sum_{j=1}^N P_{i,j}|x_j|^2 - \left|\sum_{j=1}^NP_{i,j}x_j\right|^2\right) + |A|_2\sum_{i,j=1}^N |X^{\top}Q^ie_jx_i^{\top}| \nonumber \\
        & \leq & N + |A|_2\sum_{i=1}^N \sum_{j=1}^N P_{i,j}\left|x_j - \sum_{k=1}^N P_{i,k}x_k\right|^2 + |A|_2\sum_{i,j=1}^N P_{i,j}\left|x_i^{\top}\left(x_j - X^{\top}P_{i,:}\right)\right| \nonumber \\
        & \leq & |A|_2\sum_{i=1}^N \left(P_{i,i}+\frac{1}{2}\right)\left|x_i - X^{\top}P_{i,:}\right|^2 + \underbrace{N + |A|_2\sum_{i\neq j}^N P_{i,j}\left|x_j - X^{\top}P_{i,:}\right|^2 + \frac{|A|_2}{2}\sum_{j=1}^N|x_i|^2}_{:=\Delta} \nonumber
        \end{array}$
    }
%\end{eqnarray*}
% \end{figure*}
\end{proof}
As indicated by Lemma 1, one of the key components in the upper bound of Jacobian is $|x_i - \sum_{j=1}^N P_{i,j}x_j|^2$. Thus, through the optimization, we like to reduce the size of the gradient and therefore may prefer to reduce the quantity to $\sum_{i=1}^N |x_i - \sum_{j=1}^N P_{i,j}x_j|^2$. Hence, it will be interesting to understand the choice of $W^Q$ and $W^K$ that leads to the minimization of $\sum_{i=1}^N |x_i - \sum_{j=1}^N P_{i,j}x_j|^2$, i.e. the following optimization problem 
    \scalebox{1.0}{
    $\begin{array}{rcl}
    \min\limits_{|A|_F \leq \rho} \sum_{i=1}^N \left|x_i - \sum_{j=1}^N P_{i,j} x_j\right|^2 \label{eqn:opt-1}
    \end{array}$}
where $\rho$ is introduced to control the size of $A$. 

\noindent{\bf Connection between Self-Attention and Principal Component Analysis} 

Let consider the optimization problem in (\ref{eqn:opt-1}) when $\rho$ is small, we can approximate $P_{i,j}$ as 
    \scalebox{1.0}{
    $\begin{array}{rcl}
P_{i,j} \approx \frac{1}{N} + \frac{1}{N}x_i^{\top}Ax_j
\end{array}$}
Define $\bar{x} = X^{\top}\mathbf{1}/N$. We have 
    \scalebox{1.0}{
    $\begin{array}{rcl}
\sum_{i=1}^N|x_i - X^{\top}P_{i,:}|^2 = \sum_{i=1}^N\left|x_i - \bar{x} - X^{\top}XAx_i\right|^2
\end{array}$}
By assuming that all the input patterns are zero centralized, we have $\bar{x} = 0$ and 
    \scalebox{1.0}{
    $\begin{array}{rcl}
\sum_{i=1}^N|x_i - X^{\top}XAx_i|^2 = \mbox{tr}\left((I - X^{\top}XA)^2X^{\top}X\right)
\end{array}$}
The theorem below shows that $A$ minimizing the objective $\sum_{i=1}^N |x_i - X^{\top}XAx_i|^2$ contains the largest $m$ eigenvectors of $X^{\top}X$ where $m$ is the rank of $A$.

\begin{thm}
Let $W_Q$ and $W_K$ be matrices of size $D\times m$. Let $\lambda_1 \geq \lambda_2\geq ...\geq \lambda_D$ be the eigenvalues of $X^{\top}X$ ranked in descending order, and let $v_i \in \R^D, i=1, \ldots, D$ be the corresponding eigenvectors. The optimal solution $A^*$ that minimizes $\sum_{i=1}^N |x_i - X^{\top}XAx_i|^2$ is given by
\scalebox{1.0}{
    $\begin{array}{rcl}
    A = \sum_{i=1}^m \frac{1}{\lambda_i}v_iv_i^{\top}
\end{array}$}
\end{thm}
\begin{proof}
Since $W_Q, W_K \in \R^{D\times m}$ where $m < D$, we know that $A$ is a matrix of rank $m$. Hence, we know
\scalebox{1.0}{
    $\begin{array}{rcl}
\min\limits_{A} \sum_{i=1}^N|x_i - X^{\top}XAx_i|^2 \geq \sum_{k=m+1}^N \lambda_{k}
\end{array}$}
We also know that by choosing $A$ as
\scalebox{1.0}{
    $\begin{array}{rcl}
A = \sum_{i=1}^m\frac{1}{\lambda_i} v_iv_i^{\top}
\end{array}$} we have 

\scalebox{1.0}{
$\sum_{i=1}^N|x_i - X^{\top}XAx_i|^2 = \mbox{tr}\left(\left(I - \sum_{i=1}^m v_iv_i^{\top} \right)^2X^{\top}X\right) = \sum_{k=m+1}^D \lambda_k$
}

Hence, the solution $A$ for minimizing $\sum_{i=1}^N |x_i - X^{\top}XAx_i|^2$ is essential a weighted combination of top eigenvectors of $X^{\top}X$. Since a small gradient will prefer a small quantity of $\sum_{i=1}^N |x_i - X^{\top}XAx_i|^2$, by minimizing through the self-attention layer, we essentially choose weight matrix $W_Q$ and $W_K$ to be aligned with the principal directions of $X^{\top}X$.
\end{proof}
% \begin{figure}[t]
%     \centering
%     \includegraphics[width=0.6\columnwidth]{graphs/tsne_metric.png
%     }
%     \caption{The token similarity within samples within different continuous sequence length.}
%     \label{fig:tsne_metric}
% \end{figure}

% \begin{figure}[t]
%     \centering
%     \includegraphics[width=0.8\columnwidth]{graphs/model_structure_2.pdf}
%     \caption{Model Structure 2.}
%     \label{fig:structure_2}
% \end{figure}
\section{Experiment Analysis and Other Key Results}
\label{appendix:ablations}
\subsection{Experiment analysis of GPT2-FPT model}

In this section, we conduct experiments to analyze whether the self-attention frozen pre-trained model improves performance compared with overall fine-tuning and random initialization. Firstly, we compare GPT2(6) FPT with the same model without freezing (No Freeze) and random initial model (No Pre-train). For the end-to-end paradigm No Pre-train GPT2-backbone (6 Layers), we directly train all parameters of the model. We summarize the results in Table \ref{tab:5_percent_ablation} and Table \ref{tab:10_percent_ablation}. Then we analyze the performance of various layers to clarify our selection of GPT2(6) FPT.
\begin{table}[!h]
\caption{Model analysis results on 5\% data. We use prediction length $O \in \{96, 192, 336, 720\}$ for ILI and $O \in \{24, 36, 48, 60\}$ for others.}
% GPT2-FPT represents GPT2-backbone (6 Layers) forzen pretrained transformer model. No Freeze and No Pretrain respectively represent GPT2-backbone (6 Layers) without freeze or pretrain. A lower MSE indicates better performance, and the best results are highlighted in bold. '-' means that 10\% time series is not sufficient to constitute a training set.}
\label{sample-table}
% \vskip 0.15in
\begin{center}
\begin{small}
\scalebox{0.75}{
\begin{tabular}{c|c|cccccc}
\toprule

\multicolumn{2}{c|}{Methods}&\multicolumn{2}{c|}{GPT2(6)}&\multicolumn{2}{c|}{No Freeze}&\multicolumn{2}{c}{No Pretrain}\\

\midrule

\multicolumn{2}{c|}{Metric} & MSE  & MAE & MSE & MAE& MSE  & MAE \\
\midrule

% \multirow{4}{*}{\rotatebox{90}{$ECL$}}
% & 96  &  &  &  &  &  & \\
% & 192 &  &  &  &  &  & \\
% & 336 &  &  &  &  &  & \\
% & 720 &  &  &  &  &  & \\
% \midrule

\multirow{4}{*}{\rotatebox{90}{$Weather$}}
& 96 & \textbf{0.175} &	0.230 &	0.183 & \textbf{0.229} &	0.199 &	0.254 \\
& 192 & \textbf{0.227} & \textbf{0.276} &	0.275 &	0.300 &	0.262 &	0.302 \\
& 336 & \textbf{0.286} & \textbf{0.322} &	0.297 &	0.331 &	0.326 &	0.345 \\
& 720 & \textbf{0.366} & \textbf{0.379} &	0.380 &	0.388 &	0.405 &	0.396 \\
\midrule

\multirow{4}{*}{\rotatebox{90}{$ETTh1$}}
& 96 & \textbf{0.543} &	\textbf{0.506} &	0.671 &	0.564 &	0.882 &	0.643 \\
& 192 & \textbf{0.748} &	\textbf{0.580} &	0.907 &	0.632 &	1.389 &	0.817 \\
& 336 & \textbf{0.754} &	\textbf{0.595} &	0.931 & 0.655 &	2.968 &	1.149 \\
& 720 & - &	- &	- &	- &	- &	- \\
\midrule

\multirow{4}{*}{\rotatebox{90}{$ETTh2$}}
& 96 & \textbf{0.376} &	\textbf{0.421} &	0.440 &	0.449 &	0.465 &	0.457 \\
& 192 & \textbf{0.418} &	\textbf{0.441} &	0.503 &	0.478 &	0.614 &	0.536 \\
& 336 & \textbf{0.408} &	\textbf{0.439} &	0.691 &	0.572 &	0.596 &	0.529 \\
& 720 & - &	- &	- &	- &	- &	- \\
\midrule

\multirow{4}{*}{\rotatebox{90}{$ETTm1$}}
& 96 & \textbf{0.386} &	\textbf{0.405} &	0.429 &	0.432 &	0.394 &	0.410 \\
& 192 & 0.440 &	0.438 &	0.496 &	0.470 &	\textbf{0.432} &	\textbf{0.432} \\
& 336 & \textbf{0.485} &	\textbf{0.459} &	0.535 &	0.489 &	0.491 &	0.464 \\
& 720 & \textbf{0.557} &	\textbf{0.499} &	0.786 &	0.592 &	0.564 &	0.503 \\
\midrule

\multirow{4}{*}{\rotatebox{90}{$ETTm2$}}
& 96 & \textbf{0.199} &	\textbf{0.280} &	0.217 &	0.293 &	0.301 &	0.353 \\
& 192 & \textbf{0.256} &	\textbf{0.316} &	0.300 &	0.350 &	0.321 &	0.365 \\
& 336 & \textbf{0.318} &	\textbf{0.353} &	0.331 &	0.368 &	0.371 &	0.398 \\
& 720 & \textbf{0.460} &	0.439 &	\textbf{0.460} &	\textbf{0.436} &	0.659 &	0.528 \\
% \midrule

% \multirow{4}{*}{\rotatebox{90}{$ILI$}}
% & 24 & 4.225 &	1.588 &	\textbf{4.188} &	\textbf{1.575} &	4.861 &	1.624 \\
% & 36 & - &	- &	- &	- &	- &	- \\
% & 48 & - &	- &	- &	- &	- &	- \\
% & 60 & - &	- &	- &	- &	- &	- \\
% \midrule

% \multirow{4}{*}{\rotatebox{90}{$Traffic$}}
% & 96  &  &  &  &  &  & \\
% & 192 &  &  &  &  &  & \\
% & 336 &  &  &  &  &  & \\
% & 720 &  &  &  &  &  & \\

\bottomrule
\end{tabular}
}
\label{tab:5_percent_ablation}
\end{small}
\end{center}
\vskip -0.1in
\end{table}

% tab ett ablation 10percent
\begin{table}[!h]
\caption{No Pretrain and No Freeze results on 10\% data.  We use prediction length $O \in \{96, 192, 336, 720\}$ for ILI and $O \in \{24, 36, 48, 60\}$ for others. }

%GPT2-FPT represents GPT2-backbone (6 Layers) frozen pretrained transformer model. No Freeze and No Pretrain respectively represent GPT2-backbone (6 Layers) without freeze or pretrain. A lower MSE indicates better performance, and the best results are highlighted in bold. '-' means that 10\% time seires is not sufficient to constitute a single training data.
% \vskip 0.15in
\begin{center}
\begin{small}
\scalebox{0.75}{
\begin{tabular}{c|c|cccccc}
\toprule

\multicolumn{2}{c|}{Methods}&\multicolumn{2}{c|}{GPT2(6)}&\multicolumn{2}{c|}{No Freeze}&\multicolumn{2}{c}{No Pretrain}\\

\midrule

\multicolumn{2}{c|}{Metric} & MSE  & MAE & MSE & MAE& MSE  & MAE \\
\midrule

% \multirow{4}{*}{\rotatebox{90}{$ECL$}}
% & 96  &  &  &  &  &  & \\
% & 192 &  &  &  &  &  & \\
% & 336 &  &  &  &  &  & \\
% & 720 &  &  &  &  &  & \\
% \midrule

\multirow{4}{*}{\rotatebox{90}{$Weather$}}
& 96 & \textbf{0.163} &	\textbf{0.215} &	0.168 &	0.221 &	0.175 &	0.229 \\
& 192 & \textbf{0.210} &	\textbf{0.254} &	0.238 &	0.286 &	0.244 &	0.287 \\
& 336 & \textbf{0.256} &	\textbf{0.292} &	0.289 &	0.318 &	0.301 &	0.325 \\
& 720 & \textbf{0.321} &	\textbf{0.339} &	0.398 &	0.383 &	0.390 &	0.378 \\
\midrule

\multirow{4}{*}{\rotatebox{90}{$ETTh1$}}
& 96 & \textbf{0.458} &	\textbf{0.456} &	0.605 &	0.532 &	0.680 &	0.560 \\
& 192 & \textbf{0.570} &	\textbf{0.516} &	0.713 &	0.579 &	0.738 &	0.602 \\
& 336 & \textbf{0.608} &	\textbf{0.535} &	0.747 &	0.586 &	0.893 & 0.641 \\
& 720 & \textbf{0.725} &	\textbf{0.591} &	0.945 &	0.688 &	2.994 & 1.169 \\
\midrule

\multirow{4}{*}{\rotatebox{90}{$ETTh2$}}
& 96 & \textbf{0.331} &	\textbf{0.374} &	0.369 &	0.394 &	0.422 &	0.433 \\
& 192 & \textbf{0.402} &	\textbf{0.411} &	0.464 &	0.455 &	0.482 &	0.466 \\
& 336 & \textbf{0.406} &	\textbf{0.433} &	0.420 &	0.439 &	0.540 &	0.496 \\
& 720 & \textbf{0.449} &	\textbf{0.464} &	0.535 &	0.515 &	0.564 &	0.519 \\
\midrule

\multirow{4}{*}{\rotatebox{90}{$ETTm1$}}
& 96 & 0.390 &	0.404 &	0.429 &	0.430 &	\textbf{0.385} &	\textbf{0.401} \\
& 192 & 0.429 &	0.423 &	0.463 &	0.446 &	\textbf{0.426} &	\textbf{0.421} \\
& 336 & \textbf{0.469} &	\textbf{0.439} &	0.510 &	0.470 &	0.506 &	0.455 \\
& 720 & \textbf{0.569} &	\textbf{0.498} &	0.780 &	0.591 &	0.576 &	0.505 \\
\midrule

\multirow{4}{*}{\rotatebox{90}{$ETTm2$}}
& 96 & \textbf{0.188} &	\textbf{0.269} &	0.243 &	0.311 &	0.244 &	0.315 \\
& 192 & \textbf{0.251} &	\textbf{0.309} &	0.307 &	0.352 &	0.318 &	0.363 \\
& 336 & \textbf{0.307} &	\textbf{0.346} &	0.337 &	0.364 &	0.409 &	0.412 \\
& 720 & \textbf{0.426} &	\textbf{0.417} &	0.471 &	0.440 &	0.473 &	0.450 \\
% \midrule

% \multirow{4}{*}{\rotatebox{90}{$ILI$}}
% & 24 & \textbf{3.022} &	\textbf{1.247} &	3.231 &	1.266 &	3.375 &	1.281 \\
% & 36 & 3.854 &	1.453 &	3.761 &	1.401 &	\textbf{3.056} &	\textbf{1.277} \\
% & 48 & 4.603 &	1.571 &	4.539 &	1.556 &	\textbf{3.431} &	\textbf{1.366} \\
% & 60 & - &	- &	- &	- &	- &	- \\
% \midrule

% \multirow{4}{*}{\rotatebox{90}{$Traffic$}}
% & 96  &  &  &  &  &  & \\
% & 192 &  &  &  &  &  & \\
% & 336 &  &  &  &  &  & \\
% & 720 &  &  &  &  &  & \\

\bottomrule
\end{tabular}
}
\label{tab:10_percent_ablation}
\end{small}
\end{center}
\vskip -0.22in
\end{table}

\textbf{Fine-tune More Parameters}
Compared with fine-tuning all parameters, self-attention frozen pre-trained model GPT2(6) FPT achieves better performance on most datasets and yields an overall \textbf{12.7\%} relative MSE reduction on 5\% data and \textbf{11.5\%} relative MSE reduction on 10\% data. It verifies that frozen pre-trained attention layers are effective for time series forecasting.
% and we visualize how it works in Section \ref{subsection:attention_visualization}.

\textbf{Parameters Initialization}
Compared with the random initial model, self-attention frozen pre-trained model GPT2(6) FPT achieves better performance on most datasets and yields an overall \textbf{21.2\%} relative MSE reduction on 5\% data and \textbf{14.3\%} relative MSE reduction on 10\% data. It again suggests that a model pre-trained on cross-domain data can achieve significant performance improvement in time series forecasting.

\begin{figure}[h]
    \centering
    \includegraphics[width=0.9\columnwidth]{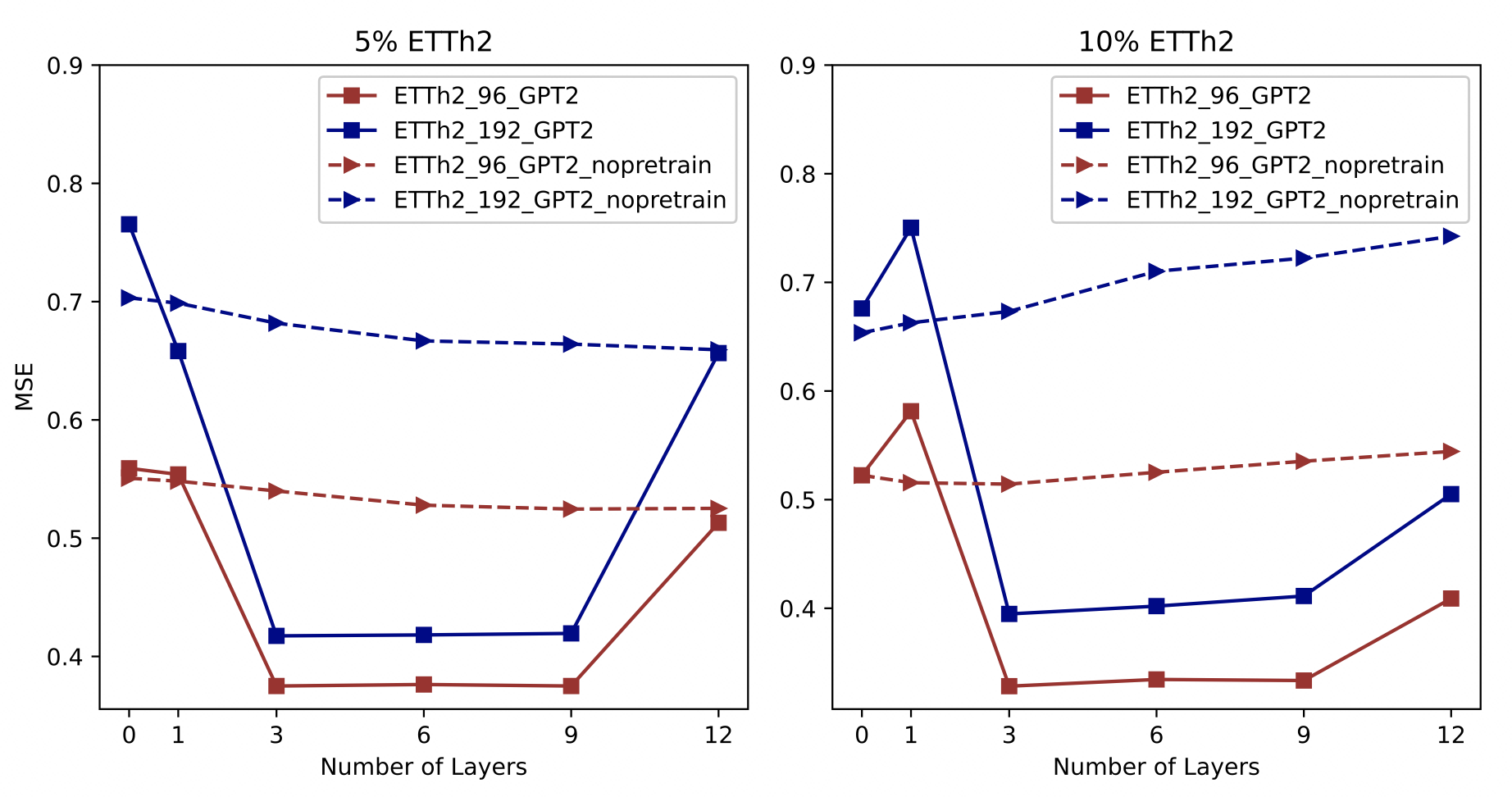}
    \caption{Comparison of pre-trained and non-pre-trained GPT2 with various layers on ETTh2. Color represents various prediction length $O \in \{96, 192\}$ and line style means different models
    \label{fig:different_layers}.}
\end{figure}

\textbf{The Number of GPT2 Layers}
For most transformer-based methods in time-series forecasting \cite{zhou2022fedformer,wu2021autoformer,Patchformer}, no more than 3 encoder layers are included. However, most pre-trained models with at least 12 layers may suffer from overfitting in time series forecasting. To better balance performance and computational efficiency, we test using various numbers of layers on ETTh2. Additionally, we train a completely random initialized non-pretrained GPT2 as a comparison. The results are shown in Figure \ref{fig:different_layers}, for both 5\% and 10\% data, the pre-trained model is unable to do well with few layers but significantly outperforms non-pre-trained GPT2 with more attention blocks transferred from NLP. It indicates that pre-trained attention layers produce a great benefit in time series forecasting. Also, the pre-trained model achieves better performance between 3 and 9 layers. Thus GPT2 with 6 layers is chosen as our default architecture.

\subsection{No Pre-training but Freezing}
\label{app:nopretrain_freeze}
For comprehensively ablation on pre-training and freezing strategies, we also add experiment for random initialized GPT2(6) with freezing. The results in Table \ref{tab:nopretrain_freeze} shows that only input and output modules can not work and pre-trained knowledge play an importance part in time series tasks.

\begin{table}[h]
\caption{Ablation on random initialized model with freezing.}
\label{tab:nopretrain_freeze}
\begin{center}
\begin{small}
\scalebox{0.7}{
\begin{tabular}{c|c|cc|cc|cc|cc}
\toprule

\multicolumn{2}{c|}{Methods}&\multicolumn{2}{c|}{GPT2(6)}&\multicolumn{2}{c|}{No Freeze}&\multicolumn{2}{c|}{No Pretrain}&\multicolumn{2}{c}{No Pretrain + Freeze}\\

\multicolumn{2}{c|}{Metric} & MSE  & MAE & MSE & MAE& MSE  & MAE& MSE  & MAE\\

\midrule

\multirow{2}{*}{\rotatebox{90}{\tiny $ETTh2$}}
& 96  & 0.376	&0.421&	0.440&	0.449	&0.465&	0.457&	0.540&	0.497\\
& 192 & 0.418&	0.441&	0.503&	0.478&	0.614&	0.536&	0.721&	0.580
\\

% \multirow{4}{*}{\rotatebox{90}{$Traffic$}}
% & 96  & 0.419 & 0.298 & 0.468 & 0.354 & 0.427 & 0.304 & \textbf{0.404} & \textbf{0.286} & 0.670 & 0.421 & 0.795 & 0.481 \\
% & 192 & 0.434 & 0.305 & 0.479 & 0.352 & 0.447 & 0.315 & \textbf{0.412} & \textbf{0.294} & 0.653 & 0.405 & 0.837 & 0.503 \\
% & 336 & 0.449 & 0.313 & 0.477 & 0.345 & 0.478 & 0.333 & \textbf{0.439} & \textbf{0.310} & 0.707 & 0.445 & 0.867 & 0.523 \\
% & 720 & - & - & - & - & - & - & - & - & - & - & - & - \\
% \hline
\bottomrule
\end{tabular}
}
\end{small}
\end{center}
\end{table}

\subsection{Fine-Tuning Parameters Selection}
In this section, we conduct ablation experiments to study which parameters are important to fine-tune. Since the input embedding and output layers are randomly initialized for adapting to a new domain, they must be trained. Then, we study adding layer normalization and positional embeddings to the list of fine-tuning parameters. Table \ref{tab:ablation} shows the results that re-train parameters of layer normalization and positional embeddings can bring certain benefits, especially in longer prediction lengths. Thus, we follow the standard practice to re-train positional embeddings and layer normalization.

\begin{table}[!h]
\caption{Ablation by fixing positional embeddings or layer normalization on 5\% ETTm1 and ETTm2. Parameters of GPT2(6) are successively added to the list of fine-tuned parameters.}
\label{tab:ablation}
\vskip 0.15in
\begin{center}
\begin{small}
\scalebox{0.8}{
\begin{tabular}{c|c|cccccc}
\toprule

\multicolumn{2}{c}{Methods}&\multicolumn{2}{c}{Input \& Output}&\multicolumn{2}{c}{+ LN}&\multicolumn{2}{c}{+ POS}\\

\midrule

\multicolumn{2}{c|}{Metric} & MSE  & MAE & MSE & MAE& MSE  & MAE\\
\midrule

% \multirow{4}{*}{\rotatebox{90}{$ETTh2$}}
% & 96  &0.394&0.424&0.376&0.420&0.376&0.421\\
% & 192 &&&0.418&0.441&0.418&0.441\\
% & 336 &&&&&0.408&0.439\\
% & 720 &-&-&-&-&-&-\\\midrule

\multirow{4}{*}{\rotatebox{90}{$ETTm1$}}
& 96  &0.395&0.410&0.392&0.409&0.386&0.405\\
& 192 &0.444&0.438&0.436&0.435&0.440&0.438\\
& 336 &0.510&0.472&0.495&0.467&0.485&0.459\\
& 720 &0.607&0.517&0.564&0.503&0.557&0.499\\\midrule

\multirow{4}{*}{\rotatebox{90}{$ETTm2$}}
& 96  &0.198&0.282&0.198&0.279&0.199&0.280\\
& 192 &0.261&0.324&0.263&0.325&0.256&0.316\\
& 336 &0.336&0.377&0.322&0.356&0.318&0.353\\
& 720 &0.473&0.444&0.457&0.435&0.460&0.436\\

\bottomrule
\end{tabular}
}
\end{small}
\end{center}
\vskip -0.1in
\end{table}

\subsection{Analysis of Data Volume}

\begin{figure}[h]
    \centering
    \includegraphics[width=0.6\columnwidth]{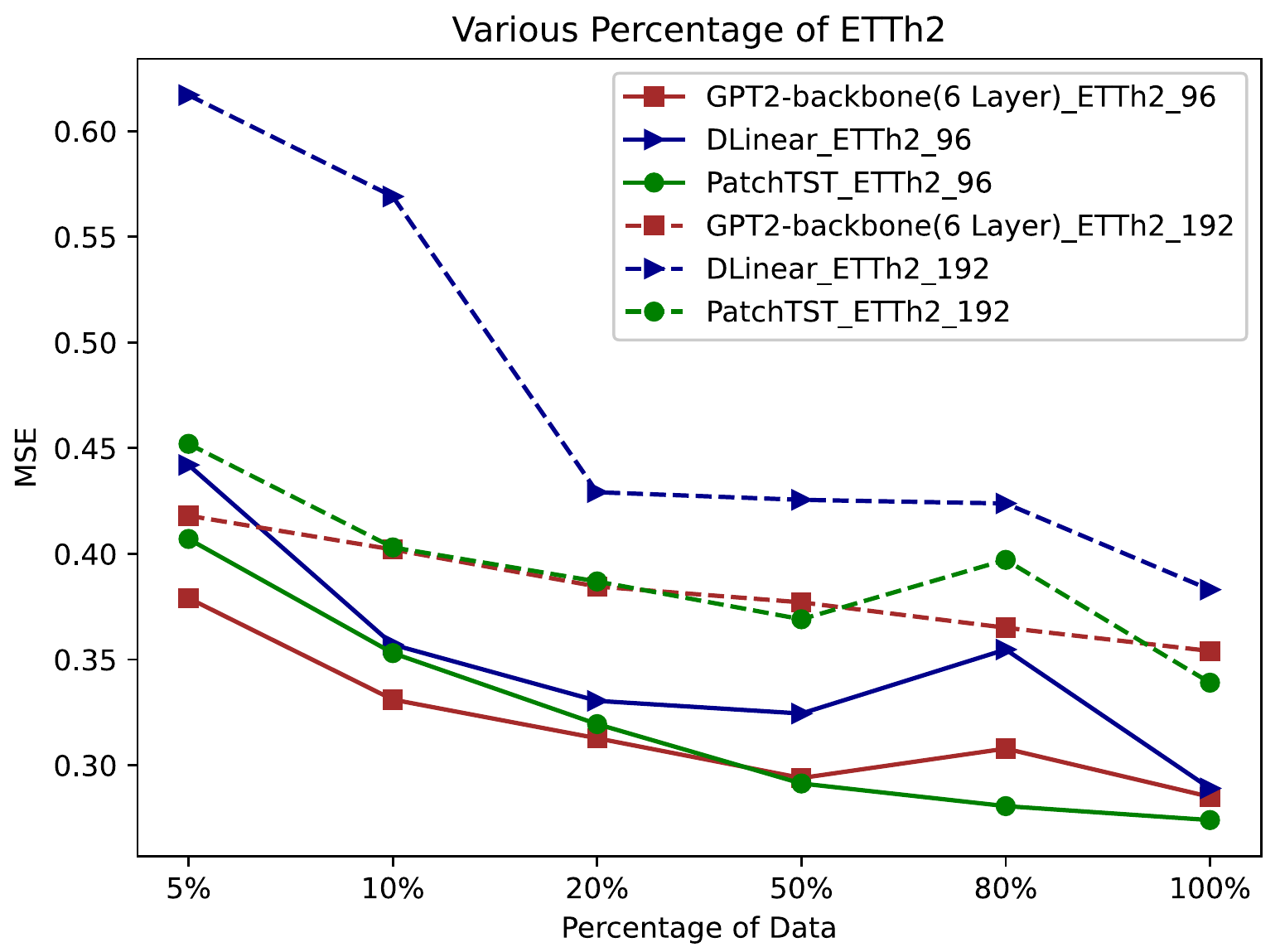}
    \caption{Results on various percentages of ETTh2. Line color represents different models and line style means various prediction lengths $O \in \{ 96, 192\}$.}
    \label{fig:different_volume}
\end{figure}

Results of few-shot learning show that GPT2(6) FPT shows SOTA performance in few-shot learning tasks in which the model is trained on 5\% data and 10\% data. Plus, it has comparable performance with the SOTA baselines PatchTST and Dlinear on full sample forecasting setting as well. This phenomenon raises a question that how performance changes with an increase in  data sample size. Thus, we conduct experiments on various percentages $P \in \{5\%, 10\%, 20\%, 50\%, 80\%, 100\%\}$ of ETTh2. Figure \ref{fig:different_volume} shows that the performance improvement for GPT2(6) FPT is almost flattened. These results illustrate that such a cross-domain FPT model is extremely efficient in few-shot time series forecasting and only requires a few fine-tuning samples to reach a SOTA performance. For more complete data, end-to-end training models start to catch up, but still, a GPT2(6) FPT model can be comparable to those SOTA end-to-end training algorithms.

\subsection{Knowledge transfer with other Pre-trained Transformer Models}
\label{app:other_model}

We investigate how other pre-trained transformer models perform and whether other domains can also help. Another NLP pre-trained model BERT \cite{Bert/NAACL/Jacob} and the CV pre-trained model BEiT \cite{bao2022beit} are trained on 5\% ETTh2 and 5\% ETTm2. Similar to GPT2, we only reserve 6 layers and freeze attention blocks. Our results are shown in Table \ref{tab:ett_other_model} that BERT(6) FPT and BEiT(6) FPT are comparable to PatchTST and remarkably surpass other baselines. We come to the conclusion that the universality of our proposed architecture holds across other pre-trained-transformer models. Moreover, the domain of successful knowledge transfer in time series forecasting is not limited to natural language. Knowledge from the CV domain can also help, supported by BEiT's experimental results.

%\resizebox{!}{\.5\paperheight}{
\begin{table}[h]
\setlength\tabcolsep{3pt} 
\centering
\begin{small}
% \begin{adjustwidth}{-1.5in}{-1in}
\caption{Results of frozen pretrained transformer variants on 5\% ETTh2 and ETTm2.  We use prediction length $O \in \{96, 192, 336, 720\}$. A lower MSE indicates better performance. \textbf{Black}: best, {\color{red} \textbf{Red}}: second best, {\color{violet} \textbf{Violet}}: third best. '-' means that 5\% time series is not sufficient to constitute a training set. }
\vskip 0.05in
\scalebox{0.70}{
\begin{tabular}{c|c|cccc|cccc}
\toprule
\multirow{2}{*}{Methods} & \multirow{2}{*}{Metric} &\multicolumn{4}{c|}{ETTh2}&\multicolumn{4}{c}{ETTm2}\\
% \midrule
% \multicolumn{2}{c|}{Metric} & MSE  & MAE & MSE & MAE& MSE  & MAE& MSE  & MAE& MSE  & MAE& MSE  & MAE\\
& &96 & 192 & 336 & 720 & 96 & 192 & 336 & 720 \\
\midrule
\multirow{2}{*}{GPT2-backbone(6 Layers)} & MSE &\textbf{0.376}&\textbf{0.421}&\textbf{0.408}&-&\textbf{0.199}&\textbf{0.256}&\textbf{0.318}&0.460\\
& MAE &\color{violet}\textbf{0.419}&\textbf{0.441}&\textbf{0.439}&-&\textbf{0.280}&\textbf{0.316}&\textbf{0.353}&\color{violet}\textbf{0.436}\\\midrule

\multirow{2}{*}{BERT-backbond(6 Layers)} & MSE &\color{violet}\textbf{0.397}&0.480&0.481&-&0.222&0.281&\color{red}\textbf{0.331}&\textbf{0.441}\\
& MAE &\textbf{0.418}&0.465&\color{violet}\textbf{0.472}&-&0.300&0.335&\color{violet}\textbf{0.367}&\textbf{0.428}\\\midrule

\multirow{2}{*}{BEiT-backbond(6 Layers)} & MSE &0.405&\color{red}\textbf{0.448}&0.524&-&\color{violet}\textbf{0.208}&\color{violet}\textbf{0.272}&\color{red}\textbf{0.331}&\color{red}\textbf{0.452}\\
& MAE &\textbf{0.418}&\color{red}\textbf{0.446}&0.500&-&\color{violet}\textbf{0.291}&\color{violet}\textbf{0.326}&\color{red}\textbf{0.362}&\color{red}\textbf{0.433}\\\midrule

\multirow{2}{*}{DLinear\cite{dlinear}} & MSE &0.442&0.617&1.424&-&0.236&0.306&0.380&0.674\\
& MAE &0.456&0.542&0.849&-&0.326&0.373&0.423&0.583\\\midrule

\multirow{2}{*}{PatchTST\cite{Patchformer}} & MSE &0.401&\color{violet}\textbf{0.452}&\color{red}\textbf{0.464}&-&\color{red}\textbf{0.206}&\color{red}\textbf{0.264}&\color{violet}\textbf{0.334}&\color{violet}\textbf{0.454}\\
& MAE &0.421&\color{violet}\textbf{0.455}&\color{red}\textbf{0.469}&-&\color{red}\textbf{0.288}&\color{red}\textbf{0.324}&\color{violet}\textbf{0.367}&0.483\\\midrule

\multirow{2}{*}{FEDformer\cite{zhou2022fedformer}} & MSE &\color{red}\textbf{0.390}&0.457&\color{violet}\textbf{0.477}&-&0.299&0.290&0.378&0.523\\
& MAE &0.424&0.465&0.483&-&0.320&0.361&0.427&0.510\\\midrule

\multirow{2}{*}{Autoformer\cite{wu2021autoformer}} & MSE &0.428&0.496&0.486&-&0.232&0.291&0.478&0.533\\
& MAE &0.468&0.504&0.496&-&0.322&0.357&0.517&0.538\\

\bottomrule
\end{tabular}
\label{tab:ett_other_model}
}
% \end{adjustwidth}
\end{small}
\vskip -0.1in
\end{table}
%}

\subsection{Full Results of Classification}
\label{appendix:classification_full}
\begin{table}[h]
\caption{Full results for the classification task. $\ast$. in the Transformers indicates the name of $\ast$former.}
\label{tab:classification}
\vskip 0.15in
\begin{center}
\begin{small}
\scalebox{0.65}{
\setlength\tabcolsep{3pt}
\begin{tabular}{c|cc|cc|c|ccccccccc|cc|c|c}
\hline

\multirow{2}{*}{Methods} &
\multicolumn{2}{c|}{Classical methods} & \multicolumn{2}{c|}{RNN} & \multirow{2}{*}{TCN}& \multicolumn{9}{c|}{Transformers} &\multicolumn{2}{c|}{MLP} & \multirow{2}{*}{TimesNet} & \multirow{2}{*}{GPT2(6)} \\
&XGBoost&Rocket&LSTNet& LSSL& & Trans.& Re.& In.& Pyra.& Auto.& Station.& FED.& ETS.& Flow. &DLinear &LightTS. & &  \\

\hline
EthanolConcentration&43.7 &45.2 &39.9 &31.1 &28.9 &32.7 &31.9 &31.6 &30.8 &31.6 &32.7 &31.2 &28.1 &33.8 &32.6 &29.7 &35.7 &34.2\\
FaceDetection&63.3 &64.7 &65.7 &66.7 &52.8 &67.3 &68.6 &67.0 &65.7 &68.4 &68.0 &66.0 &66.3 &67.6 &68.0 &67.5 &68.6 &69.2\\
Handwriting&15.8 &58.8 &25.8 &24.6 &53.3 &32.0 &27.4 &32.8 &29.4 &36.7 &31.6 &28.0 &32.5 &33.8 &27.0 &26.1 &32.1 &32.7\\
Heartbeat&73.2 &75.6 &77.1 &72.7 &75.6 &76.1 &77.1 &80.5 &75.6 &74.6 &73.7 &73.7 &71.2 &77.6 &75.1 &75.1 &78.0 &77.2\\
JapaneseVowels&86.5 &96.2 &98.1 &98.4 &98.9 &98.7 &97.8 &98.9 &98.4 &96.2 &99.2 &98.4 &95.9 &98.9 &96.2 &96.2 &98.4 &98.6\\
PEMS-SF&98.3 &75.1 &86.7 &86.1 &68.8 &82.1 &82.7 &81.5 &83.2 &82.7 &87.3 &80.9 &86.0 &83.8 &75.1 &88.4 &89.6 &87.9\\
SelfRegulationSCP1&84.6 &90.8 &84.0 &90.8 &84.6 &92.2 &90.4 &90.1 &88.1 &84.0 &89.4 &88.7 &89.6 &92.5 &87.3 &89.8 &91.8 &93.2\\
SelfRegulationSCP2&48.9 &53.3 &52.8 &52.2 &55.6 &53.9 &56.7 &53.3 &53.3 &50.6 &57.2 &54.4 &55.0 &56.1 &50.5 &51.1 &57.2 &59.4\\
SpokenArabicDigits&69.6 &71.2 &100.0 &100.0 &95.6 &98.4 &97.0 &100.0 &99.6 &100.0 &100.0 &100.0 &100.0 &98.8 &81.4 &100.0 &99.0 &99.2\\
UWaveGestureLibrary&75.9 &94.4 &87.8 &85.9 &88.4 &85.6 &85.6 &85.6 &83.4 &85.9 &87.5 &85.3 &85.0 &86.6 &82.1 &80.3 &85.3 &88.1\\
\hline
Average &66.0 &72.5 &71.8 &70.9 &70.3 &71.9 &71.5 &72.1 
&70.8 &71.1 &72.7 &70.7 &71.0 &73.0 &67.5 &70.4 &\color{red}\textbf{73.6} &\textbf{74.0}\\
\hline

\end{tabular}
}
\end{small}
\end{center}
\vskip -0.1in
\end{table}

\subsection{Full Results of Anomaly Detection}
\label{appendix:anomaly_full}
\begin{table}[h]
\caption{Full results for the anomaly detection.}
\label{tab:anomaly_full}
\vskip 0.15in
\begin{center}
\begin{small}
\scalebox{0.65}{
\begin{threeparttable}[b]
\begin{tabular}{c|ccc|ccc|ccc|ccc|ccc|c}
\toprule

Methods &
\multicolumn{3}{c|}{SMD} & \multicolumn{3}{c|}{MSL} & \multicolumn{3}{c|}{SMAP}& \multicolumn{3}{c|}{SWaT} &\multicolumn{3}{c|}{PSM} & Avg F1 \\
Metrics&P&R&F1&P&R&F1&P&R&F1&P&R&F1&P&R&F1&\%  \\

\midrule
GPT(6)&{\bf88.89}&{\bf84.98}&{\bf86.89}&82.00&{\bf82.91}&{\bf82.45}&{\bf90.60}&{\bf60.95}&{\bf72.88}&{\bf92.20}&{\bf96.34}&{\bf94.23}&{\bf98.62}&95.68&97.13&{\bf86.72}\\
TimesNet\tnote{*} &87.91&81.54&84.61&{\bf89.54}&75.36&81.84&90.14&56.40&69.39&90.75&95.40&93.02&98.51&{\bf96.20}&{\bf97.34}&85.24\\
PatchTST&87.26&82.14&84.62&88.34&70.96&78.70&90.64&55.46&68.82&91.10&80.94&85.72&98.84&93.47&96.08&82.79\\
ETSformer&87.44&79.23&83.13&85.13&84.93&85.03&92.25&55.75&69.50&90.02&80.36&84.91&99.31&85.28&91.76&82.87\\
FEDformer&87.95&82.39&85.08&77.14&80.07&78.57&90.47&58.10&70.76&90.17&96.42&93.19&97.31&97.16&97.23&84.97\\
LightTS&87.10&78.42&82.53&82.40&75.78&78.95&92.58&55.27&69.21&91.98&94.72&93.33&98.37&95.97&97.15&84.23 \\
DLinear&83.62&71.52&77.10&84.34&85.42&84.88&92.32&55.41&69.26&80.91&95.30&87.52&98.28&89.26&93.55&82.46 \\
Stationary&88.33&81.21&84.62&68.55&89.14&77.50&89.37&59.02&71.09&68.03&96.75&79.88&97.82&96.76&97.29&82.08 \\
Autoformer&88.06&82.35&85.11&77.27&80.92&79.05&90.40&58.62&71.12&89.85&95.81&92.74&99.08&88.15&93.29&84.26 \\
Pyraformer&85.61&80.61&83.04&83.81&85.93&84.86&92.54&57.71&71.09&87.92&96.00&91.78&71.67&96.02&82.08&82.57 \\
Anomaly Transformer\tnote{**}&88.91&82.23&85.49&79.61&87.37&83.31&91.85&58.11&71.18&72.51&97.32&83.10&68.35&94.72&79.40&80.50 \\
Informer&86.60&77.23&81.65&81.77&86.48&84.06&90.11&57.13&69.92&70.29&96.75&81.43&64.27&96.33&77.10&78.83 \\
Reformer&82.58&69.24&75.32&85.51&83.31&84.40&90.91&57.44&70.40&72.50&96.53&82.80&59.93&95.38&73.61&77.31 \\
LogTransformer&83.46&70.13&76.21&73.05&87.37&79.57&89.15&57.59&69.97&68.67&97.32&80.52&63.06&98.00&76.74&76.60 \\
Transformer&83.58&76.13&79.56&71.57&87.37&78.68&89.37&57.12&69.70&68.84&96.53&80.37&62.75&96.56&76.07&76.88 \\

\bottomrule

\end{tabular}

\begin{tablenotes}
 \item[*] We reproduce the results of TimesNet by \href{https://github.com/thuml/Time-Series-Library}{https://github.com/thuml/Time-Series-Library}. 
 \item[**] We replace the joint criterion in Anomaly Transformer with reconstruction error for fair comparison.
\end{tablenotes}
\end{threeparttable}

}
\end{small}
\end{center}
\vskip -0.1in
\end{table}

\subsection{Full Results of Imputation}
\label{appendix:inputation_full}
\begin{table}[h]
\caption{Full results for the imputation task.}
% We randomly mask {12.5\%, 25\%, 37.5\%, 50\%} to compare the model performance under different missing degrees. A lower MSE indicates better performance. \textbf{Black}: best,  {\color{red}\textbf{Red}}: second best.}
\label{tab:imputation_full}
\begin{center}
\begin{small}
\scalebox{0.65}{
\setlength\tabcolsep{3pt}
\begin{tabular}{cc|cc|cc|cc|cc|cc|cc|cc|cc|cc|cc|cc}
\toprule

\multicolumn{2}{c|}{Methods} 
&\multicolumn{2}{c|}{GPT2(3)} & \multicolumn{2}{c|}{TimesNet}& \multicolumn{2}{c|}{PatchTST}&\multicolumn{2}{c|}{ETSformer}&\multicolumn{2}{c|}{LightTS}&\multicolumn{2}{c|}{DLinear}&\multicolumn{2}{c|}{FEDformer}&\multicolumn{2}{c|}{Stationary}&\multicolumn{2}{c|}{Autoformer}&\multicolumn{2}{c}{Informer}&\multicolumn{2}{c}{Reformer} \\
Mask&Ratio&MSE&MAE&MSE&MAE&MSE&MAE&MSE&MAE&MSE&MAE&MSE&MAE&MSE&MAE&MSE&MAE&MSE&MAE&MSE&MAE&MSE&MAE \\

\midrule
\multirow{5}{*}{\rotatebox{90}{$ETTm1$}}
& 12.5\% &{\bf 0.017}&{\bf0.085}&0.023&0.101&0.041&0.130&0.096&0.229&0.093&0.206&0.080&0.193&0.052&0.166&0.032&0.119&0.046&0.144&0.063&0.180&0.042&0.146 \\
& 25\% &{\bf0.022}&{\bf0.096}&0.023&0.101&0.044&0.135&0.096&0.229&0.093&0.206&0.080&0.193&0.052&0.166&0.032&0.119&0.046&0.144&0.063&0.180&0.042&0.146 \\
& 37.5\% &{\bf0.029}&{\bf0.111}&0.029&0.111&0.049&0.143&0.133&0.271&0.113&0.231&0.103&0.219&0.069&0.191&0.039&0.131&0.057&0.161&0.079&0.200&0.063&0.182 \\
& 50\% &0.040&0.128&{\bf0.036}&{\bf0.124}&0.055&0.151&0.186&0.323&0.134&0.255&0.132&0.248&0.089&0.218&0.047&0.145&0.067&0.174&0.093&0.218&0.082&0.208 \\
& Avg &0.028&{\bf0.105}&{\bf0.027}&0.107&0.047&0.140&0.120&0.253&0.104&0.218&0.093&0.206&0.062&0.177&0.036&0.126&0.051&0.150&0.071&0.188&0.055&0.166 \\
\midrule

\multirow{5}{*}{\rotatebox{90}{$ETTm2$}}
& 12.5\% &{\bf0.017}&{\bf0.076}&0.018&0.080&0.026&0.094&0.108&0.239&0.034&0.127&0.062&0.166&0.056&0.159&0.021&0.088&0.023&0.092&0.133&0.270&0.108&0.228 \\
& 25\% &{\bf0.020}&{\bf0.080}&0.020&0.085&0.028&0.099&0.164&0.294&0.042&0.143&0.085&0.196&0.080&0.195&0.024&0.096&0.026&0.101&0.135&0.272&0.136&0.262 \\
& 37.5\% &{\bf0.022}&{\bf0.087}&0.023&0.091&0.030&0.104&0.237&0.356&0.051&0.159&0.106&0.222&0.110&0.231&0.027&0.103&0.030&0.108&0.155&0.293&0.175&0.300 \\
& 50\% &{\bf0.025}&{\bf0.095}&0.026&0.098&0.034&0.110&0.323&0.421&0.059&0.174&0.131&0.247&0.156&0.276&0.030&0.108&0.035&0.119&0.200&0.333&0.211&0.329 \\
& Avg &{\bf0.021}&{\bf0.084}&0.022&0.088&0.029&0.102&0.208&0.327&0.046&0.151&0.096&0.208&0.101&0.215&0.026&0.099&0.029&0.105&0.156&0.292&0.157&0.280 \\
\midrule

\multirow{5}{*}{\rotatebox{90}{$ETTh1$}}
& 12.5\% &{\bf0.043}&{\bf0.140}&0.057&0.159&0.093&0.201&0.126&0.263&0.240&0.345&0.151&0.267&0.070&0.190&0.060&0.165&0.074&0.182&0.114&0.234&0.074&0.194 \\
& 25\% &{\bf0.054}&{\bf0.156}&0.069&0.178&0.107&0.217&0.169&0.304&0.265&0.364&0.180&0.292&0.106&0.236&0.080&0.189&0.090&0.203&0.140&0.262&0.102&0.227 \\
& 37.5\% &{\bf0.072}&{\bf0.180}&0.084&0.196&0.120&0.230&0.220&0.347&0.296&0.382&0.215&0.318&0.124&0.258&0.102&0.212&0.109&0.222&0.174&0.293&0.135&0.261 \\
& 50\% &0.107&0.216&{\bf0.102}&{\bf0.215}&0.141&0.248&0.293&0.402&0.334&0.404&0.257&0.347&0.165&0.299&0.133&0.240&0.137&0.248&0.215&0.325&0.179&0.298 \\
& Avg &{\bf0.069}&{\bf0.173}&0.078&0.187&0.115&0.224&0.202&0.329&0.284&0.373&0.201&0.306&0.117&0.246&0.094&0.201&0.103&0.214&0.161&0.279&0.122&0.245 \\
\midrule

\multirow{5}{*}{\rotatebox{90}{$ETTh2$}}
& 12.5\% &{\bf0.039}&{\bf0.125}&0.040&0.130&0.057&0.152&0.187&0.319&0.101&0.231&0.100&0.216&0.095&0.212&0.042&0.133&0.044&0.138&0.305&0.431&0.163&0.289 \\
& 25\% &{\bf0.044}&{\bf0.135}&0.046&0.141&0.061&0.158&0.279&0.390&0.115&0.246&0.127&0.247&0.137&0.258&0.049&0.147&0.050&0.149&0.322&0.444&0.206&0.331 \\
& 37.5\% &{\bf0.051}&{\bf0.147}&0.052&0.151&0.067&0.166&0.400&0.465&0.126&0.257&0.158&0.276&0.187&0.304&0.056&0.158&0.060&0.163&0.353&0.462&0.252&0.370 \\
& 50\% &{\bf0.059}&{\bf0.158}&0.060&0.162&0.073&0.174&0.602&0.572&0.136&0.268&0.183&0.299&0.232&0.341&0.065&0.170&0.068&0.173&0.369&0.472&0.316&0.419 \\
& Avg &{\bf0.048}&{\bf0.141}&0.049&0.146&0.065&0.163&0.367&0.436&0.119&0.250&0.142&0.259&0.163&0.279&0.053&0.152&0.055&0.156&0.337&0.452&0.234&0.352 \\
\midrule

\multirow{5}{*}{\rotatebox{90}{$ECL$}}
& 12.5\% &{\bf0.080}&{\bf0.194}&0.085&0.202&0.055&0.160&0.196&0.321&0.102&0.229&0.092&0.214&0.107&0.237&0.093&0.210&0.089&0.210&0.218&0.326&0.190&0.308 \\
& 25\% &{\bf0.087}&{\bf0.203}&0.089&0.206&0.065&0.175&0.207&0.332&0.121&0.252&0.118&0.247&0.120&0.251&0.097&0.214&0.096&0.220&0.219&0.326&0.197&0.312 \\
& 37.5\%  &0.094&{\bf0.211}&{\bf0.094}&0.213&0.076&0.189&0.219&0.344&0.141&0.273&0.144&0.276&0.136&0.266&0.102&0.220&0.104&0.229&0.222&0.328&0.203&0.315\\
& 50\% &0.101&{\bf0.220}&{\bf0.100}&0.221&0.091&0.208&0.235&0.357&0.160&0.293&0.175&0.305&0.158&0.284&0.108&0.228&0.113&0.239&0.228&0.331&0.210&0.319 \\
& Avg &{\bf0.090}&{\bf0.207}&0.092&0.210&0.072&0.183&0.214&0.339&0.131&0.262&0.132&0.260&0.130&0.259&0.100&0.218&0.101&0.225&0.222&0.328&0.200&0.313 \\
\midrule

\multirow{5}{*}{\rotatebox{90}{$Weather$}}
& 12.5\% &0.026&0.049&{\bf0.025}&{\bf0.045}&0.029&0.049&0.057&0.141&0.047&0.101&0.039&0.084&0.041&0.107&0.027&0.051&0.026&0.047&0.037&0.093&0.031&0.076 \\
& 25\% &{\bf0.028}&{\bf0.052}&0.029&0.052&0.031&0.053&0.065&0.155&0.052&0.111&0.048&0.103&0.064&0.163&0.029&0.056&0.030&0.054&0.042&0.100&0.035&0.082 \\
& 37.5\% &0.033&0.060&{\bf0.031}&{\bf0.057}&0.035&0.058&0.081&0.180&0.058&0.121&0.057&0.117&0.107&0.229&0.033&0.062&0.032&0.060&0.049&0.111&0.040&0.091 \\
& 50\% &0.037&0.065&{\bf0.034}&{\bf0.062}&0.038&0.063&0.102&0.207&0.065&0.133&0.066&0.134&0.183&0.312&0.037&0.068&0.037&0.067&0.053&0.114&0.046&0.099 \\
& Avg &0.031&0.056&{\bf0.030}&{\bf0.054}&0.060&0.144&0.076&0.171&0.055&0.117&0.052&0.110&0.099&0.203&0.032&0.059&0.031&0.057&0.045&0.104&0.038&0.087 \\
\bottomrule

\end{tabular}
}
\end{small}
\end{center}
\end{table}

\subsection{Full Results of Short-term Forecasting}
\label{appendix:short-term_full}
\begin{table*}[h]
\renewcommand\arraystretch{1.5}
\vskip -0.10in
\captionsetup{font=small} 
\caption{Full results of short-term forecasting.}
\label{tab:short_term_full}
%\vskip 0.15in
\begin{center}
\begin{small}
\scalebox{0.7}{
\setlength\tabcolsep{3pt}
\begin{tabular}{cc|ccccccccccccc}
\toprule

\multicolumn{2}{c|}{Methods}&GPT2(6)&TimesNet&PatchTST&N-HiTS&N-BEATS& ETSformer& LightTS& DLinear &FEDformer &Stationary &Autoformer  &Informer&Reformer \\

% \multirow{2}{*}{Methods} 
% &\multicolumn{2}{c|}{GPT2(6)} & \multicolumn{2}{c|}{TimesNet}&\multicolumn{2}{c|}{ETSformer}&\multicolumn{2}{c|}{ETSformer}&\multicolumn{2}{c|}{LightTS}&\multicolumn{2}{c|}{DLinear}&\multicolumn{2}{c|}{FEDformer}&\multicolumn{2}{c|}{Stationary}&\multicolumn{2}{c|}{Autoformer}&\multicolumn{2}{c}{Informer}&\multicolumn{2}{c}{Reformer} \\

\midrule
\multirow{3}{*}{\rotatebox{90}{$Yearly$}}
&SMAPE&13.531&{\bf13.387}&13.477&13.418&13.436&18.009&14.247&16.965&13.728&13.717&13.974&14.727&16.169\\
&MASE&3.015&{\bf2.996}&3.019&3.045&3.043&4.487&3.109&4.283&3.048&3.078&3.134&3.418&3.800\\
&OWA&0.793&{\bf0.786}&0.792&0.793&0.794&1.115&0.827&1.058&0.803&0.807&0.822&0.881&0.973\\
\bottomrule

\multirow{3}{*}{\rotatebox{90}{$Quarterly$}}
&SMAPE&10.177&{\bf10.100}&10.38&10.202&10.124&13.376&11.364&12.145&10.792&10.958&11.338&11.360&13.313\\
&MASE&1.194&{\bf1.182}&1.233&1.194&1.169&1.906&1.328&1.520&1.283&1.325&1.365&1.401&1.775\\
&OWA&0.898&{\bf0.890}&0.921&0.899&0.886&1.302&1.000&1.106&0.958&0.981&1.012&1.027&1.252\\
\bottomrule

\multirow{3}{*}{\rotatebox{90}{$Monthly$}}
&SMAPE&12.894&{\bf12.670}&12.959&12.791&12.677&14.588&14.014&13.514&14.260&13.917&13.958&14.062&20.128\\
&MASE&0.956&{\bf0.933}&0.970&0.969&0.937&1.368&1.053&1.037&1.102&1.097&1.103&1.141&2.614\\
&OWA&0.897&{\bf0.878}&0.905&0.899&0.880&1.149&0.981&0.956&1.012&0.998&1.002&1.024&1.927\\
\bottomrule

\multirow{3}{*}{\rotatebox{90}{$Others$}}
&SMAPE&4.940&{\bf4.891}&4.952&5.061&4.925&7.267&15.880&6.709&4.954&6.302&5.485&24.460&32.491\\
&MASE&3.228&3.302&3.347&{\bf3.216}&3.391&5.240&11.434&4.953&3.264&4.064&3.865&20.960&33.355\\
&OWA&1.029&{\bf1.035}&1.049&1.040&1.053&1.591&3.474&1.487&1.036&1.304&1.187&5.879&8.679\\
\bottomrule

\multirow{3}{*}{\rotatebox{90}{$Average$}}
&SMAPE &11.991 &{\bf11.829}&12.059& 11.927& 11.851& 14.718& 13.525& 13.639 &12.840 &12.780 &12.909 &14.086 &18.200 \\
&MASE & 1.600 & {\bf1.585}&1.623 & 1.613 & 1.599 &2.408 &2.111 &2.095 &1.701 &1.756 &1.771  &2.718 &4.223\\
&OWA &0.861 & {\bf0.851}&0.869 &0.861 &0.855 &1.172 &1.051 &1.051 &0.918 &0.930 &0.939 & 1.230 & 1.775\\

\bottomrule

\end{tabular}
}
\end{small}
\end{center}
\vskip -0.25in
\end{table*}

\subsection{Input Length Setting Discussion}
\label{app:input_length}
The consideration of input length is of great importance. It is widely believed that longer input lengths have the potential to generate superior results. However, in practice, certain algorithms  might fall short in effectively utilizing long input signals due to overfitting issues either. Here, we conducted long-term forecasting experiments by comparing our approach with the best reported values from different baseline papers. This was done to avoid any biases stemming from selective tuning of baseline parameters. While some may argue in favor of a fairer comparison using a fixed input length, we are starting to shift our focus towards pursuing more accurate algorithms that possess enhanced capabilities for handling longer inputs. Instead of restricting the input length to a fixed small value, it is pragmatic to tune both the input length and model parameters based on performance, as it is often the primary concern in practical usage. Exploring the utilization of extremely long inputs, such as Chatgpt or LLM, is among our future research directions.
% You can have as much text here as you want. The main body must be at most $8$ pages long.
% For the final version, one more page can be added.
% If you want, you can use an appendix like this one, even using the one-column format.
%%%%%%%%%%%%%%%%%%%%%%%%%%%%%%%%%%%%%%%%%%%%%%%%%%%%%%%%%%%%%%%%%%%%%%%%%%%%%%%
%%%%%%%%%%%%%%%%%%%%%%%%%%%%%%%%%%%%%%%%%%%%%%%%%%%%%%%%%%%%%%%%%%%%%%%%%%%%%%%

\end{document}